\documentclass[10pt,journal,compsoc]{IEEEtran}

\usepackage{mwe}
\usepackage{microtype}
\usepackage{graphicx}
\usepackage{subcaption}
\usepackage{hyperref}
\usepackage{soul}

\usepackage{multirow}
\usepackage{multicol}
\usepackage{siunitx}
\usepackage{amsmath}
\usepackage{amssymb}
\usepackage{esint}
\usepackage{times}
\usepackage{epsfig}
\usepackage{epstopdf}
\usepackage{gensymb}

\usepackage{dirtree}

\usepackage[utf8]{inputenc} 
\usepackage[T1]{fontenc}    
\usepackage{url}            
\usepackage{amsfonts}       
\usepackage{nicefrac}       
\usepackage{xcolor}
\usepackage{wrapfig,lipsum,booktabs}
\usepackage[makeroom]{cancel}
\usepackage{diagbox}

\usepackage{xspace}

\usepackage{flushend}

\DeclareMathOperator*{\argmax}{argmax}

\newcommand{\ie}{\emph{i.e.}\xspace}
\newcommand{\etc}{\emph{etc}\xspace}
\newcommand{\eg}{\emph{e.g.}\xspace}

\newcommand{\Eq}{Eq.\xspace}
\newcommand{\Fig}{Fig.\xspace}

\newcommand{\Sec}{Sec.\xspace}
\newcommand{\Tab}{Tab.\xspace}

\newcommand{\revised}[1]{\textcolor{black}{#1}}
\newcommand{\revisedTwo}[1]{\textcolor{black}{#1}}
\newcommand{\revisedThree}[1]{\textcolor{black}{#1}}
\newcommand{\revisedF}[1]{\textcolor{black}{#1}}

\newif\ifarxiv
\arxivfalse

\newcommand{\appropto}{\mathrel{\vcenter{
			\offinterlineskip\halign{\hfil$##$\cr
				\propto\cr\noalign{\kern2pt}\sim\cr\noalign{\kern-2pt}}}}}

\ifCLASSOPTIONcompsoc
  \usepackage[nocompress]{cite}
\else
  \usepackage{cite}
\fi

\ifCLASSINFOpdf
  
\else
  
\fi

\DeclareCaptionFormat{cont}{#1 (cont.)#2#3\par}

\makeatletter
\newcommand\footnoteref[1]{\protected@xdef\@thefnmark{\ref{#1}}\@footnotemark}
\makeatother

\begin{document}

\title{JRDB: A Dataset and Benchmark of\\Egocentric \revisedThree{Robot Visual Perception of Humans in Built Environments}
}

\author{Roberto Mart\'in-Mart\'in$^*$, Mihir Patel$^*$, Hamid Rezatofighi$^*$, Abhijeet Shenoi, JunYoung Gwak\\Eric Frankel, Amir Sadeghian, Silvio Savarese
\IEEEcompsocitemizethanks{\IEEEcompsocthanksitem$^*$ indicates equal contribution.
\IEEEcompsocthanksitem Roberto Mart\'in-Mart\'in, Mihir Patel, Abhijeet Shenoi, JunYoung Gwak, Eric Frankel, Amir Sadeghian and Silvio Savarese are with the Stanford Vision and Learning Laboratory, Stanford University, USA.\protect\\E-mail:\texttt{[robertom,mihirp,ashenoi,jgwak,esfrankel,} \texttt{amirabs,ssilvio]@cs.stanford.edu}.
\IEEEcompsocthanksitem Hamid Rezatofighi is with the department of Data Science and AI, Faculty of Information Technology, Monash University, Clayton, VIC, 3800, Australia.\protect\\
Email: \texttt{hamid.rezatofighi@monash.edu.}
}}

\markboth{}%
{Shell \MakeLowercase{\textit{et al.}}: Bare Advanced Demo of IEEEtran.cls for IEEE Computer Society Journals}

\IEEEtitleabstractindextext{%
\begin{abstract}
We present JRDB, a novel egocentric dataset collected from our social mobile manipulator JackRabbot. The dataset includes 64 minutes of annotated multimodal sensor data including stereo cylindrical 360\degree RGB video at 15 fps, 3D point clouds from two 16 planar rays Velodyne LiDARs, line 3D point clouds from two Sick Lidars, audio signal, RGB-D video at 30 fps, 360\degree spherical image from a fisheye camera and encoder values from the robot's wheels. Our dataset incorporates data from traditionally underrepresented scenes such as indoor environments and pedestrian areas, all from the ego-perspective of the robot, both stationary and navigating. The dataset has been annotated with over 2.4 million bounding boxes spread over 5 individual cameras and 1.8 million associated 3D cuboids around all people in the scenes totaling over 3500 time consistent trajectories. Together with our dataset and the annotations, we launch a benchmark and metrics for 2D and 3D person detection and tracking. With this dataset, which we plan on extending with further types of annotation in the future, we hope to provide a new source of data and a test-bench for research in the areas of egocentric robot vision, autonomous navigation, and all perceptual tasks around social robotics in human environments.
\end{abstract}

}

\maketitle

\IEEEdisplaynontitleabstractindextext

%
\IEEEpeerreviewmaketitle

\ifCLASSOPTIONcompsoc
\newcommand{\fixmea}[1]{\textcolor{red}{AS: #1}}

\IEEEraisesectionheading{\section{Introduction}\label{sec:introduction}}
\else
\section{Introduction}
\label{sec:introduction}
\fi
\IEEEPARstart{A}~utonomous robots, and other intelligent mobile agents need to operate and navigate in dynamic human environments such as homes, offices, universities and streets. In these uncontrolled and human-populated environments, egocentric perceptual solutions allow robots to perceive, comprehend, and reason about the surroundings, \ie, to extract the information required to achieve their tasks safely from an often multimodal sensor stream. In recent years, the computer vision and robotics communities have proposed several standarized benchmarks to evaluate and compare different perception solutions, many of them from an egocentric visual perspective. The tasks evaluated include, on the one hand, scene understanding problems such as object detection~\cite{Everingham:2012:VOC, coco}, semantic and instance segmentation~\cite{Everingham:2012:VOC,coco,cordts2016cityscapes}, 3D reconstruction~\cite{Seitz:2006:ACE}, optical flow computation~\cite{Baker:2011:MID}, and stereo estimation~\cite{Scharstein:2002:TED}, and, on the other hand, problems related to understanding, analyzing and predicting human motion and behaviour such as pedestrian detection and tracking~\cite{Geiger:2012:AWR, MOTChallenge:arxiv:2015, MOTChallenge:arxiv:2016, dendorfer2019cvpr19, dendorfer2020mot20}, human pose estimation~\cite{andriluka2018posetrack} and human activity understanding~\cite{caba2015activitynet}. These benchmarks have enabled fair comparison between different perceptual solutions and have supported progress with new solutions in the respective research fields, often serving as training data to develop modern data-driven approaches.

\begin{figure}[th]
    \centering
    \includegraphics[width=0.45\textwidth]{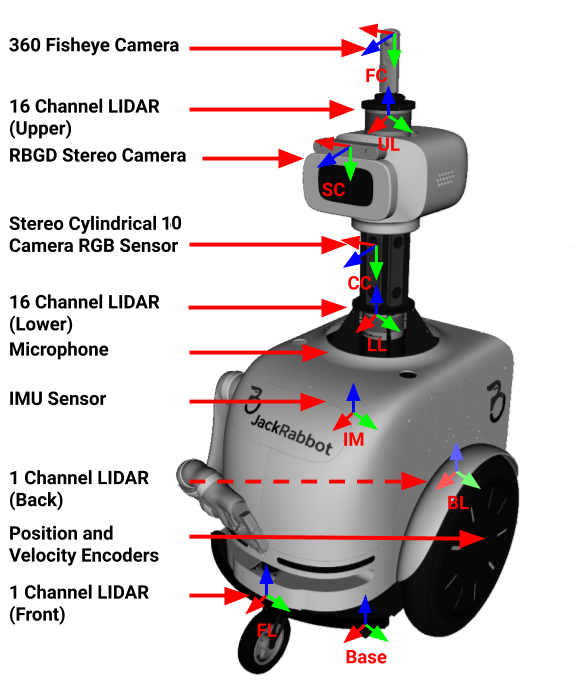}\vspace{-.5em}
    \caption{JackRabbot, our social robot and data collection platform, is equipped with four LiDAR sensors, three cameras, motion encoders, an IMU sensor, and a microphone.}
    \label{fig:JR}
\end{figure}

However, existing benchmarks mainly focus on visual perception tasks defined on single RGB images~\cite{Everingham:2012:VOC,coco,imagenet} or RGB video sequences~\cite{cordts2016cityscapes, andriluka2018posetrack, MOTChallenge:arxiv:2016, caba2015activitynet}. With the rise of popularity of 3D sensory data systems based on LiDAR, some benchmarks have begun to provide both 2D and 3D sensor data, and to define new scene understanding tasks on this geometric information. Some examples of these multimodal datasets are KITTI~\cite{Geiger:2012:AWR}, Oxford's Robotic Car~\cite{RobotCarDatasetIJRR},  Apolloscape~\cite{ ma2019trafficpredict}, nuScenes~\cite{caesar2019nuscenes} and Waymo~\cite{sun2019scalability}. Nonetheless, their targeted domain application is autonomous driving: the data they provide is captured exclusively from sensor suites on top of cars and the data only depicts streets, roads and highways. 

In this paper, we target a unique visual domain tailored to the egocentric perceptual tasks related to navigation in {\bf human environments}, both {\bf indoors} and {\bf outdoors}. We hope that this new domain provides the community an opportunity to develop egocentric visual perception frameworks for various types of autonomous navigation agents, not only self-driving cars but also autonomous agents such as social mobile robots\footnote{In 2030 up to 20\% of the jobs in retail (including malls, restocking, last-mile delivery, guidance, \etc.) could be carried out by social robots navigation among humans~\cite{manyika2017jobs}}. These agents need to perceive and understand the context of both indoor and outdoor scenes from a first-person view perspective in order to interact successfully with humans, predict their behaviour in these environments, and incorporate this behavior in agent's planning and decision processes.

With this motivation, we present the \textbf{JackRabbot Dataset and Benchmark (JRDB)}, a novel dataset and several visual benchmarks associated to it. Our dataset contains {\bf 64} minutes of sensor data acquired from our mobile robot JackRabbot\footnote{\url{http://svl.stanford.edu/projects/jackrabbot/}} comprising {\bf 54} sequences indoors and outdoors in a university campus environment. \revised{Capturing data from university campus, both indoors and outdoors and in crowded scenes, makes our dataset unique and tailored to perceptual tasks related to navigation in human environments, and different from previously collected datasets used for autonomous driving scenarios (\eg KITTI ~\cite{Geiger:2012:AWR}).We hope this
dataset will support and drive research in a variety of domains related to social robotics.} To this end, we have used multiple visual, depth, and motion sensors to collect this dataset. The sensor data includes stereo RGB 360\degree cylindrical video streams, continuous 3D point clouds from two LiDAR sensors, audio and GPS sensing. Currently, for this first phase, we provide the following ground truth annotations for our captured data\footnote{In near future, we have plans to augment with other types of annotations for many other visual perception tasks, \eg 2D human skeleton pose and individual, group and social activity labels.}: a)~2D bounding boxes in the images from each physical camera and the resulting cylindrical composed image for human/pedestrian class, b) occlusion attributes for each 2D bounding box, c) 3D oriented bounding boxes in the point clouds from the LiDAR sensors for human/pedestrian class, d) the association between 2D and 3D bounding boxes for corresponding humans/pedestrians, and e)~time consistent trajectories (tracks) for all annotated human/pedestrian in both 2D and 3D data. The labels have been annotated manually at high frame-rate providing accurate and sharp 2D and 3D trajectories. Based on this annotations, we provide a unified and standardized benchmark for 2D-3D person detection and tracking. Our 2D-3D person detection and tracking benchmark supports the evaluation of computer vision algorithms for object detection, 3D orientation estimation, and multi-object tracking. 

\revised{
With the novel JRDB dataset we aim at filling a need of the robotics and computer vision communities in terms of large scale multimodal annotated indoor and outdoor egocentric data. We hope the data in JRDB and associated benchmarks will foster research and progress in first-person view perception for autonomous agents in human environments. In the rest of this manuscript, we first describe the sensor setup used to acquire the multimodal signals in our dataset (Sec.~\ref{JRsensor}), the data captured and contained in the JRDB (Sec.~\ref{data}), and the ground truth annotations included in JRDB (Sec.~\ref{annotation2d}). Then, we compare to other existing datasets (Sec.~\ref{comparison}). We subsequently introduce the benchmarks and metrics for evaluation on JRDB (Sec.~\ref{benchmark}) and provide our own evaluation of state-of-the-art detection and tracking algorithms on the challenging data of JRDB (Sec.~\ref{evaluation}), followed by some conclusions.}

\section{Sensor Setup: The JackRabbot Platform}
\label{JRsensor}

To collect our multimodal dataset we used the sensor suite on-board of our mobile-manipulator JackRabbot (see Fig.~\ref{fig:JR}). JackRabbot (JR) is a custom-design robot platform tailored to navigate and interact in human environments. For navigation, JR is equipped with a Segway base with two active wheels and two passive caster wheels that improve stability. For manipulation and physical interaction, the robot is equipped with a six degrees-of-freedom Kinova Mico arm with two-fingered end-effector. The head of the robot is equipped with a LCD display that renders multiple facial expressions that allow JR to engage with humans and communicate intentions. The display and the RGB-D camera mounted on it are actively controlled by two Maxon motors providing two degrees-of-freedom pan-tilt motion.

The JackRabbot platform is equipped with multiple visual, audio, depth and motion sensors, as shown in Fig.~\ref{fig:JR}. In the following we list the most relevant sensor characteristics. 
\begin{itemize}
    \item 2 $\times$  Velodyne 16 Puck LITE rotating 3D laser scanner at \SI{10}{\hertz}, with
16 beams, 0.09\degree angular resolution, \SI{2}{\centi\meter} distance
accuracy, collecting approx. 1.3 million points/second, with a field of
view of 360\degree horizontally and 26.8\degree vertically, and a measuring range between \SI{2}{\meter} and \SI{100}{\meter}.
    \item 2 $\times$ SICK LMS500-20000 rotating 3D laser scanner at \SI{25}{\hertz}, with
a single beam, 0,167\degree angular resolution, \SI{3}{\centi\meter} distance
accuracy with a field of
view of 190\degree horizontally, and a measuring range between \SI{0.1}{\meter} and \SI{80}{\meter}.
    \item Cylindrical stereo sensor suite by Occam composed of two rows of five RGB cameras each. Each individual camera has a resolution of $752 \times 480$. The images of each row is composed into a cylindrical 360\degree panorama image.
    \item 360\degree fisheye RGB camera by Ricoh (Theta model) with a resolution of $1280 \times 720$ at 15 fps.
    \item RGB-D stereo camera by ZED with a resolution of $672 \times 376$ at 15 fps.
    \item Digital microphone.
    \item 6D inertial measurement unit (IMU).
    \item Position encoders on each of the active wheels of the base, the joints and fingers of the arm, and the motors of the pan-tilt unit of the head.
\end{itemize}

While our dataset contains information from all the aforementioned sensors, we only annotated the point clouds from the two Velodyne 16 LiDARs and the images from one of the rows of the 360\degree stereo cylindrical camera, in this first phase of the labelling effort.

JackRabbot is also equipped with an on-board computational unit including two Nvidia GPUs. The onboard computer runs Linux 16.04 and ROS (Robot Operating System) Kinetic that we use to collect multimodal synchronized signals in the form of rosbags.

\section{Captured Data}
\label{data}

\begin{figure*}[t]
\centering
\begin{subfigure}[b]{0.49\textwidth}
    \centering
	\includegraphics[width=.9\linewidth]{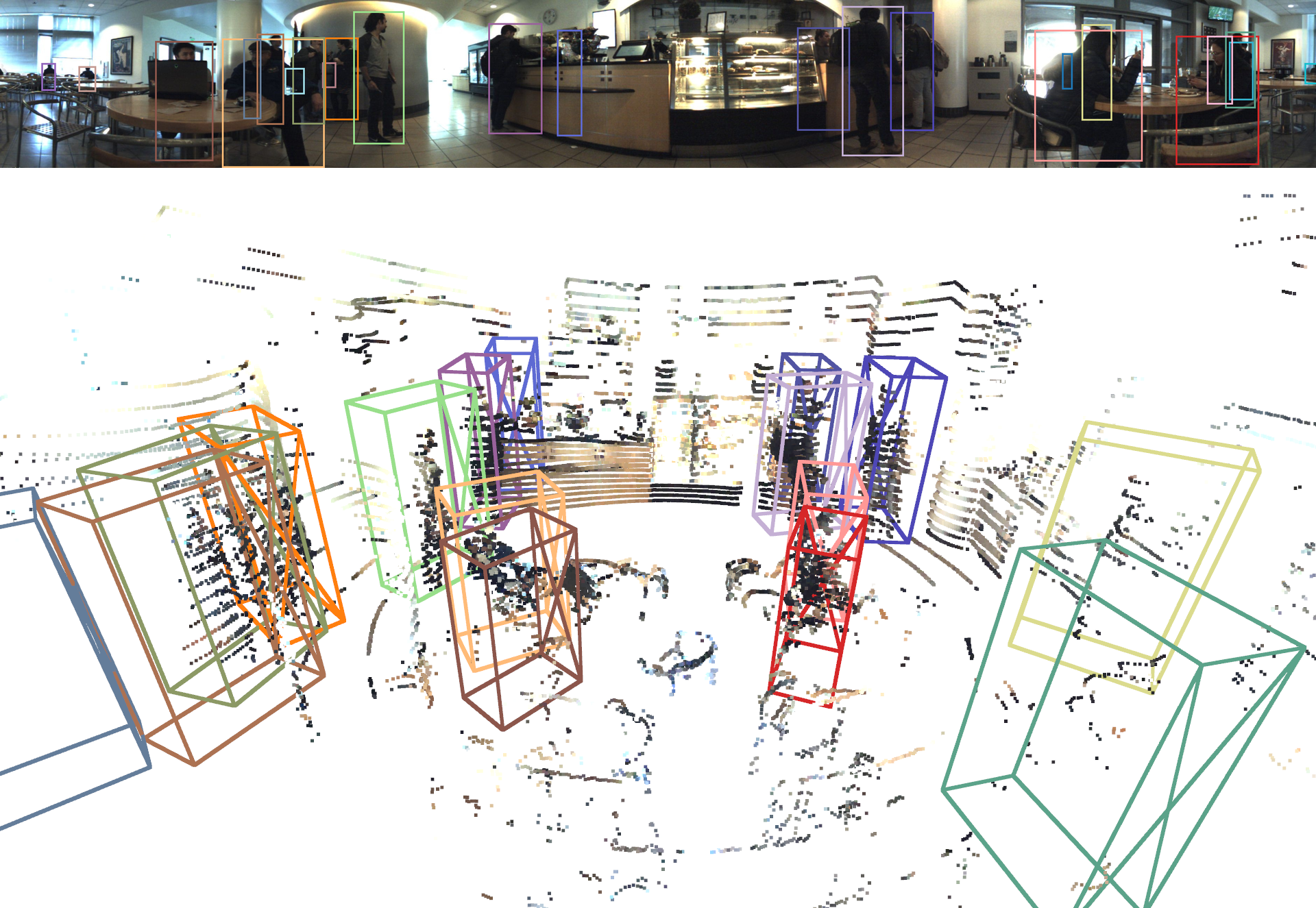}
	\caption{Stationary robot, indoor scene (\texttt{\scriptsize bytes-cafe-2019-02-07\_0})} 
\end{subfigure}
\hfill
\begin{subfigure}[b]{0.49\textwidth}
    \centering
	\includegraphics[width=.9\linewidth]{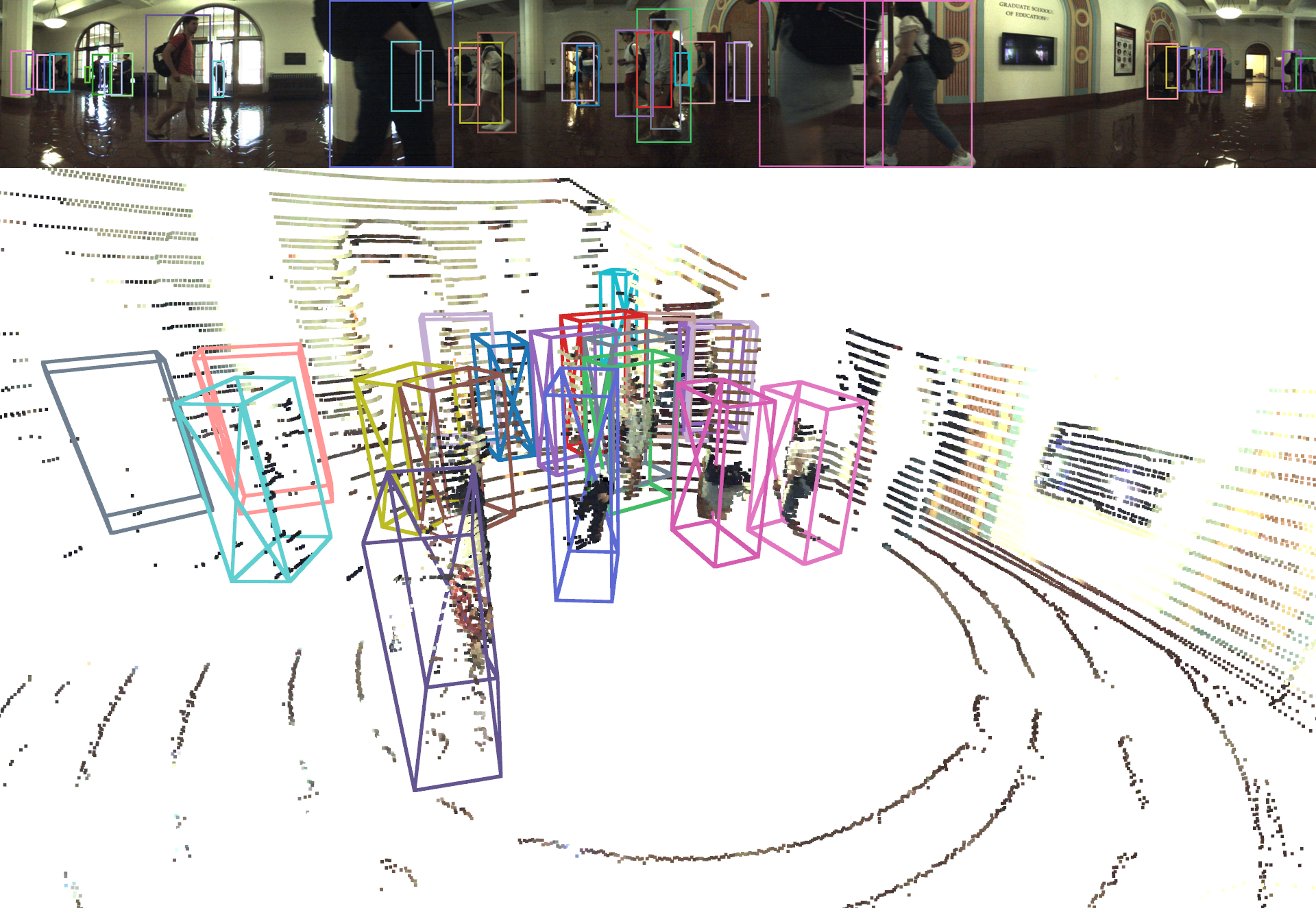}
	\caption{Moving robot, indoor scene (\texttt{\scriptsize cubberly-auditorium-2019-04-22\_0})} 
\end{subfigure}
\vskip\baselineskip
\begin{subfigure}[b]{0.49\textwidth}
    \centering
	\includegraphics[width=.9\linewidth]{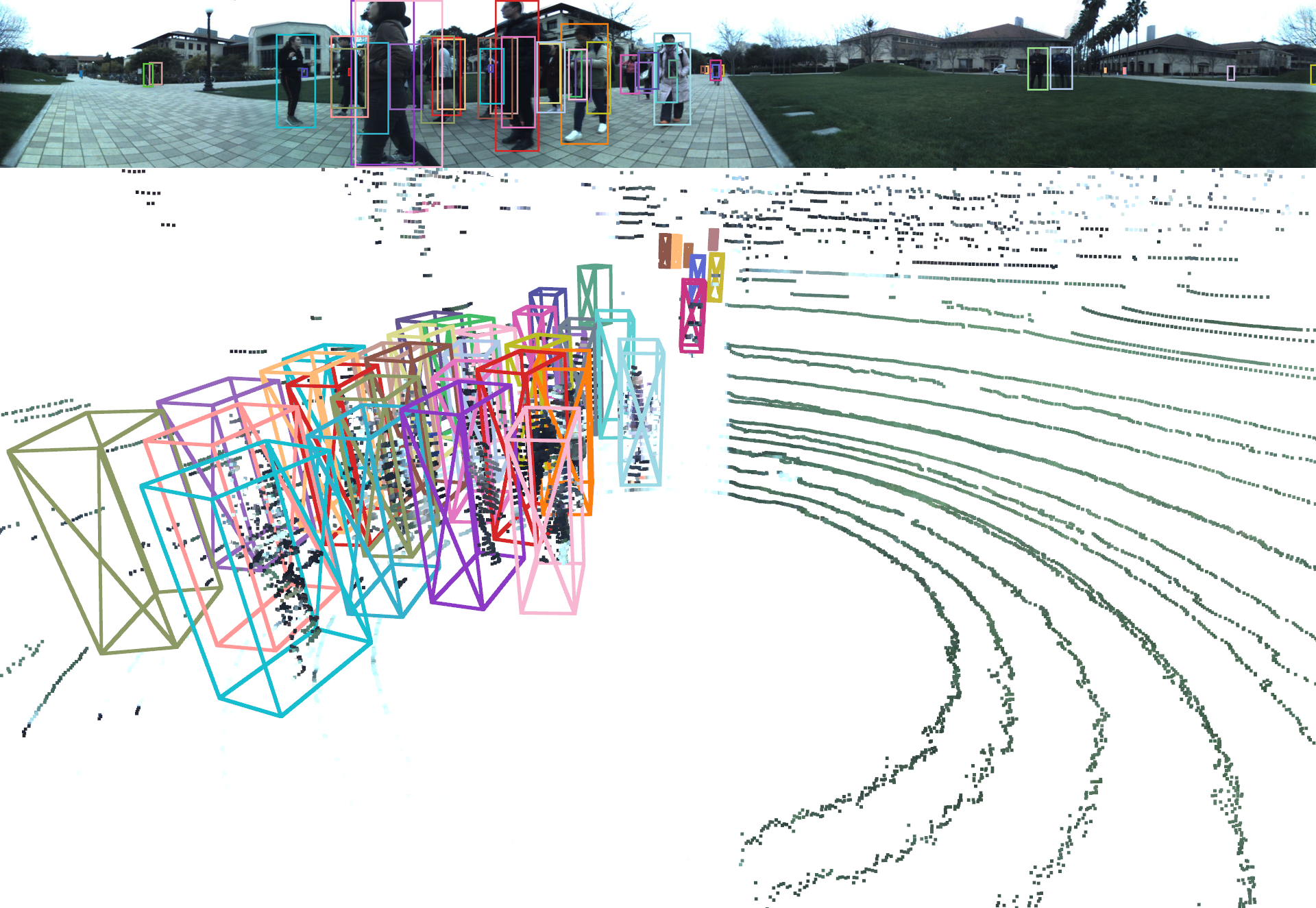}
	\caption{Stationary robot, outdoor scene (\texttt{\scriptsize huang-lane-2019-02-12\_0})} 
\end{subfigure}
\hfill
\begin{subfigure}[b]{0.49\textwidth}
    \centering
	\includegraphics[width=.9\linewidth]{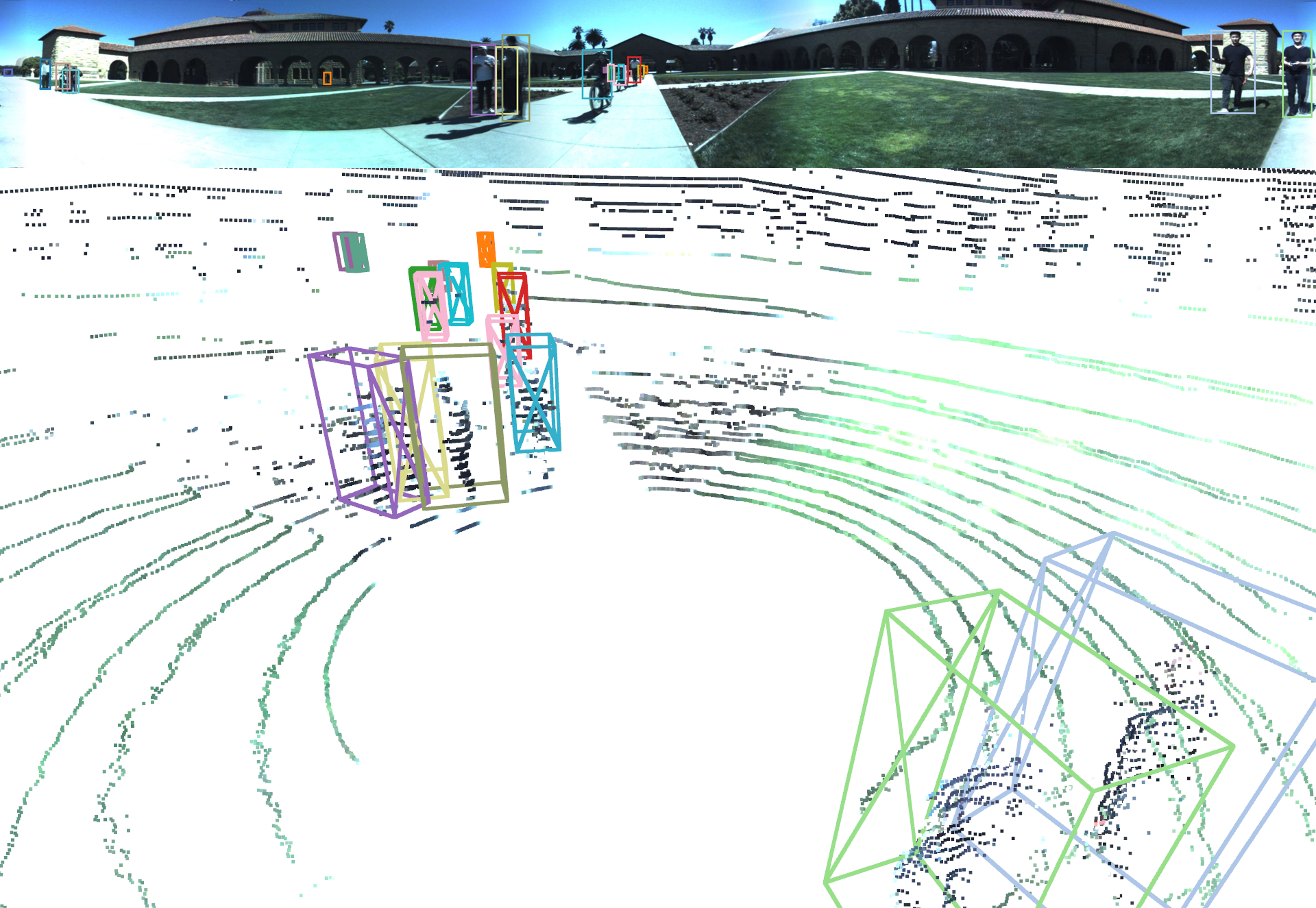}
	\caption{Moving robot, outdoor scene (\texttt{\scriptsize memorial-court-2019-03-16\_0})} 
\end{subfigure}

 	\caption{Sample visualization of the dataset (best seen in color). For each subfigure, top: 2D stitched 360\degree panoarama with human-annotated 2D bounding boxes, bottom: 3D Velodyne point clouds with human-annotated 3D oriented bounding boxes. In order to demonstrate the accuracy of camera registration, we visualize 3D point clouds with color extracted from the projected 2D RGB stitched 360 panorama image. Our dataset captures a variety of pedestrian density, indoor/outdoor scenes, and moving/stationary robot. Additional visualizations for all scenes in Fig.~A.3 in the Appendix.}
 	\label{fig:visualization}
 \end{figure*}
 
To generate the JRDB, we collected time synchronized data from all sensors and save it on the on-board computer. The data is recorded from 30 different locations indoors and outdoors \revised{resulting in 33 indoor and 21 outdoor sequences} in a university campus environment.  \revised{There are 32 sequences captured from the moving robot, and 22 from a stationary position}, with varying and uncontrolled environmental conditions such as illumination and other natural and dynamical elements. \revised{For repeating locations, the data is acquired from different points of view, \eg the foyer and a corridor in the same building, or stationary vs. moving setups, in the same time of the day. The environments in which the data is collected includes university building halls and corridors, laboratories and exits of large classrooms, pedestrian areas, plazas, food malls, sidewalks, pedestrian crossing areas, indoor patios and cafeterias (for a complete overview, see Fig. A.3)}. We also ensure that the recorded data captures a variation of natural human posture, behavior and social activities in different crowd densities in the indoor and outdoor environments.

The JackRabbot dataset includes $54$ sensor sequences captured from the multimodal robot’s sensor stream containing:
\begin{itemize}
    \item video streams at 15 fps from stereo cylindrical 360\degree RGB cameras (raw data acquired from two rows of five RGB cameras each),
    \item continuous 3D point clouds from two Velodyne LiDAR signals of 16 planar rays,
    \item RGB-D video stream at 30 fps,
    \item 360\degree spherical image from a fisheye camera,
    \item line 3D point clouds from two Sick LiDARs, 
    \item audio signal,
    \item encoder values from the robot's wheels, joint arm and motors of the head.
\end{itemize}

\section{Annotation and Labels in JRDB}
\label{annotations}

Our JRDB dataset includes annotations for all pedestrians in 2D stereo images and 3D point clouds with 2D and 3D bounding boxes, respectively. We also label all persons in the scene with a unique ID that is consistent across 2D and 3D annotations both within the same sequence, and along the entire sequence. Such 2D-3D time consistent annotations allow us to train and evaluate multiple solutions as we present in Sec.~\ref{benchmark}. In the following, we will discuss the details of the annotation such as the annotation procedure, bounding box parameterization and additional metadata.


\subsection{Annotation Procedure}
\label{annotationprocedure}

\revised{The annotation is performed by a group of labelers specially instructed for the task. Due to the rapidly changing and irregular motion patterns of pedestrians in indoor environments, we annotate at \SI{7.5}{\hertz}, once for every two received LiDAR point clouds and set of synchronized images from the 2D RGB stereo cylindrical sensor suite. The annotation procedure is as follows: First, 3D bounding boxes around each pedestrian in the point cloud are annotated. Each pedestrian is annotated manually in the combined point cloud resulting from merging the signals from both Velodyne 16 LiDAR sensors. The bounding boxes are annotated with the \textit{assumed} full pedestrian size, if they are not visible. The boxes have a constant height, width and length per pedestrian that fits from shoulder to shoulder and has enough front and back space to cover the arm swinging during walking. Since the point clouds are sparse, especially at longer distances, the 2D RGB images associated to the scene are shown during annotation to support the labelers. 
\revisedF{We annotate a pedestrian in all frames if it is visible in at least one frame, \eg at least 50 points of the point cloud belong to the pedestrian.}
The annotations are performed per sequence such that the same ID labels are assigned to the same pedestrians over the entire sequence.}

\revised{After the annotation in 3D, the bounding boxes are projected into the image plane of each of the five cameras of the top row of the stereo cylindrical camera suite (see Appendix~A.1 for the details of the transformation). The projection provides a first approximation for the 2D bounding boxes of the pedestrians visible in both the point cloud and the images. The projections are then manually refined to fit accurately the pedestrian size. The 2D bounding boxes are annotated with the \textit{assumed} full pedestrian size, if they are not visible, and should enclose all the human body, including legs and arms. Additional 2D bounding boxes are added to annotate pedestrians visible in the 2D RGB images but not in the point cloud. After this annotation step we obtain temporally consistent labels across the point cloud as well as the images of the five cameras.}

\revised{To provide annotations for researchers interested on methods based on cylindrical panorama images, we merge the 2D annotations in the images of the five cameras of the top row of the cylindrical stereo sensor suite into a unified cylindrical image. We use the same mapping applied to compose the cylindrical images (see Appendix~A.1) to obtain the coordinates of four corners of each 2D bounding box in the 360\degree cylindrical image. Using the unique ID per pedestrian, we compute the tightest enclosing axis-aligned rectangle for the corners of each pedestrian. In few cases, since our cylindrical image spans the entire 360\degree, the same pedestrian appears in both the right and left borders of the image. Those cases are indicated with the $(u,v)$ coordinates of the top left corner of the box and a width that extends beyond the right edge of the image. The resulting of merging the annotations in the 360~\degree images can be observed in Fig~\ref{fig:visualization}.} \revised{As a final step, we upsample the annotations from \SI{7.5}{\hertz} to \SI{15}{\hertz} with a linear interpolation.}


\subsection{3D Bounding Box Labels}
\label{annotation3d}
For each pedestrian visible in the point cloud, we annotate a 3D bounding box alongside with the following metadata:

\begin{itemize}
    \item \texttt{label\_id}: Unique ID of the pedestrian that is consistent across 2D and 3D annotation and within the same sequence
    \item \texttt{box}: We assume axis aligned 3D bounding boxes. The 3D bounding box is parameterized by the location of the center of the cuboid, $(c_x, c_y, c_z)$, the size of the bounding box, $(h, w, l)$, and the rotation angle around the gravity ($z$) axis, $r_z$.
    \item \texttt{interpolated}: Flag indicating whether the labels in this frame are a direct product of the human annotation (at \SI{7.5}{\hertz}) or generated by the interpolation of human annotated frames (interpolation to \SI{15}{\hertz}).
    \item \texttt{num\_points}: Number of points of the point cloud within the 3D bounding box.
    \item \texttt{distance}: Distance from the sensor to the pedestrian.
    \item \texttt{no\_eval}: Flag that indicates if the annotation does not fulfill all the conditions to be included in the evaluation of a detector/tracker as described in Sec.~\ref{benchmark}.
    \item \texttt{observation\_angle}: Angle with respect to the front facing direction of the robot. We consider the robot to be pointing in the $x$ direction. Therefore, the observation angle for a bounding box with center at $(c_x, c_y, c_z)$ is given by $\tan^{-1}\left(\frac{-c_y}{c_x}\right)$.
\end{itemize}

\subsection{2D Bounding Box Labels}
\label{annotation2d}
For each pedestrian visible the 2D RGB images from the five cameras of the top row of the stereo cylindrical sensor suite, we annotate a 2D bounding box alongside with the following metadata:

\begin{itemize}
    \item \texttt{label\_id}: Unique ID of the person that is consistent across 2D and 3D annotation within the same sequence, and over time for the entire sequence. The IDs are consistent between the five annotated camera images.
    \item \texttt{box}: 2D bounding box parameterized by the location of the upper left corner of the bounding box, $(u, v)$, and the size of the 2D bounding box, $(w, h)$.
    \item \texttt{truncated}: Flag indicating whether the object is leaving the upper or lower border of the image boundary. 
    \item \texttt{occlusion}: Level of occlusion of the pedestrian in the image. We consider the following four levels: \texttt{Fully\_visible} (no occlusion), \texttt{Mostly\_visible} (more than 50\% visible), \texttt{Severely\_occluded} (more than 50\% occluded), \texttt{Fully\_occluded} (not visible at all).
    \item \texttt{interpolated}: lag indicating whether the labels in this frame are a direct product of the human annotation (at \SI{7.5}{\hertz}) or generated by the interpolation of human annotated frames (interpolation to \SI{15}{\hertz}).
    \item \texttt{no\_eval}: Flag indicating whether the annotation does not fulfill all the conditions to be included in the benchmark evaluation for detectors/trackers as described in Sec.~\ref{benchmark}.
    \item \texttt{area}: Area of the bounding box in pixel$^2$.
\end{itemize}

\subsection{Data Split and Statistics}
\label{stats}

\begin{figure}[t]
    \centering
   \begin{subfigure}[b]{0.48\linewidth}
    \centering
	\includegraphics[width=1.05\linewidth]{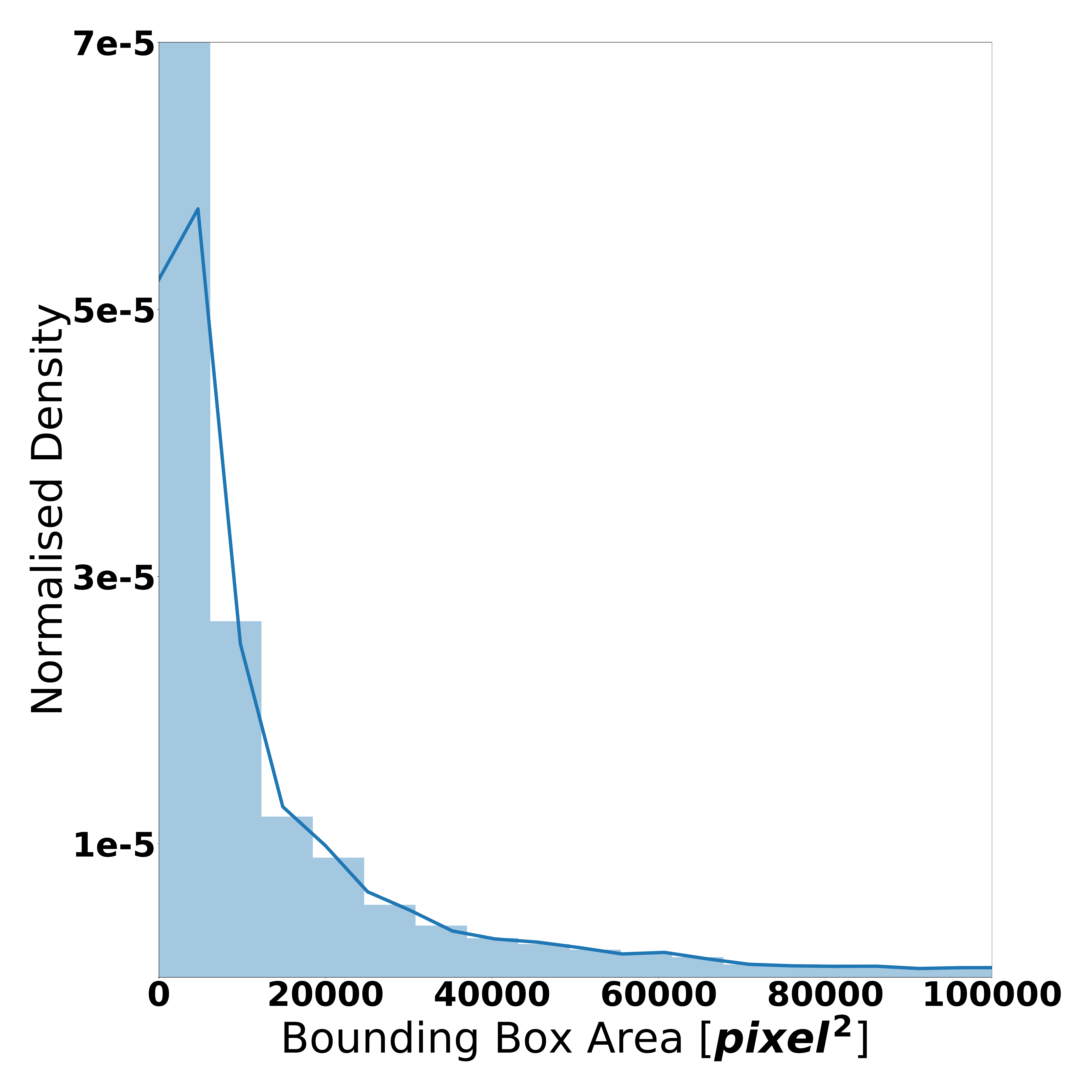}
	\caption{Train\revised{-Validation} Set} 
\end{subfigure}
\hfill
\begin{subfigure}[b]{0.48\linewidth}
    \centering
	\includegraphics[width=1.05\linewidth]{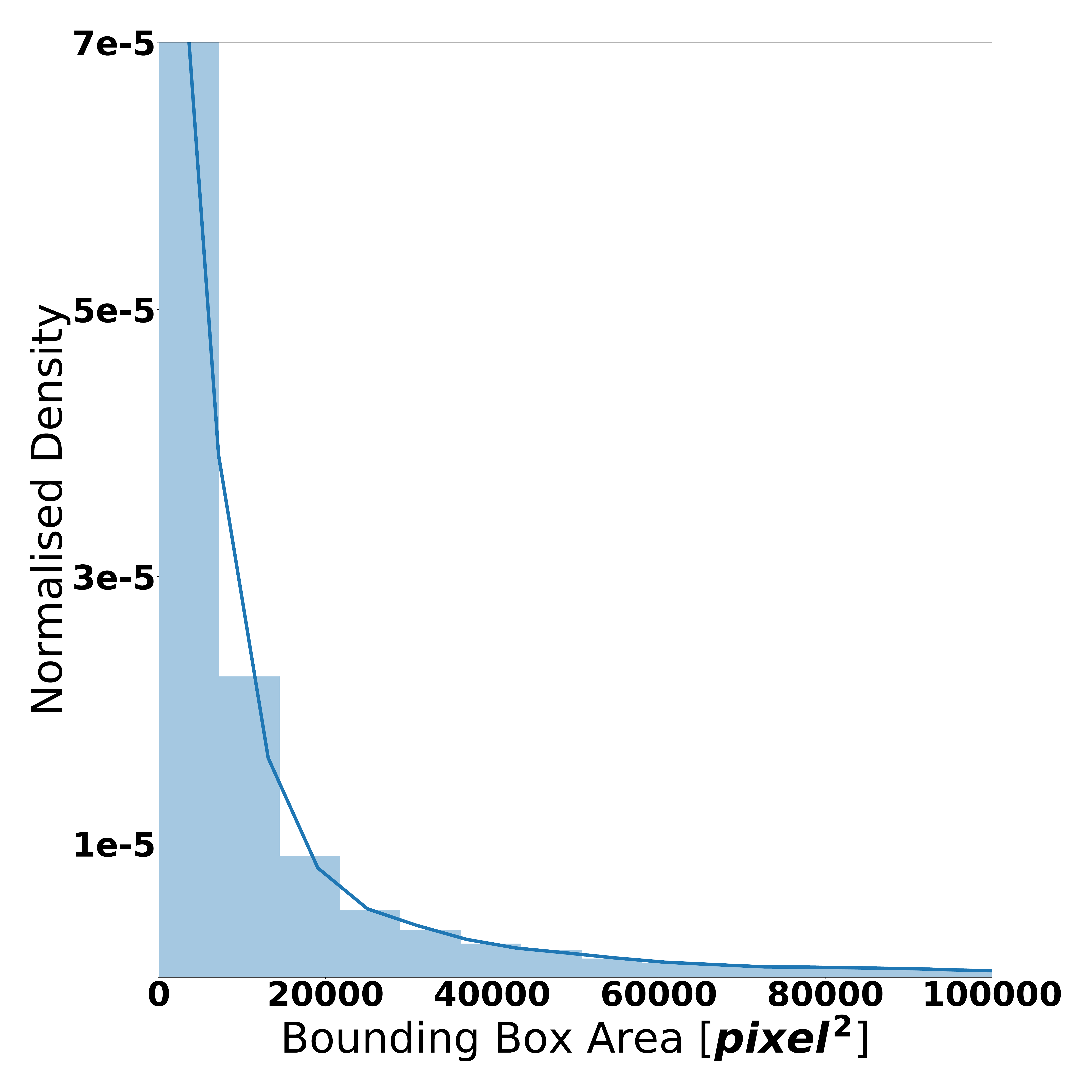}
	\caption{Test Set} 
\end{subfigure}
    \caption{Histogram of areas of annotated 2D bounding boxes along with the Kernel Density Estimate. Train\revised{-validation} and test sets show similar distributions \revised{(comparison between train and validation in Appendix, Fig.~A.2)}. Our dataset contains a large number of frames in indoor scenes where people are closer to the robot, resulting in boxes with areas over 10000 pixel$^2$. However, most boxes present an area of less than 10000 pixel$^2$ as most annotated pedestrians are at \SI{5}{\meter} or further.}
    \label{fig:AreaDist}
\end{figure}

\begin{figure}[t]
\centering
\begin{subfigure}[b]{0.48\linewidth}
    \centering
	\includegraphics[width=1.05\linewidth]{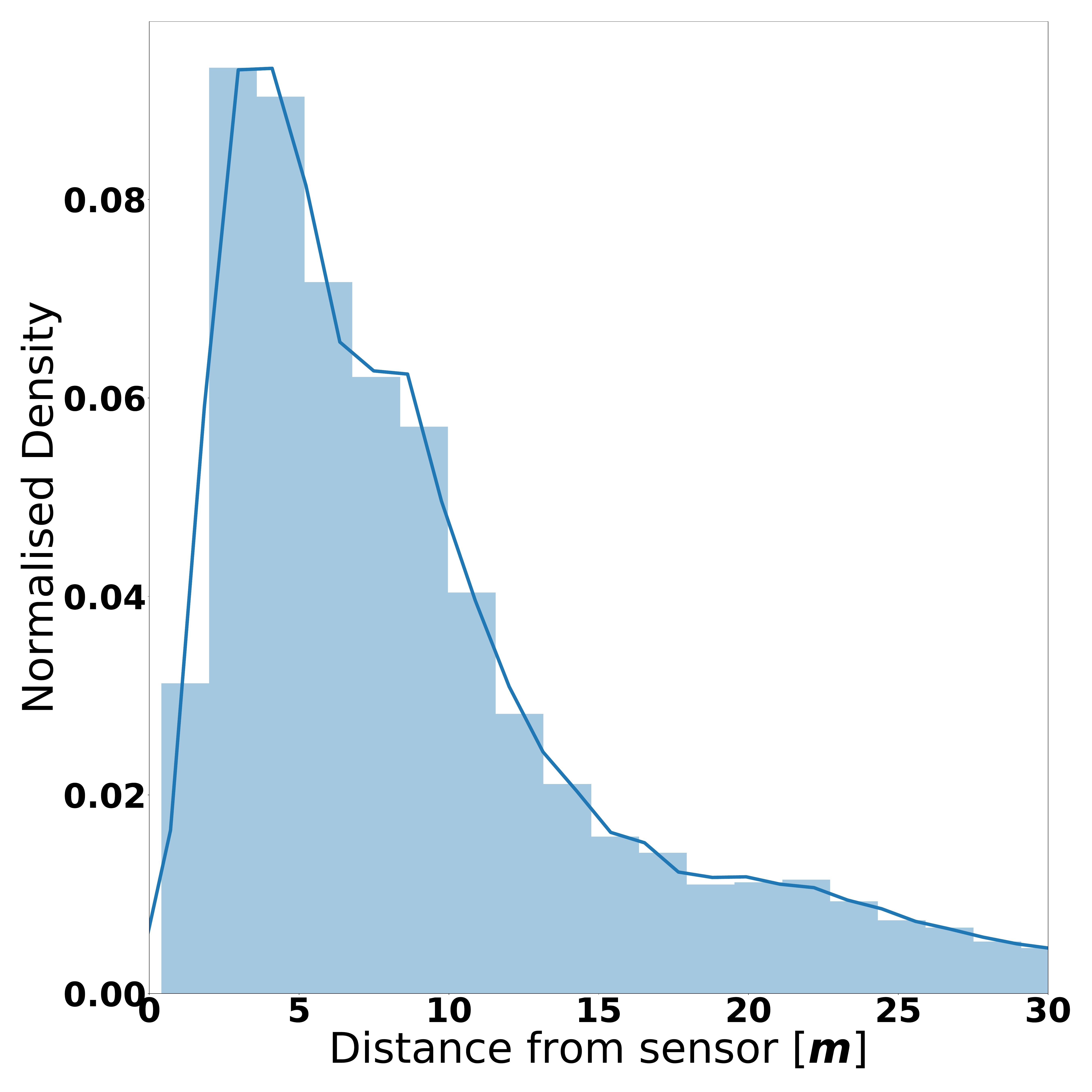}
	\caption{Train\revised{-Validation} Set} 
\end{subfigure}
\hfill
\begin{subfigure}[b]{0.48\linewidth}
    \centering
	\includegraphics[width=1.05\linewidth]{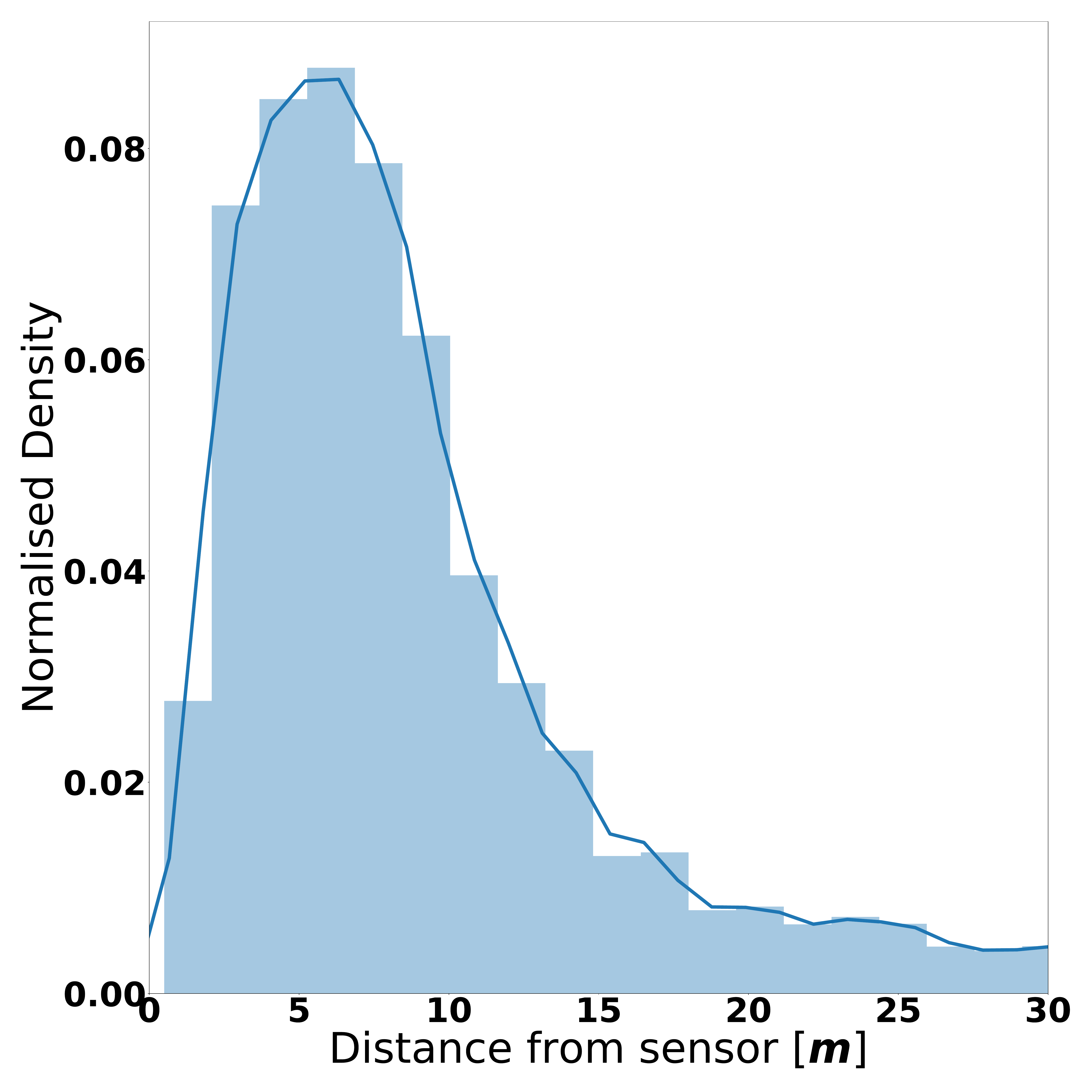}
	\caption{Test Set} 
\end{subfigure}
 	\caption{Histogram of distances of 3D bounding boxes to the sensor location. The distribution remains relatively unchanged between train\revised{-validation} and test sets \revised{(comparison between train and validation in Appendix, Fig.~A.2)}. Our dataset contains a large number of frames of indoor scenes where people tend to be closer to the robot, yielding a distribution with a mode at approximately 5m, which is quite different to existing datasets for autonomous cars. The long-tailed nature of the distribution is due to outdoor scenes, with some annotations as far as 80m from JackRabbot.}
 	\label{fig:3D_distance_stats}
 \end{figure}

\begin{figure}[tbh]
\centering
\begin{subfigure}[b]{0.48\linewidth}
    \centering
	\includegraphics[width=1.05\linewidth]{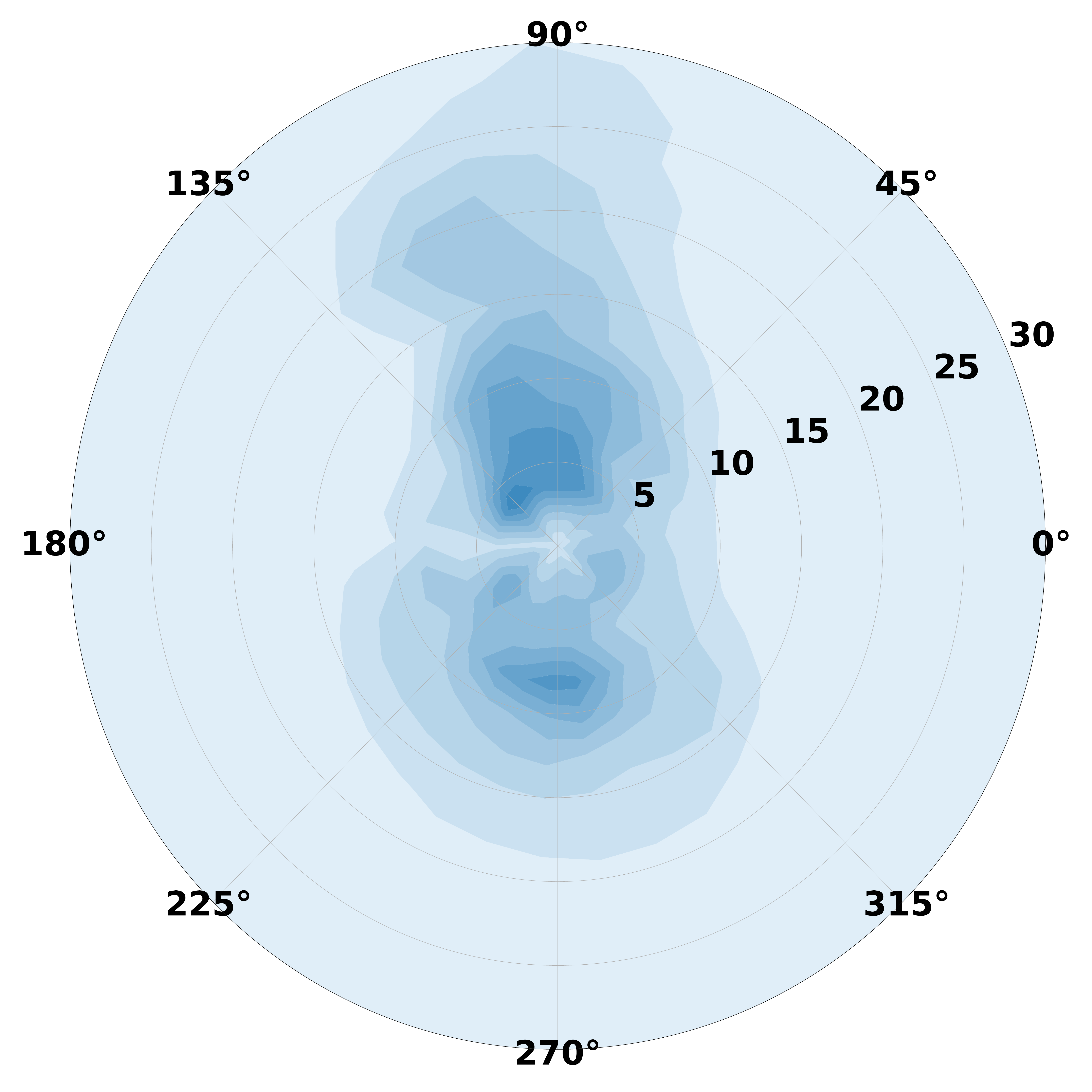}
	\caption{Train\revised{-Validation} Set} 
\end{subfigure}
\hfill
\begin{subfigure}[b]{0.48\linewidth}
    \centering
	\includegraphics[width=1.05\linewidth]{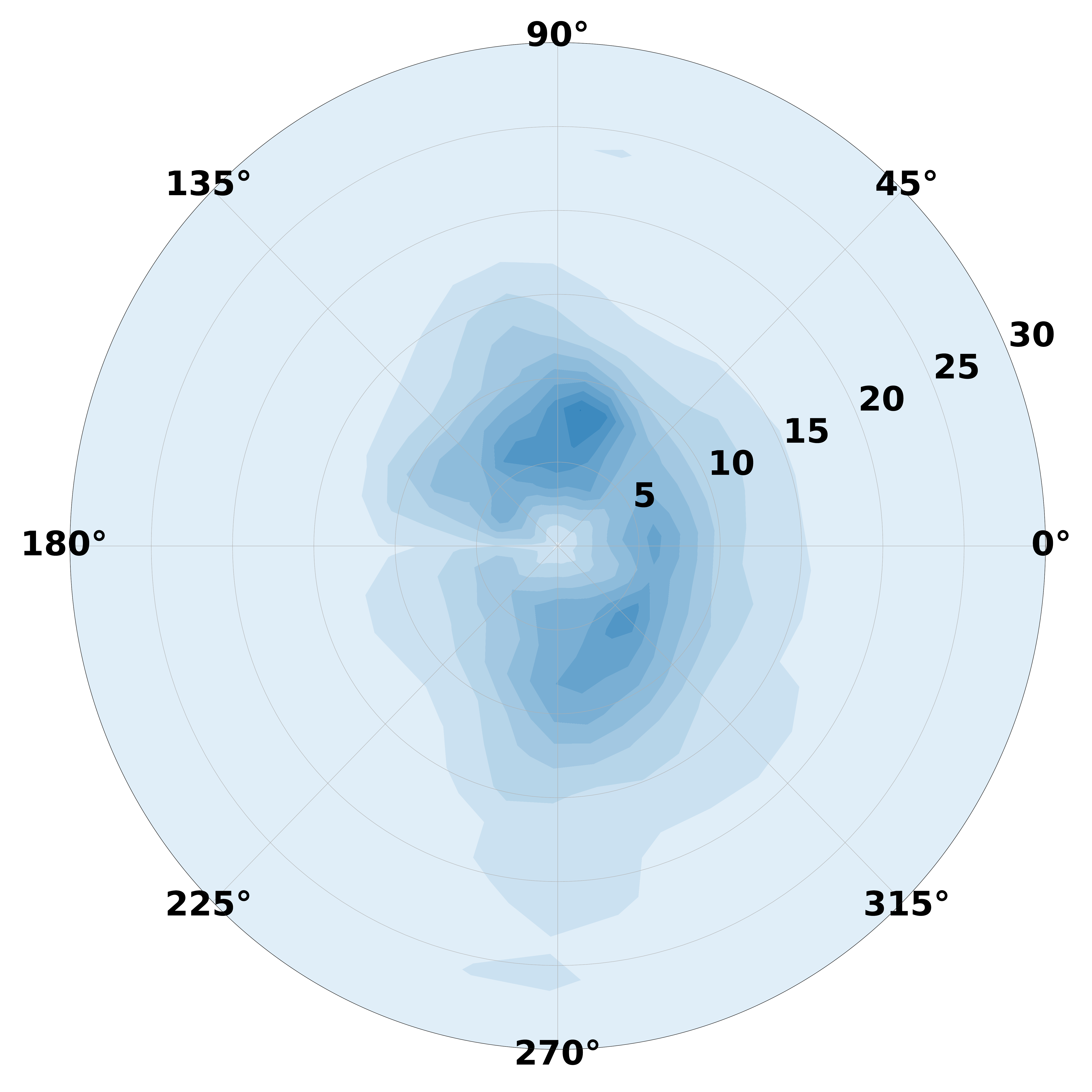}
	\caption{Test Set} 
\end{subfigure}
\hfill
 	\caption{Kernel Density Estimates (KDEs) of the spatial distribution of people around the robot, across the train\revised{-validation} and test splits of the dataset \revised{(comparison between train and validation in Appendix, Fig.~A.2)}. In close proximity of JackRabbot, people are evenly spread across all directions, but tend to be along the X-axis (front and back of the robot with respect to the base frame) as the distance increases. This is due to the robot being driven along straight walkways when outdoors, and along corridors when indoors. Distances are in meters.}
 	\label{fig:3D_distance_angle_stats}
 \end{figure}

\begin{figure}[tbh]
\centering
\begin{subfigure}[b]{0.48\linewidth}
    \centering
	\includegraphics[width=\linewidth]{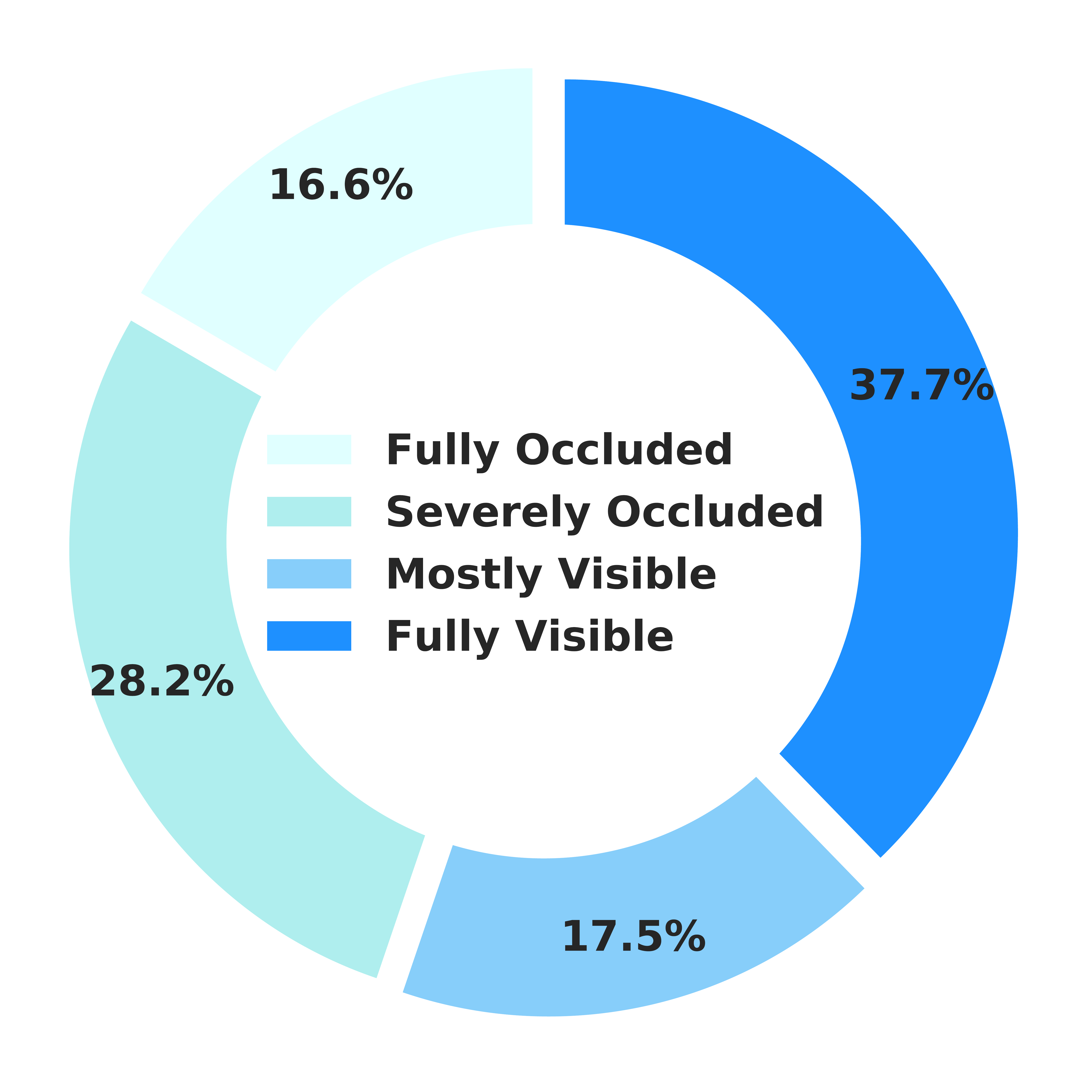}
	\caption{Train\revised{-Validation} Set} 
\end{subfigure}
\hfill
\begin{subfigure}[b]{0.48\linewidth}
    \centering
	\includegraphics[width=\linewidth]{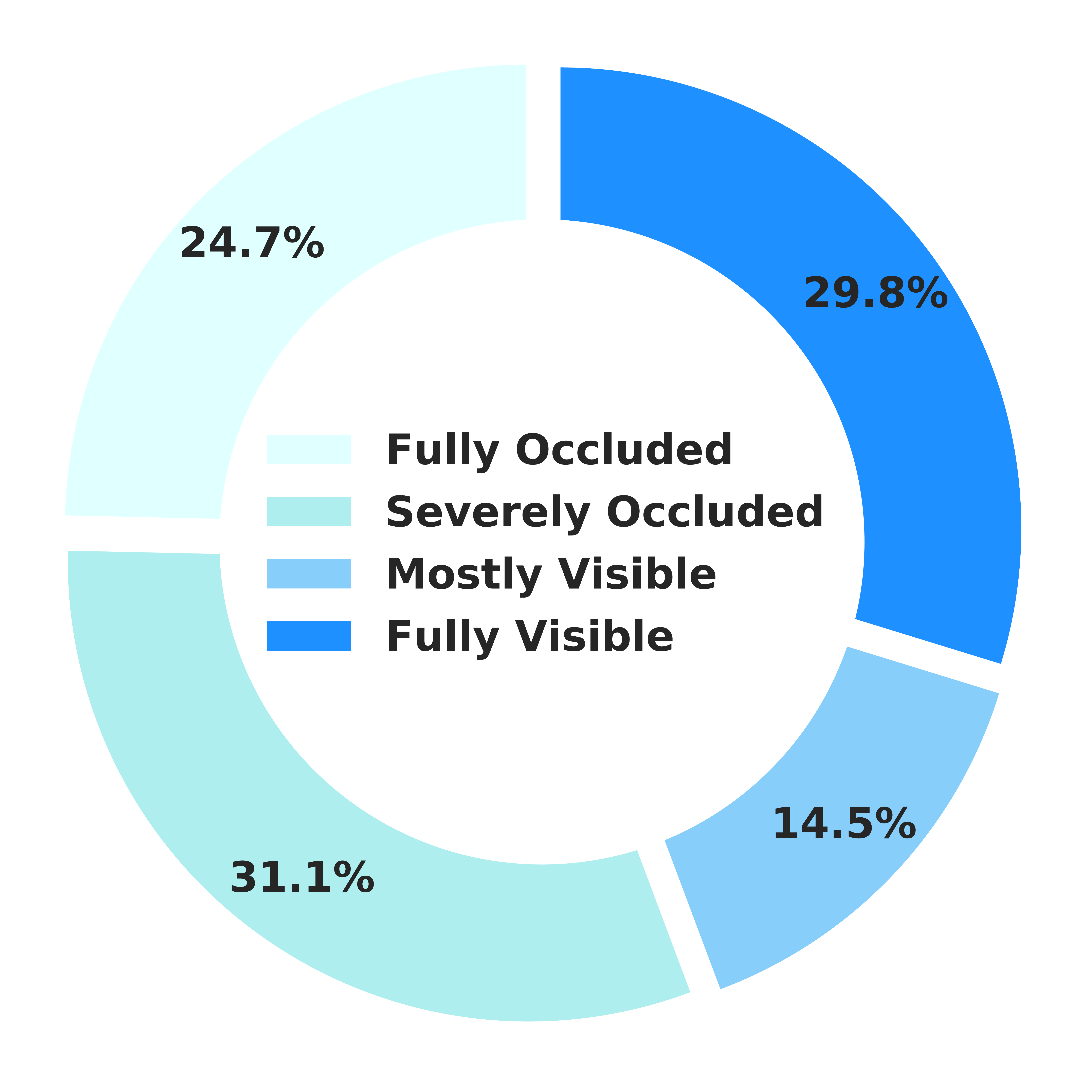}
	\caption{Test Set} 
\end{subfigure}
 	\caption{Distribution of occlusion levels (or lack thereof) in train\revised{-validation} and test sets \revised{(comparison between train and validation in Appendix, Fig.~A.2)}. JRDB considers four levels of occlusion: fully occluded, severely occluded, mostly occluded and fully visible. 
 	 Approximately two thirds of the total number of boxes in both the train and test sets are more than 50\% visible.
 	}
 	\label{fig:2D_occlusion_stats}
 \end{figure}

In order to provide a benchmark, we divide the 54 captured sequences into \revised{train-validation} and test splits equally. \revised{We select 7 out of 27 sequences (25\% of the split) as the suggested validation split.} We aim to have the same number of indoor and outdoor sequences captured using stationary or moving robot in both train-validation and test splits. Moreover, we ensure that the train-validation and test splits contain very diverse scenes while having comparable annotation statistics (See \Tab~\ref{tab:numerical_stats} and \Fig~\ref{fig:AreaDist}-\ref{fig:2D_occlusion_stats}). 

All relevant captured data, \ie point cloud streams, images from the individual cameras of the cylindrical stereo sensor suite as well as 360\degree panorama composed images and additional sensor signals for both train and test split, and annotations for the train set are publicly available to download at \url{https://jrdb.stanford.edu}. We exclude \SI{5.2}{\second} at the end of each sequence; this data is not released publicly in preparation for a future forecasting set of benchmarks.

\begin{table*}[tbh]
    \centering
\begin{tabular}{ |c| c c| c c|c|| c c| c c| c|}
\hline
&\multicolumn{5}{c||}{Train\revised{-Validation} Set}&\multicolumn{5}{c|}{Test Set}\\
\cline{2-6}\cline{7-11}
\backslashbox{Data Statistics}{Data Split}& Indoor & Outdoor & \revisedThree{Stationary} & \revisedThree{Moving}& Total & Indoor & Outdoor &\revisedThree{Stationary} & \revisedThree{Moving} & Total\\ 
   \hline\hline
 Number of sequences &17&10 & \revisedThree{11} & \revisedThree{16} & \bf 27&16&11& \revisedThree{11} & \revisedThree{16} &\bf 27\\ 
 Number of frames & 19248 & 8699 & \revisedThree{12248} & \revisedThree{15699} &\bf 27947& 17408 & 12359 & \revisedThree{12217} & \revisedThree{17550} & \bf 29767\\
 \hline\hline
 Number of 2D Boxes & 765K & 324K & \revisedThree{419K} & \revisedThree{670K} &\bf 1.09M & 868K & 405K & \revisedThree{587K} & \revisedThree{686K} &\bf 1.27M\\  
 Number of 2D tracks & 1229 & 564 & \revisedThree{622} & \revisedThree{1171} & \bf 1793 & 1201 & 617 & \revisedThree{785} & \revisedThree{1033} & \bf 1818\\
 Avg. track length in frames (2D)  & 475.7 & 457.8 & \revisedThree{506.3} & \revisedThree{450.9} & \bf 470.0 & 567.9 & 534.3 & \revisedThree{600.1} & \revisedThree{523.3} &\bf 556.5 \\
 \hline
 Number of 3D Boxes & 578K & 254K & \revisedThree{312K} & \revisedThree{520K} & \bf 832K & 671K & 324K & \revisedThree{466K} & \revisedThree{529K} & \bf 995K\\
 Number of 3D tracks  & 1220 & 554 & \revisedThree{612} & \revisedThree{1162} & \bf 1774 & 1186 & 611 & \revisedThree{780} & \revisedThree{1017} & \bf 1797\\
 Avg. track length in frames (3D)  & 473.7 & 459.3 & \revisedThree{509.8} & \revisedThree{447.8} & \bf 469.2 & 566.1 & 529.9 & \revisedThree{597.7} & \revisedThree{520.1} &\bf 553.8\\
\hline
\end{tabular}
    \caption{Statistical analysis of the data in JRDB. 2D statistics aggregate the five individual cameras used to construct the 360\degree cylindrical RGB images.}
    \label{tab:numerical_stats}
\end{table*}

\revised{\section{JRDB: A Dataset with Unique Features}}
\label{comparison}

\revised{\begin{table*}[tbh]
\centering
\begin{tabular}{ c c c c c c c c}
\hline
Name (year) & \# Sequences & Indoor/Outdoor & \# Frames & \# 2D BBx & \# 3D BBx & \# 2D Tracks & \# 3D Tracks \\ 
\hline\hline
MOT17 (2017) & 14 & Outdoor only & 34K & 290K & -- & 1.3K & --\\ 
MOT20 (2020) & 8 & Indoor \& Outdoor& 13K & 1.6M & -- & 3.6K & --\\ 
\hline
KITTI (2013) & 22 & Outdoor only & 15K & 6K & 6K & 900 & --\\ 
ApolloScape (2018) & 4 & Outdoor only & 93K & -- & 16.2K & -- & 646\\ 
BDD100K (2018) & 1600 & Outdoor only & 318K & 440K & -- & 24K & -- \\ 
NuScenes (2019) & 1000 & Outdoor only &  1.4M & -- & 232K & -- & 11.4K \\ 
Waymo (2019) & 1150 & Outdoor only & 600K & 2.1M & 2.8M & 45K & 23K \\
\hline
\textbf{JRDB (2020)} & \textbf{54} & \textbf{Indoor \& Outdoor} & \textbf{28K} & \textbf{2.4M} & \textbf{1.8M} & \textbf{3.5K} & \textbf{3.5K} \\
\hline
\end{tabular}
\caption{\revised{Comparing statistics between JRDB and similar existing datasets. Reported numbers are for annotations of pedestrians/humans in the tracking portion of these datasets. \# = Number, K = Thousand, M = Million, and BBx = Bounding Box.}}
\label{tab:dataset_comparison}
\end{table*}}

 \begin{figure*}[tbh]
\centering
\begin{subfigure}[b]{0.19\linewidth}
    \centering
	\includegraphics[width=1.05\linewidth]{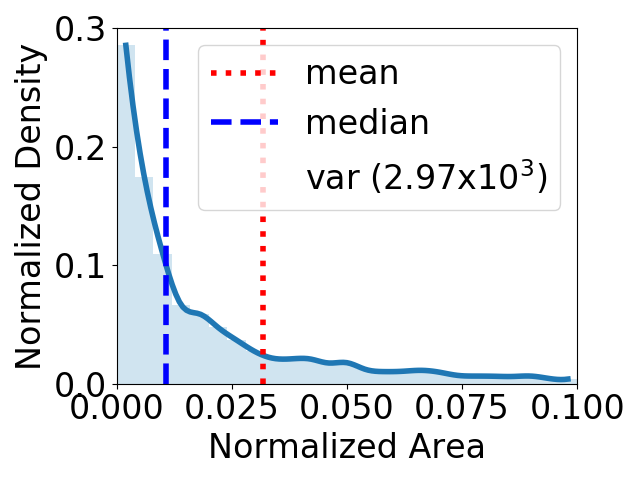}
\end{subfigure}
\hfill
\begin{subfigure}[b]{0.19\linewidth}
    \centering
	\includegraphics[width=1.05\linewidth]{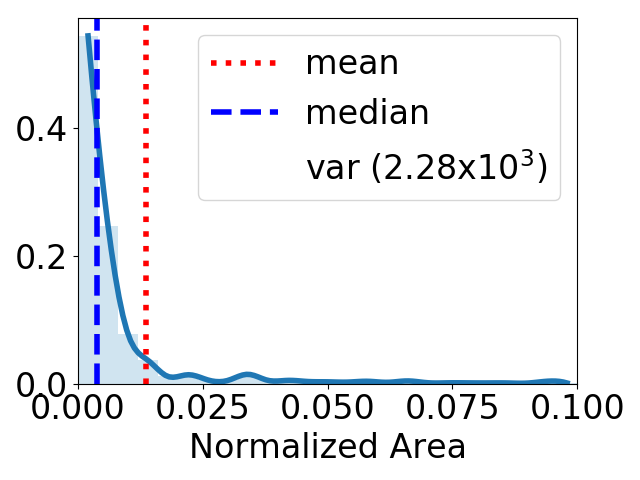}
\end{subfigure}
\hfill
\begin{subfigure}[b]{0.19\linewidth}
    \centering
	\includegraphics[width=1.05\linewidth]{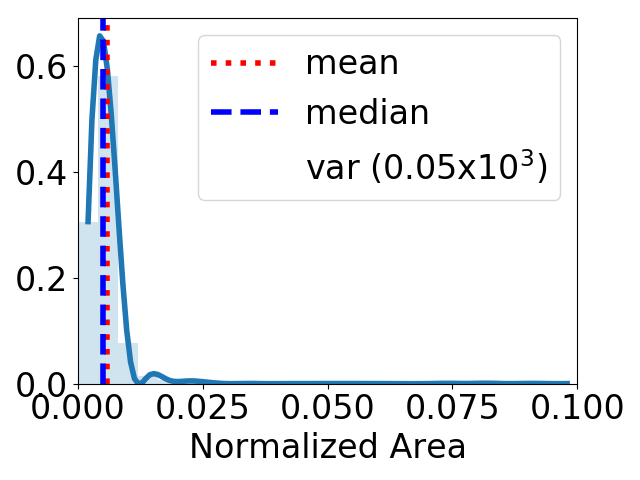}
\end{subfigure}
\hfill
\begin{subfigure}[b]{0.19\linewidth}
    \centering
	\includegraphics[width=1.05\linewidth]{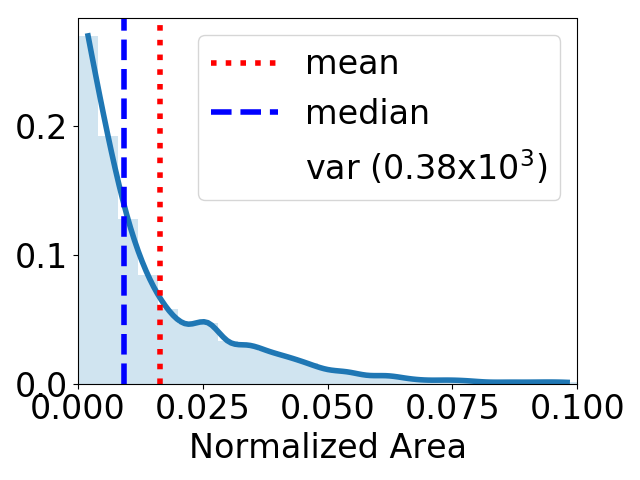}
\end{subfigure}
\hfill
\begin{subfigure}[b]{0.19\linewidth}
    \centering
	\includegraphics[width=1.05\linewidth]{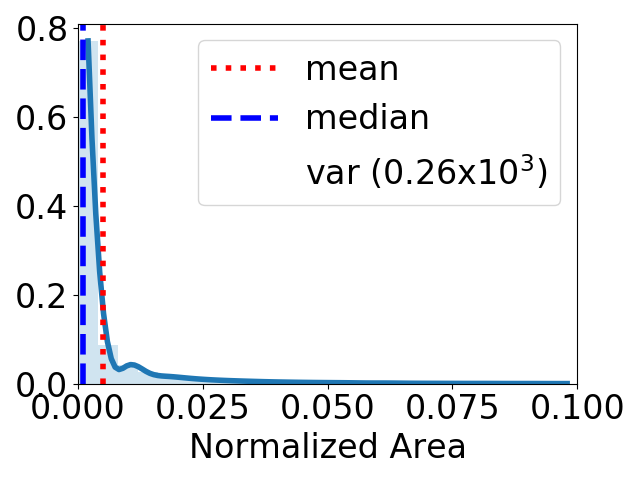}
\end{subfigure}

\begin{subfigure}[b]{0.19\linewidth}
    \centering
	\includegraphics[width=1.05\linewidth]{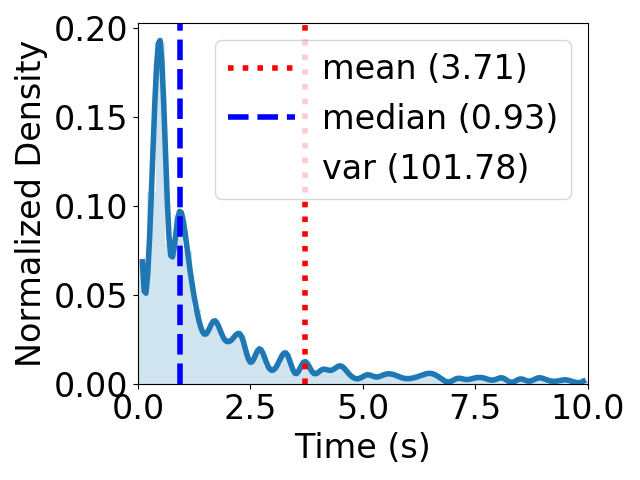}
	\caption{JRDB} 
\end{subfigure}
\hfill
\begin{subfigure}[b]{0.19\linewidth}
    \centering
	\includegraphics[width=1.05\linewidth]{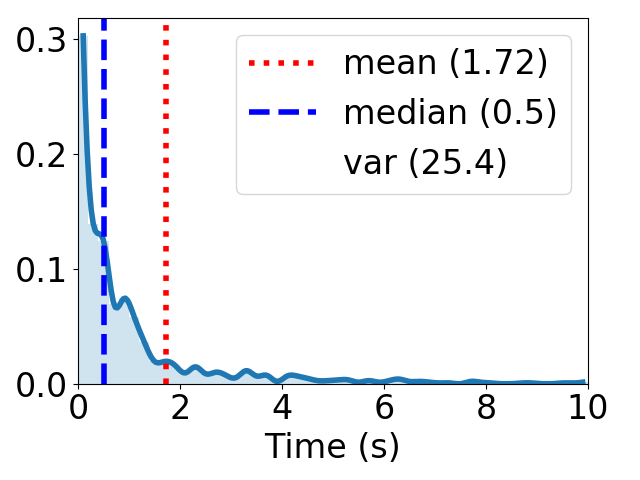}
	\caption{MOT17~\cite{MOTChallenge:arxiv:2016}} 
\end{subfigure}
\hfill
\begin{subfigure}[b]{0.19\linewidth}
    \centering
	\includegraphics[width=1.05\linewidth]{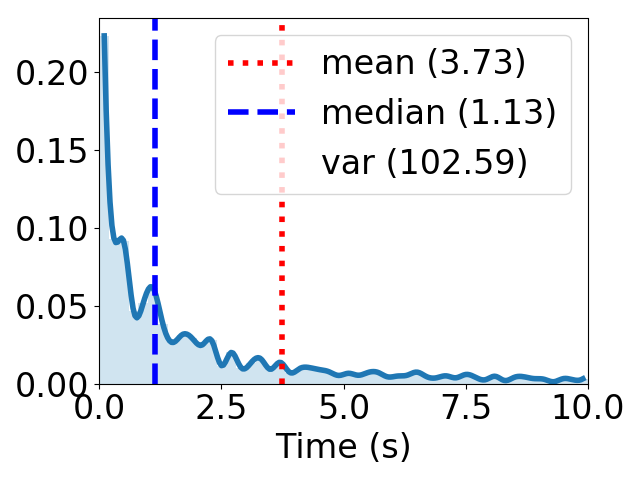}
	\caption{MOT20~\cite{dendorfer2020mot20}} 
\end{subfigure}
\hfill
\begin{subfigure}[b]{0.19\linewidth}
    \centering
	\includegraphics[width=1.05\linewidth]{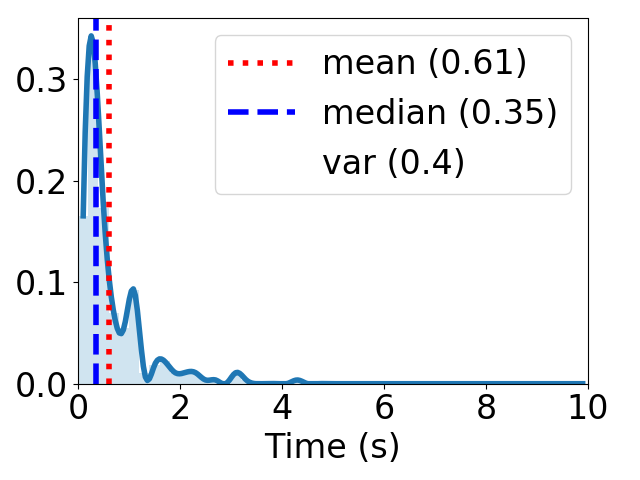}
	\caption{KITTI~\cite{Geiger2012CVPR}} 
\end{subfigure}
\hfill
\begin{subfigure}[b]{0.19\linewidth}
    \centering
	\includegraphics[width=1.05\linewidth]{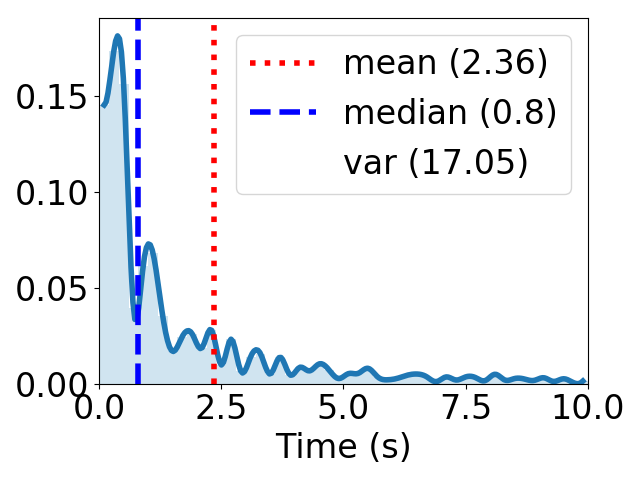}
	\caption{Waymo~\cite{sun2019scalability}} 
\end{subfigure}
 	\caption{\revised{Comparison between JRDB and similar existing datasets and benchmarks (MOT17, MOT20, KITTI, Waymo). \textit{Top row:} 2D bounding box area divided by image area in training dataset. JRDB contains images where the pedestrians occupy a larger portion due to its closer proximity to people, while keeping a greater variance in sizes than other datasets. \textit{Bottom row:} Amount of continuous time tracks are fully-occluded in training datasets. JRDB has longer and more diverse lengths in occluded track periods indicating more complex occlusion patterns caused by the robot's point of view and the higher density of the crowd. Note that long tails past end of graph are not shown.
 \label{fig:occlusion-size}	}}
 \end{figure*}
 
 \begin{figure*}[tb]
\centering
\begin{subfigure}[b]{0.49\linewidth}
    \centering
	\includegraphics[width=0.49\linewidth]{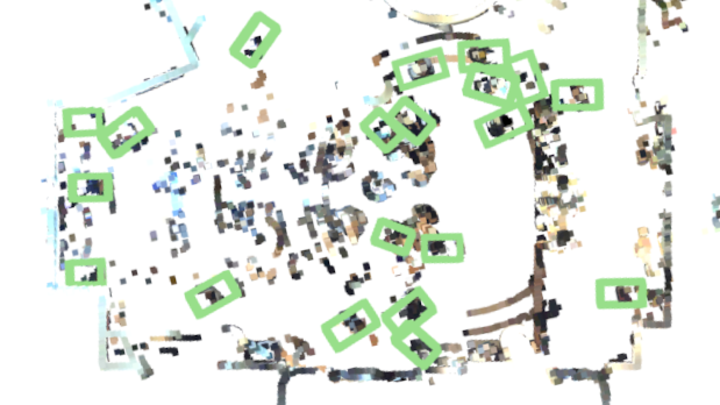}
	\includegraphics[width=0.49\linewidth]{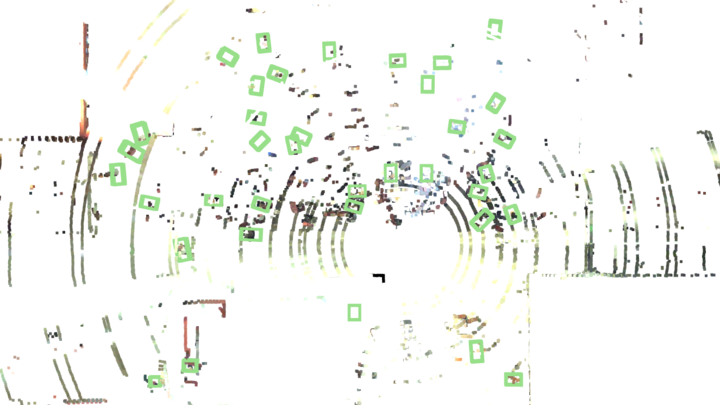}
	\caption{Indoor scenes including many natural indoor occluders such as desks and chairs (\texttt{\scriptsize bytes-cafe-2019-02-07\_0} and \texttt{\scriptsize forbes-cafe-2019-01-22\_0}) } 
\end{subfigure}
\hfill
\begin{subfigure}[b]{0.49\linewidth}
    \centering
	\includegraphics[width=0.49\linewidth]{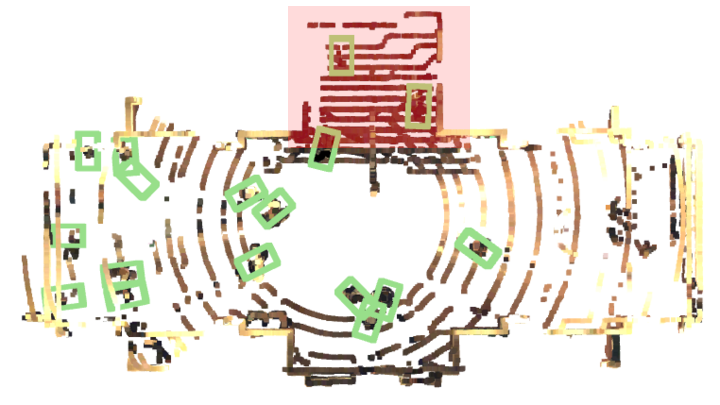}
	\includegraphics[width=0.49\linewidth]{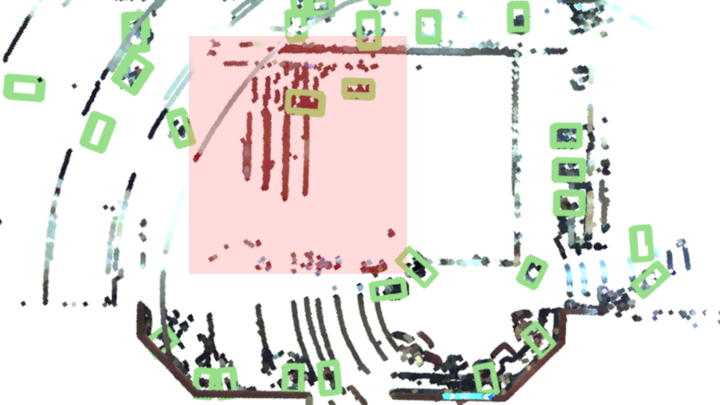}
	\caption{Indoor scenes including stairs. Stairs highlighted in red box. (\texttt{\scriptsize gates-basement-elevators-2019-01-17\_1} and \texttt{\scriptsize forbes-cafe-2019-01-22\_0})  } 
\end{subfigure}
\begin{subfigure}[b]{0.49\linewidth}
    \centering
	\includegraphics[width=0.49\linewidth]{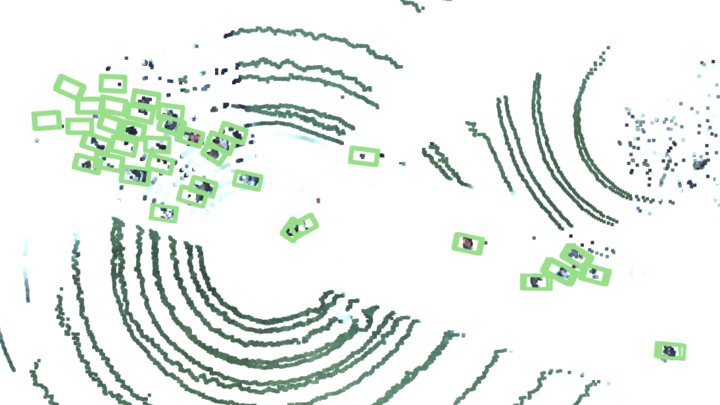}
	\includegraphics[width=0.49\linewidth]{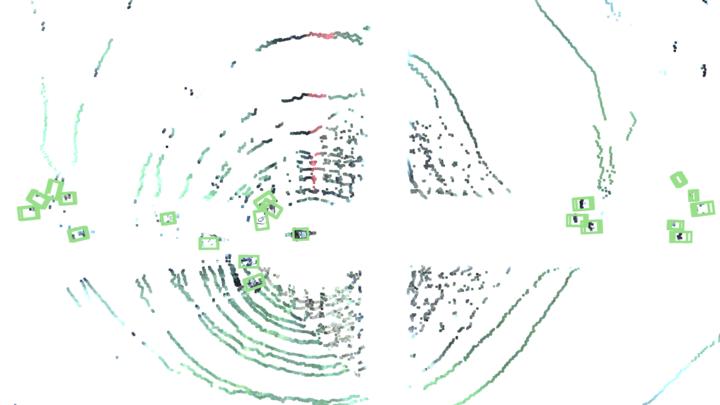}
	\caption{Outdoor scenes with large variation of distance from the robot (\texttt{\scriptsize huang-lane-2019-02-12\_0} and \texttt{\scriptsize memorial-court-2019-03-16\_0})}  
\end{subfigure}
\hfill
\begin{subfigure}[b]{0.49\linewidth}
    \centering
	\includegraphics[width=0.49\linewidth]{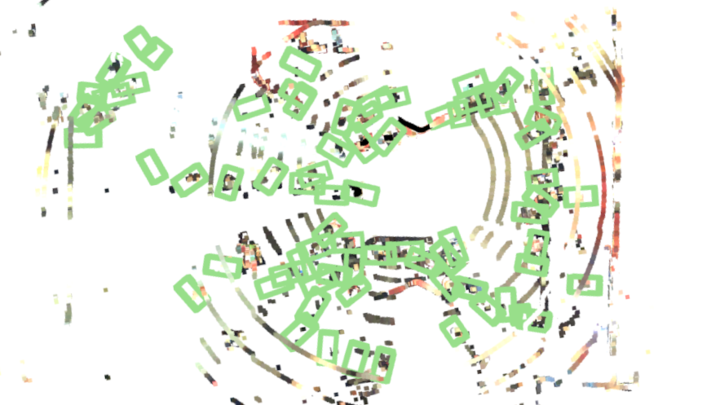}
	\includegraphics[width=0.49\linewidth]{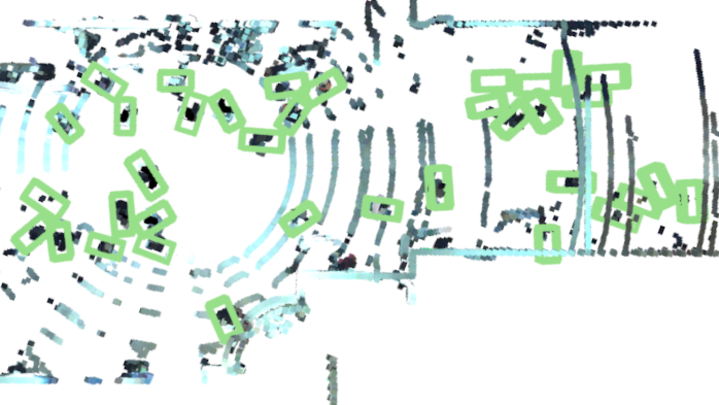}
	\caption{Very crowded scenes (\texttt{\scriptsize tressider-2019-04-26\_2} and \texttt{\scriptsize packard-poster-session-2019-03-20\_2})}  
\end{subfigure}

 	\caption{\revised{Bird's-eye view of LiDAR point clouds and 3D bounding box annotations of eight different JRDB sequences, including indoors and outdoors. The green boxes indicate pedestrians. JRDB includes an extremely large number of pedestrians at different distances and orientation to the sensors, including very close proximity ($\approx$\SI{1}{\meter}) or very distant ($>$\SI{50}{\meter}) , with social groups and clusters related to human walking patterns and activities, and navigation patterns strongly affected by the indoor and outdoor physical constraints. Other natural complexities such as cluttered indoor objects and stairs add to its challenges.} \label{fig:jrdbexamples}}
 \end{figure*}

\revised{In this section, we further highlight the unique and challenging features of our JRDB dataset compared to the relevant existing datasets such as popular pedestrian tracking benchmarks from moving camera or surveillance footage, \eg MOT17~\cite{MOTChallenge:arxiv:2016} and MOT20~\cite{dendorfer2020mot20}, and autonomous driving benchmarks, \eg KITTI~\cite{Geiger2012CVPR}, ApolloScape~\cite{wang2019apolloscape}, BDD100K~\cite{yu2018bdd100k}, NuScenes~\cite{caesar2019nuscenes} and Waymo~\cite{sun2019scalability}. Table~\ref{tab:numerical_stats} provides a summary of the most relevant statistics of the data in our novel dataset. Table~\ref{tab:dataset_comparison} provide a comparison between JRDB and all aforementioned datasets, in respect to their most significant statistics. We observe that JRDB is the only multimodal dataset that includes indoor and outdoor data, with the largest number of annotated 2D bounding boxes around pedestrians and second largest number of annotated 3D bounding boxes.} 

\revised{The MOTChallenge datatsets~\cite{MOTChallenge:arxiv:2015,MOTChallenge:arxiv:2016, dendorfer2019cvpr19, dendorfer2020mot20}, as one of the most popular benchmark datasets for pedestrian detection and tracking, provide high quality manual 2D detection and tracking annotations for all the pedestrians in videos which are mainly captured from different outdoor scenes using a moving or surveillance camera. Compared to the MOTChallenge datasets, JRDB data contains more scene diversity, higher number of sequences, frames and number of 2D bounding box annotations with competitive quality in the annotation tightness around the pedestrians. In addition, JRDB data includes 3D sensor signals, \ie LiDAR, and 3D bounding box and tracking annotations in 3D, providing an opportunity to study and develop methods to perceive human motion and behaviour in 3D Euclidean coordinates, a more natural space.}

Compared to the recent autonomous driving datasets, JRDB data may not contain as many number of sequences, frames and tracks. However, the number of annotated 2D and 3D bounding boxes is in par or even higher than those in these datasets (see \Tab~\ref{tab:dataset_comparison}). This confirms that JRDB contains longer pedestrian trajectories in each sequence and has been annotated with considerably higher temporal resolution, \SI{7.5}{\hertz}, and interpolated to the original sensor frequency of \SI{15}{\hertz}, and thus, 50\% of the sensor data has been annotated with ground truth 2D and 3D bounding boxes. This high frame rate annotation is essential for the perception and analysis of the fine-grain and intricate human motion, specially in indoor scenes: while human trajectories are usually more linear and regular in outdoor street environments, in indoor environments and other pedestrian areas, or when the number of pedestrians is very high, the trajectories of humans are more intricate and with more abrupt changes. Furthermore, since the JRDB sequences are captured in both indoor and outdoor environment in close proximity to human, it contains very unique scenarios and diversity compared to the self driving data-sets (for visualizations of all scenes, see Fig.~A.3 in Appendix). \revisedTwo{Moreover, while the number of sequences included is not large, it is enough to split in training and test sets that represent similarly the distribution of scenes, as we support empirically in our experimental evaluation (see discussion on Sec.~\ref{ss_2ddet} and Sec.~\ref{ss_3ddet}).}

Additionally, in comparison to high view's sensory setup in surveillance datasets or self-driving datasets that are captured from the top of a car, the JRDB sequences have been captured from a human-size robot. Consequently, the egocentric and perceptive view of its sensors, naturally introduces additional unique features such as larger size variations of 2D bounding boxes in camera view and longer partial and full occlusion of pedestrians. The occlusions become exacerbated in the indoor sequences where there exist many natural occluders. In \Fig~\ref{fig:occlusion-size}, we include the results of an evaluation on the size variations and full occlusion duration in JRDB sequences compared to competitive datasets such as MOT17, MOT20, KITTI and Waymo.

\Fig~\ref{fig:jrdbexamples} depicts a few examples of JRDB sequences illustrating some of its unique features and challenges, \eg a) indoor cluttered sequences such as cafes, restaurants or classrooms, b) sequences with walking areas at different heights including some sequences with stairs, c) outdoor sequences with a large variation of pedestrian distance and orientation respect to the robot, and d) very crowded sequences with more than 200 people in the field of view. In \Sec~\ref{evaluation}, we will demonstrate how these challenges will affect the performance of the state-of-the art 2D and 3D detection and tracking methods.
~\\


\section{Benchmarks and Metrics}
\label{benchmark}

Based on our associated annotations in RGB images and LiDAR point clouds, we currently propose the following four benchmarks: 
\begin{itemize}
    \item 2D person detection (in images)
    \item 3D person detection
    \item 2D person tracking (in images)
    \item 3D person tracking
\end{itemize}
Participants are free to use any of the available sensor modalities for each of the tasks, \eg  they can use RGB images and 3D point clouds for 2D person detection or for 3D person tracking.

In the following we describe the evaluation criteria and metrics we use for detection and tracking.

\subsection{\revised{Intersection over Union (IoU)}}
\revised{Intersection over Union (IoU) is the most commonly used similarity measures between two shapes, \eg bounding boxes, and hence the defacto base-similarity measure of many detection and tracking performance criteria. We also rely on this similarity measure as our base metric to compare two bounding boxes in 2D or 3D.}

\revised{Given two bounding boxes, \ie $b_p, b_g \subset  \mathbb{R}^\ell$, where $\ell = 2$ or $3$ in the case of 2D or 3D respectively, IoU is attained by $$ IoU(b_p,b_g)=\frac{\left|b_p\cap b_g\right|}{\left|b_p\cup b_g\right|}=\frac{\left|b_p\cap b_g\right|}{\left|b_p\right|+ \left|b_g\right|-\left|b_p\cap b_g\right|},$$ where $IoU(\cdot,\cdot)\in[0;1]$ and $\left|\cdot \right|$ denotes hyper-volume. }

\revised{For the 2D detection and tracking problems where the
task is to compare two sets of axis aligned bounding boxes, the calculation of IoU between each pair of bounding box has a straightforward analytical solution~\cite{rezatofighi2019generalized}. In this case,
the intersection, \ie $I(b_p,b_g) =\left|b_p\cap b_g\right|$,  between two axis aligned bounding boxes is also an axis aligned bounding box, which its vertices' coordinates are simply attained by comparing each vertices’ coordinates of $b_p$ and $b_g$ using min and max
functions~\cite{rezatofighi2019generalized}.} 

\revised{In the 3D case in our dataset, the predicted and truth bounding boxes, \ie $b_p$ and $b_g$, can be generally two arbitrary size cuboids with one degree of rotation around the axis, perpendicular to the world surface plan. Therefore, any intersection between these two cuboids, $I(b_p,b_g)$, would form a right prism with a base which can be any arbitrary 2D convex polygon. To calculate the volume of the right prism, \ie $I(b_p,b_g)$, we first calculate the area of the 2D polygon and then multiply the area with the minimum height between $b_p$ and $b_g$.} 

\revised{To attain the area of the 2D polygon surface, we use a combination of the Sutherland-Hodgman algorithm~\cite{sutherland1974reentrant} and the shoelace formula (surveyor's formula) to determine the area of 2D polygon. The Sutherland-Hodgman algorithm is an algorithm used to clip convex polygons. Given two non-axis aligned 2D bounding boxes, $\hat{b}_p$ and $\hat{b}_g$ (considering the projection of the cuboids seen from bird-eye view), we extend each line segment of $\hat{b}_g$ to clip the edges of $\hat{b}_p$, \ie the edges of $\hat{b}_p$ are restricted to fall within $\hat{b}_g$. The resulting clipped polygon represents the area of intersection. Given the set of vertices of this polygon, sorted in clockwise order, we apply the shoelace formula to find its area. The shoelace formula is given by:
$$ I(\hat{b}_p,\hat{b}_g) = \frac{1}{2} \lvert \sum_{i=1}^{n-1}\alpha_i \beta_{i+1} + \alpha_n \beta_1 - \sum_{i=1}^{n-1}\alpha_{i+1} \beta_i - \alpha_1 \beta_n \rvert.$$
where $I(\hat{b}_p,\hat{b}_g)$ is the area of the intersecting polygon between $\hat{b}_p$ and $\hat{b}_g$ with $n$ vertices $(\alpha_1, \beta_1), (\alpha_2, \beta_2) \dots (\alpha_n, \beta_n)$. } 
\\
\revised{\subsection{Evaluations for Detection Benchmarks}}

\revised{We evaluate 2D and 3D detection performance using two criteria : \emph{(i)} Average Precision (AP), as the most widely used performance measure for object detection benchmarks~\cite{pascal-voc-2008,coco,Geiger2012CVPR}, \emph{(ii)} a new IoU-based optimal sub-pattern assignment (OSPA$_{IoU}$) metric for calculating a distance between two set of shapes, \eg 2D and 3D bounding boxes~\cite{rezatofighi2020how}.} 

\revised{Here, the aim is to find a measure (or a metric), defining the similarity (or distance) between a set of predicted bounding boxes, $B_p=\{b^1_{p},...,b^n_{p}\}$, and a set of the truth bounding boxes,  $B_g=\{b^1_{g},...,b^m_{g}\}$. }

\revised{{\bf Average Precision (AP):} The value of AP is based on the notion of \textit{true positives} of the prediction set, which are determined by matching predicted bounding boxes with truth bounding boxes such that the IoU value between them is larger than a specified threshold, $tr$.}

\revised{Considering the set of predicted and truth bounding boxes as $B_p=\{b^1_{p},...,b^n_{p}\}$, and $B_g=\{b^1_{g},...,b^m_{g}\}$ respectively, the \textit{true positives} set is a subset of the predicted boxes, \ie $B_{TP} \subset B_{p}$, which fulfills the following condition:}
\revised{\begin{align}
    B_{TP} = \{b^i_{p} | \forall i\in \{1,\cdots,n\}, \: IoU\left(b^{i}_p,b^{\pi^*(i)}_g\right)\geq tr \}, 
\end{align}
where $b^{\pi^*(i)}_g$ is a corresponding unique matched bounding box from the the truth bounding box set $B_g$ for the predicted box $b^{i}_p$. All matching assignments between the prediction and truth sets are attained by $\pi^*=\argmax_{\pi\in\Pi_{m}}\sum_{i=1}^{n} IoU(b^{i}_p,b^{\pi(i)}_g)$. Note that if $m\geq n$ for each prediction, $b^{i}_p$, after solving the assignment problem, there is a match $b^{\pi^*(i)}_g$. Otherwise, some $b^{i}_p$ may not be matched to any truth box and thus, they are also excluded from true positive list.  Similar to all the existing popular object detection benchmarks~\cite{pascal-voc-2008,Geiger2012CVPR, coco}, in our benchmark, the non-optimal greedy assignment strategy is used for solving the assignment problem\footnote{\revised{The predictions are sorted according to their confidence scores and then  matching is greedily performed for the prediction with highest confidence scores first.}}.  }

\revised{Next, precision and recall are respectively calculated by $ Pr = \frac{|B_{TP}|}{n} $ and $Rc = \frac{|B_{TP}|}{m}$, where $|B_{TP}|$ is the cardinality of true positive sets (or simply the number of true positives). }

\revised{Finally, we take average of 11-point interpolated precision to compute mean average precision, just as in Pascal VOC2008~\cite{pascal-voc-2008}:
\begin{align*}
    AP = \frac{1}{11}\sum_{Rc = 0:0.1:1}\max_{\widetilde{Rc}:\widetilde{Rc}\geq Rc}\quad\textit{Pr}(\widetilde{Rc})
\end{align*}
where $\textit{Pr}(Rc)$ is the measured precision at recall $Rc$.}
~

\revised{
In our benchmark, a prediction is considered true positive if the IoU between the prediction and the ground-truth bounding box is over 0.5 for 2D and 0.3 for 3D. For 2D object detection benchmark, the bounding boxes with occlusion level lower than \textit{Fully\_occluded} and area greater than 500 pixel$^2$ are considered for the benchmark. For 3D object detection benchmark, the 3D boxes that encloses more than 10 points and are within 25m of the sensor are considered for the benchmark. }

\revised{In spite of popularity of AP, in~\cite{rezatofighi2020how} it is demonstrated that AP can be an unreliable performance measure when the aim is use this measure for ranking competitive submissions in a benchmark. Moreover, the rankings can change dramatically with different choices of IoU threshold. To this end, in our benchmark, we also report OSPA$_{IoU}$ as a reliable metric for object detection as shown in~\cite{rezatofighi2020how}.}

\revised{{\bf IoU-based Optimal Sub-Pattern Assignment (OSPA$_{IoU}$):} In contrast to the true positive-based measures, \eg AP, which relies on an IoU threshold  between the pairs of predictions and truth to determine true positives, OSPA$_{IoU}$ is a set-based metric which can directly capture a distance, $d_{O}$, between two sets of bounding boxes, \ie  $B_p=\{b^1_{p},...,b^n_{p}\}$ and $B_g=\{b^1_{g},...,b^m_{g}\}$, without a thresholding parameter. If $m\geq n>0$, $d_{O}$ is calculated as follows:
\begin{align}
d_{O}(B_p,B_g)\!=
& \frac{1}{m}\left(\!\min_{\pi\in\Pi_{m}}\sum_{i=1}^{n}\!d_{IoU}\left(b^{i}_p,b^{\pi(i)}_g\right)\!+\!\left(n-m\right)\right)\label{eq:OSPA-dist}
\end{align}\\
 where $\Pi_{m}$ is the set of all permutations of $\left\{ 1,2,...,m\right\}$ and $d_{IoU}\left(b^{i}_p,b^{j}_g\right) = 1- IoU\left(b^{i}_p,b^{j}_g\right)\in[0;1]$. Consequently, $d_{O}(B_p,B_g)\in[0;1]$. Note that if
$n>m>0$; $d_{O}(B_p,B_g)=d_{O}(B_g,B_p)$. Moreover, if one of the set is empty; $d_{O}(B_p,B_g)=1$,  and $d_{O}(\emptyset,\emptyset)=0$. }
 
\revised{Please note that the OSPA distance is the combination of two types of errors: \emph{i)} the term $(n-m)$ representing cardinality mismatch between two sets, penalizing missed or false detections without an explicit definition for them, and \emph{ii)} the term $\min_{\pi\in\Pi_{m}}\sum_{i=1}^{n}\!d_{IoU}\left(b^{i}_p,b^{\pi(i)}_g\right)$ which is an equivalent matching problem discussed in the previous section and is simply an assignment problem between the predicted and ground-truth bounding boxes based on IoU distances between each pair. However, in contrast to AP, there does not exist any notation of true positives based on an arbitrary threshold and the cost of optimal assignment is interpreted as localization error. Additionally, the optimal assignment for the OSPA distance can be attained in polynomial time complexity using a Munkres (Hungarian) algorithm~\cite{kuhn1955hungarian}. }
\\
\revised{\subsection{Evaluations for Multi-Object Tracking Benchmarks}}
\revised{We evaluate 2D and 3D multi-object tracking performance using two criteria: \emph{(i)} the Clear-MOT measures~\cite{bernardin2008evaluating}, as the most widely used performance criteria for multi-object tracking benchmarks~\cite{MOTChallenge:arxiv:2016,Geiger2012CVPR}, \emph{(ii)} a IoU-based OSPA$^{(2)}$ (OSPA$^{(2)}_{IoU}$) metric for calculating a distance between two set of tracks of 2D and 3D bounding boxes~\cite{rezatofighi2020how}.}

\revised{Here, the aim is to find a measure (or a metric) defining a similarity (or a distance) between a set of $n$ predicted tracks, $\mathcal{B}_p=\{B^{1}_{p}(t^1_{s:e}),...,B^{n}_{p}(t^n_{s:e})\}$, and a set of $m$ truth tracks,  $\mathcal{B}_g=\{B^{1}_{g}(t^1_{s:e}),...,B^{m}_{g}(t^m_{s:e})\}$, where $B^{i}(t^i_{s:e})$ is an $i^{th}$ trajectory which is an ordered collection of bounding boxes in time between $t^i_s$ and $t^i_e$, \ie $B^{i}(t^i_{s:e}) = \left(b^i(t^i_s),\cdots, b^i(t^i_e)\right)$. }

\revised{{\bf Clear-MOT measures:} These measures are known as multiple object tracking accuracy (MOTA) and multiple object tracking precision (MOTP), which are attained for a sequence with $T$ frames as follows:}

\revised{\emph{i)} For all the predicted and ground truth tracks having a bounding box at time $t$, we find the \textit{true positives} set at time $t$,  by:}
\revised{\begin{align}
    \mathcal{B}_{TP}(t) = \{& b^i_{p}(t) | \forall i\in \{1,\cdots,n\}, \nonumber\\ & t^i_s \leq t\leq t^i_e, \: IoU\left(b^{i}_p(t),b^{\pi^*(i)}_g(t)\right)\geq tr \}, 
\end{align} 
 where $b^{\pi^*(i)}_g(t)$ is a corresponding unique matched bounding box from the available truth set at time $t$ for the predicted box $b^{i}_p (t)$ at the same frame. All matching assignments between the prediction and truth sets are optimally attained by $\pi^*=\argmax_{\pi\in\Pi_{m}}\sum_{i=1}^{n} IoU(b^{i}_p(t),b^{\pi(i)}_g(t))$ using the Hungarian algorithm.}

\revised{\emph{ii)} Next, the false positive set $\mathcal{B}_{FP}(t)$ at time $t$ is calculated by excluding true positive set from the prediction set at the same frame. Similarly, the false negative set, $\mathcal{B}_{FN}(t)$ at time $t$, is attained by excluding the set of the matched truth bounding boxes from the truth set  at time $t$. ID switches $\mathcal{B}_{ID}(t)$ at time $t$ will be the mismatch the assignments between the matching solutions at time $t$ and $t-1$ for the same truth track.   }

\revised{\emph{iii)} Finally, MOTA and MOTP are respectively calculated by
\begin{align}
\textit{MOTA} &= 1 -  \frac{\sum_{t=1}^T|\mathcal{B}_{FP}(t)|+|\mathcal{B}_{FN}(t)|+|\mathcal{B}_{ID}(t)|}{\sum_{t=1}^T|\mathcal{B}_{g}(t)|} \\
\textit{MOTP} &=  \frac{\sum_{t=1}^T\sum_{i=1}^{n} IoU(b^{i}_p(t),b^{\pi^*(i)}_g(t))\geq tr}{\sum_{t=1}^T |\mathcal{B}_g^{\pi^*}(t)|}
\end{align}\\
where $|.|$ represent the cardinally and $|\mathcal{B}_g^{\pi^*(t)}|$ is the total number of ground truth tracks, matched to the predictions at time $t$.
}

\revised{
The Clear-MOT measures are also based on the notion of \textit{true positives} of the prediction set, which are determined by matching predicted bounding boxes at each frame with truth bounding boxes at the same frame such that the IoU value between them is larger than a specified threshold. In our benchmark, a prediction is considered true positive if the IoU between the prediction and the ground-truth bounding box is over 0.5 for 2D and 0.3 for 3D. We do not require matches for boxes beyond 25 meters.
}

\revised{Similar to AP, Clear-MOT measures are shown to be very sensitive to choices of IoU threshold and the rankings of submissions can vary noticeably depending on the choices of IoU threshold. To this end, in our benchmark, we also report OSPA$^{(2)}_{IoU}$ as a reliable metric for multi-object tracking as discussed in~\cite{rezatofighi2020how}.}

\revised{{\bf IoU-based OSPA$^{(2)}$ (OSPA$^{(2)}_{IoU}$):} OSPA$^{(2)}_{IoU}$ is a set-based metric which can directly capture a distance, $d_{O}$, between two sets of trajectories, \ie  $\mathcal{B}_p=\{B^{1}_{p}(t^1_{s:e}),...,B^{n}_{p}(t^n_{s:e})\}$, and  $\mathcal{B}_g=\{B^{1}_{g}(t^1_{s:e}),...,B^{m}_{g}(t^m_{s:e})\}$, without a thresholding parameter.}

\revised{A meaningful distance between two sets of tracks requires a meaningful base-distance between two tracks first. In~\cite{rezatofighi2020how}, the following time-averaged OSPA distance
between two tracks $B^i_p(t^i_{s:e})$ and $B^j_g(t^j_{s:e})$ are suggested:
\begin{align}
\underline{\widetilde{d}}\left(B^i_p(t^i_{s:e}), B^j_g(t^j_{s:e})\right)= & \sum _{t=t^m_s}^{t^M_e}\!\frac{d_{O}\left(B^i_p(t) ,B^j_g(t)\right)}{t^M_e-t^m_s},
\label{eq:OSPA2-base}
\end{align}\\
where $d_O(\cdot,\cdot)$ is OSPA distance in \Eq~\ref{eq:OSPA-dist} and $t^M_e = \max(t^i_e,t^j_e)$ and $t^m_s=\min(t^i_s,t^j_s)$. Note that, in some frames, both $B^i_p(t)$ and $B^j_g(t)$ may include a bounding box, however since each track can start and end in different time, in some frames, one of them may exist only, \ie $B^i_p(t)=\emptyset$ while $B^j_g(t) = b^j_g(t)$ or vice versa. In both scenarios $d_O(\cdot,\cdot)$ has a value as a set distance, \ie measuring a distance between two singleton sets (two bonding boxes) or one singleton set and one empty set.}

\revised{Having $\underline{\widetilde{d}}$ as a base distance between two tracks, the final distance between two sets of track can be calculated with another OSPA distance (\Eq~\ref{eq:OSPA-dist}) when the base distance is replaced with $\underline{\widetilde{d}}$ in \Eq~\ref{eq:OSPA2-base}, \ie if $m\geq n>0$, $d_{O}$ then,  
\begin{align}
d_{O}(\mathcal{B}_p,\mathcal{B}_g)=
\frac{1}{m}\Bigg(\min_{\pi\in\Pi_{m}}\sum_{i=1}^{n}\underline{\widetilde{d}}\left(B^{i}_p(\cdot), B^{\pi{*}(i)}_g(\cdot)\right)&\nonumber \\+\left(n-m\right)&\Bigg),\label{eq:OSPA2b-dist}\end{align}
and if
$n>m>0$; $d_{O}(\mathcal{B}_p,\mathcal{B}_g)=d_{O}(\mathcal{B}_p,\mathcal{B}_g)$. Note that $d_{O}(\mathcal{B}_p,\mathcal{B}_g)\in[0;1]$.}

\revised{Since OSPA distance is calculated in two levels (first in \Eq~\ref{eq:OSPA2-base} and next in \Eq~\ref{eq:OSPA2b-dist}), this distance is known as OSPA$^{(2)}$. Please note that the OSPA$^{(2)}$ distance is also the combination of two types of errors: \emph{i)} the term $(n-m)$ representing cardinality mismatch between two sets, penalizing missed or false tracks without an explicit definition for them, and \emph{ii)} the term $\min_{\pi\in\Pi_{m}}\sum_{i=1}^{n}\left(B^{i}_p(\cdot), B^{\pi{*}(i)}_g(\cdot)\right)$ representing different tracking error such as the displacement and size errors, track ID switches, track fragmentation or even track late initiation/early termination. }

~
\section{Evaluation}
\label{evaluation}

In this section, we evaluate the performance of several state-of-the-art/popular approaches for detection and tracking on the egocentric data of the JRDB dataset. By analyzing their performance and comparing in different scenarios, we also provide further insights on the special nature of our dataset and on areas of research that should be further explored to support robotic agents that need to perceive robustly in public human environments.

\subsection{2D Person Detection}
\label{ss_2ddet}

We first analyze several popular algorithms for 2D object detection. We provide results in our challenge of three widely-adopted and top-ranking solutions: YOLOv3, RetinaNet and Faster R-CNN. These solutions include single and double stage 2D detectors.

\begin{table*}[tb!]
\revised{\scriptsize
    \centering
\begin{tabular}{ |l||c| c |c||c | c |c c c||}
\hline
\multirow{2}{*}{\backslashbox{Method}{Metric}}& \multicolumn{3}{c||}{\%AP$_{IoU=.5}$$\uparrow$}& \multicolumn{5}{c||}{OSPA$_{IoU}$$\downarrow$}\\
\cline{2-9} 
& Indoor & Outdoor & Overall &  Indoor & Outdoor& Overall  & Card. & Loc. \\ \hline
Faster R-CNN &  \bf 54.1 & \bf 48.1646 & \bf 52.2 & \bf \revisedTwo{0.619}& \bf\revisedTwo{0.736}& \bf\revisedTwo{0.668}& \bf\revisedTwo{0.408}& \revisedTwo{0.260} \\ \hline
RetinaNet & 53.0 & 43.1 &50.4 &\revisedTwo{0.628}&\revisedTwo{0.787}&\revisedTwo{0.694}&\revisedTwo{0.476}& \revisedTwo{0.218}\\ \hline
YOLOv3 &  40.5 & 45.3 &41.7 &\revisedTwo{0.687}&\revisedTwo{0.764}&\revisedTwo{0.719}&\revisedTwo{0.558}& \bf\revisedTwo{0.161} \\ \hline
\end{tabular}
    \caption{2D detection results in the JRDB benchmark using average precision (AP) and OSPA$_{IoU}$. Card. $=$ Cardinality, Loc. $=$ Localization.}
    \label{tab:2ddetection}}
\end{table*}

\begin{figure}[tb]
\centering
\begin{subfigure}[b]{0.49\linewidth}
    \centering
	\includegraphics[width=\linewidth]{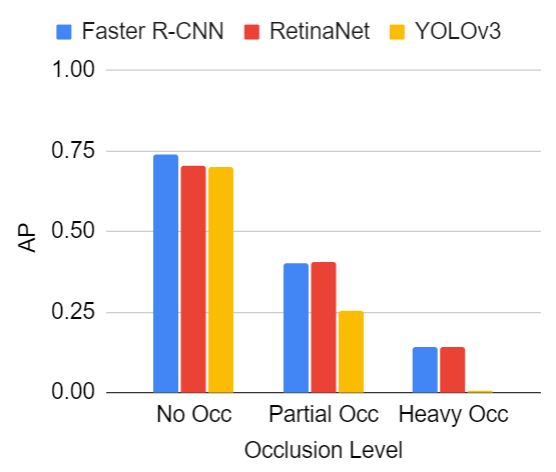}
	\caption{Average Precision (AP)} 
\end{subfigure}
\hfill
\begin{subfigure}[b]{0.49\linewidth}
    \centering
	\includegraphics[width=\linewidth]{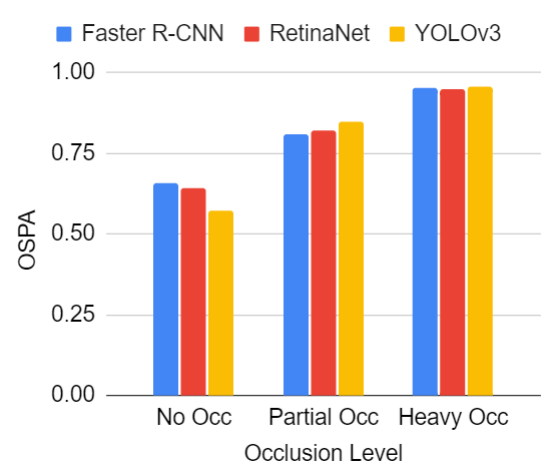}
	\caption{OSPA$_{IoU}$} 
\end{subfigure}
 	\caption{\revised{Sensitivity to occlusion of methods in the JRDB 2D detection benchmark. We include average precision (AP) and OSPA$_{IoU}$ at 3 different occlusion (Occ.) levels. All methods get significantly worse with occlusion, but YOLOv3 is the most strongly affected.}}
 	
    \label{fig:2ddetection_occ}
 \end{figure}

\emph{YOLOv3~\cite{DBLP:journals/corr/abs-1804-02767}\footnote{\revised{Code: \url{https://github.com/eriklindernoren/PyTorch-YOLOv3}}}:} is one of the most popular single stage detector. We use a publicly available implementation with weights trained on the COCO dataset~\cite{DBLP:journals/corr/LinMBHPRDZ14}. Further, we fine-tune these weights on our egocentric JRDB dataset with the default parameters \revised{until the training loss converges, \ie the descent rate on the training loss is negligible.}

The original YOLOv3 model uses square images of $416\times416$ pixels. However, our RGB sensor provides 360$^\circ$ cylindrical images of $3760\times480$ pixels, a small width-to-height ratio. To adapt to this difference, we segment each cylindrical image into subimages with width-to-height ratio close to the unit and use them both for training and inference. We finally map the detections in the segmented images to the original cylindrical image. In the intersection between two segmented images, we search for detections that correspond to the same person by finding pairs of boxes at similar height and with similar size, and merge them together. While this process may introduce very few false merged detections, we found the rate of these failures negligible and the performance and computational speed of YOLOv3 is greatly improved compared to using the entire cylindrical image. 

\emph{RetinaNet~\cite{retinanet}\footnote{\revised{\label{note1}Code: \url{https://github.com/facebookresearch/detectron2}}}:} is a very popular single stage detector well-known for improved performance thanks to a novel focal loss. We use the publicly available implementation with the ResNet-50 backbone. We start with pretrained weights on the COCO dataset and fine-tune to the JRDB dataset with the default parameters \revised{until the training loss converges, \ie the descent rate on the training loss is negligible.}

\emph{Faster R-CNN~\cite{NIPS2015_5638}\footnoteref{note1}:} is the most widely used two stage detector, which first generates several good proposals for the objects of interest in the scene independent from their category labels, then in the second stage they are accurately localized and classified to their category labels. \revised{We use the publicly available and recent implementation of this detector which uses ResNet-50 backbone, a feature pyramid network and RoI Align pooling technique. Similar to the other detectors, we start with pretrained weights from the COCO dataset and fine-tune on the JRDB dataset with the default hyper-parameters until the training loss converges, \ie the descent rate on the training loss is negligible.}

\subsubsection*{Results \& Discussion}
The results of our experiments are summarized in Table~\ref{tab:2ddetection}. \revised{According to the results, all these state-of-the-art detectors perform poorly in our egocentric dataset, as indicated by AP scores ($\approx \%50$) and very high OSPA$_{IoU}$ (the values are closer to 1 than zero). We hypothesize that their low performance is caused by the very challenging scenes included in our egocentric dataset, with up-to hundreds of humans moving, the robot (sensors) own motion, and high-level of partial and heavy occlusions due to the first-person view setup. This further indicates that our dataset and benchmark poses challenges to existing solutions due to the novel type of data and scenes, and as a benchmark can guide research into these unsolved and necessary new areas.}

\revised{When we compare the detectors against each other using the different performance measures, the best performing one according to AP and OSPA$_{IoU}$ is Faster R-CNN, although YOLOv3 has better localization error compared to Faster R-CNN and RetinaNet.}
\revisedTwo{We hypothesise this is because YOLOv3 is able to successfully localize well the larger pedestrians but has a high cardinality error as it struggles with smaller boxes due to sparser anchor boxes. When decreasing the confidence threshold for OSPA to decrease cardinality error and consider more detections, we observe locality error rapidly rises and surpasses the locality error for other methods. This suggests YOLOv3 gives low confidence low IoU detections for small and occluded pedestrians, giving the lower AP score. As Faster-RCNN gives the best AP scores with the current IoU threshold ($tr = 0.5$) and the best overall OSPA$_{IoU}$ score, we use the output of Faster-RCNN as public detections for our 2D tracking benchmarks.}

\begin{table*}[!htb]
\revised{\scriptsize
    \centering
\begin{tabular}{ |l||c|c|c|c|c|c|c||c|c|c c c||}
\hline
\multirow{3}{*}{\backslashbox{Method}{Metric}}& \multicolumn{7}{c||}{Clear-MOT$_ {IoU=.5}$}&\multicolumn{5}{c||}{OSPA$^{(2)}_{IoU}$$\downarrow$}\\
\cline{2-8} 
 & \multicolumn{3}{c|}{\%MOTA$\uparrow$} & \multirow{2}{*}{\%MOTP$\uparrow$}  & \multirow{2}{*}{$|\mathcal{B}_{ID}|\downarrow$} & \multirow{2}{*}{$|\mathcal{B}_{FP}|\downarrow$} & \multirow{2}{*}{$|\mathcal{B}_{FN}|\downarrow$}&\multicolumn{5}{c||}{}\\ \cline{2-4}\cline{9-13} 
 & Indoor & Outdoor & Overall&&&&&Indoor & Outdoor & Overall & Card. & Loc. 
 \\\hline
    DeepSORT &\bf 25.5 & 18.5& \bf 23.2 &  24.6 &  \bf 5296 & 78947 &  \bf 650478 &\bf0.947&\bf0.956&\bf0.951&\bf0.536&0.415\\\hline
JRMOT & 24.3 & \bf 19.0 & 22.5 & 23.6 & 7719 & \bf 65550 & 667783 &0.973&0.980&0.977&0.748&0.229\\ \hline
Tracktor & 19.9 & 14.2 & 18.1 & \bf 29.2 & 7564 & 88480 & 687842  &0.971&0.985&0.978&0.759&\bf0.219\\ \hline
\end{tabular}
    \caption{2D tracking results in the JRDB benchmark using Clear-MOT metrics and OSPA$^{(2)}_{IoU}$. IDS $=$ ID Switches, FP $=$ False Positives, FN $=$ False Negatives, Card. $=$ Cardinality, Loc. $=$ Localization.}
    \label{tab:2dtracking}}
\end{table*}

\emph{Indoor vs. Outdoor Scenes:} 
\revised{Based on both performance measures, AP and OSPA$_{IoU}$, all three detectors seem to perform better in indoor environments. In outdoor scenes, pedestrians are commonly further away from the robot (sensors). Therefore, it would be difficult task for all detectors to detect these small bounding boxes. In addition, YOLOv3 performs much weaker in localizing pedestrians very close to robot, \ie for very large bounding boxes, compared to the pedestrians in medium distance from the robot. This is because its anchor boxes align poorly to the size of pedestrians as images of pedestrians are often heavily cutoff and fill the screen at short distance to the robot, making accurate localization challenging. Therefore, for a $IoU$ threshold equal to $0.5$, these bounding boxes might be considered as false positives resulting in lower AP in indoor environments.}


\emph{Sensitivity to Occlusion Level:}
As expected, the performance of all three methods decrease with increased level of occlusions, as shown in Table~\ref{tab:2ddetection}. Interestingly, YOLOv3 performs drastically worse even with partial occlusions indicating a strong sensitivity to occlusions. \revisedTwo{This significantly hurts overall performance as our dataset has nearly twice the number of partially occluded pedestrians as unoccluded pedestrians due to large crowds and the egocentric human-height view point.} The significant drop of performance of all algorithms due to occlusion highlights the need for for new detectors with more robust performance that can support egocentric artificial perception for autonomous navigating agents in public human environments.

\begin{figure}[tb]
\centering
\begin{subfigure}[b]{0.49\linewidth}
    \centering
	\includegraphics[width=\linewidth]{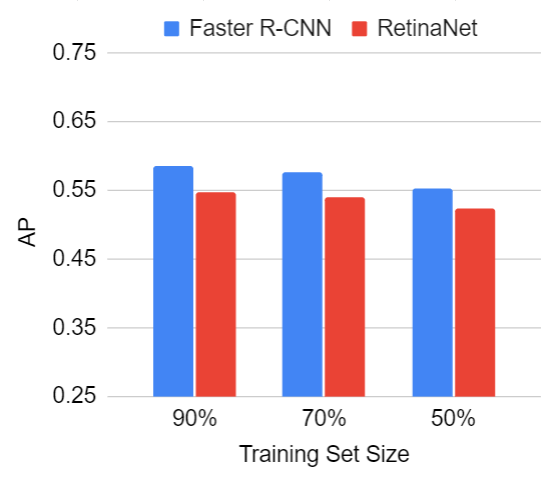}
	\caption{Average Precision (AP)} 
\end{subfigure}
\hfill
\begin{subfigure}[b]{0.49\linewidth}
    \centering
	\includegraphics[width=\linewidth]{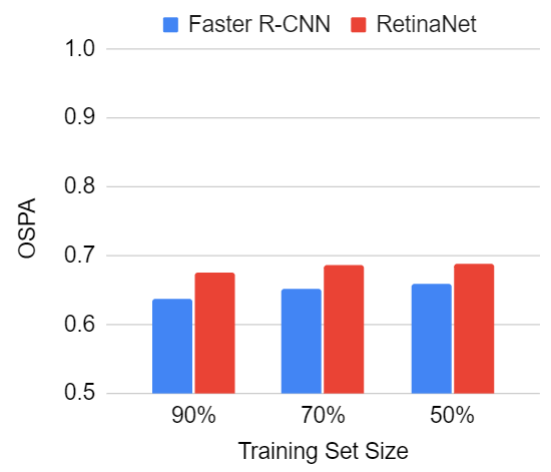}
	\caption{OSPA$_{IoU}$} 
\end{subfigure}
     \caption{\revised{Sensitivity to training set size of methods in the JRDB 2D detection benchmark. We include average precision (AP) and OSPA$_{IoU}$ at 3 different levels. Method performance marginally decreases with less data.}}
    \label{fig:2ddet_trainsetsize}
 \end{figure}

\revisedTwo{\emph{Effect of the Size of JRDB on the Results:} A possible reason for the poor performance observed from the state-of-the-art 2D detectors on the JRDB data could be the small size of the training data compared to other datasets, with only 27 different scenes. This may cause that the distribution of the training and test splits be different. To analyze the possible effect of the size and diversity of the training split, we create a new training split by combining JRDB's train and test splits (approximately 90\% of the entire dataset) and train Faster-RCNN and RetinaNet on different fractions of this new training split. In this experiment, we use the original evaluation split as test split, approximately 10\% of the entire dataset. If JRDB's small number of sequences in the original training set is not enough to correctly cover the diversity of scenes in the dataset (and the original test set) causing the observed poor 2D detector performance, using more data from the test split should drastically improve the results. The results of this experiment are summarized in Fig~\ref{fig:2ddet_trainsetsize}. We observer an almost negligible improvement from using 50\% to 90\% of the entire dataset for training.  \revisedThree{While these results can be explained by several possible causes, \eg limited network capacity in evaluated models and/or partial mismatch between the distribution of the new training split (full training+test data) and the distribution of the selected validation set,
we believe this experiment shows strong evidence that the number of sequence in our originally suggested training set is large and diverse enough to cover most of the distribution and allow for generalization. } }

\begin{figure}[tb]
\centering
\begin{subfigure}[b]{0.49\linewidth}
    \centering
	\includegraphics[width=\linewidth]{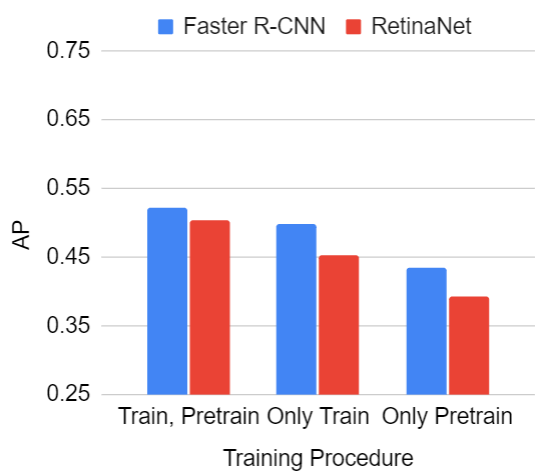}
	\caption{Average Precision (AP)} 
\end{subfigure}
\hfill
\begin{subfigure}[b]{0.49\linewidth}
    \centering
	\includegraphics[width=\linewidth]{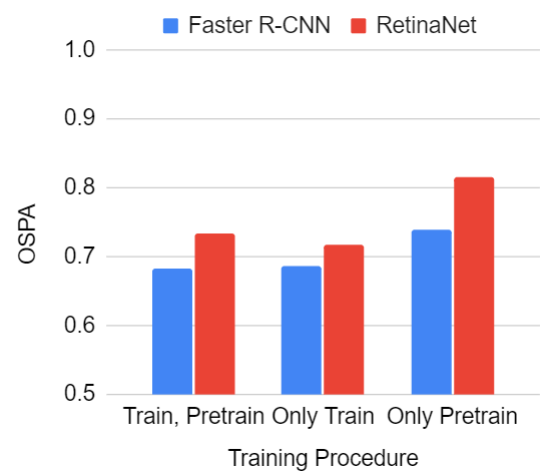}
	\caption{OSPA$_{IoU}$} 
\end{subfigure}
     \caption{\revised{Effect of training procedure in the JRDB 2D detection benchmark. We include average precision (AP) and OSPA$_{IoU}$ for 1) both pretraining on MS COCO and training on JRDB 2) Only training on JRDB 3) Only pretraining on MS COCO. Training on JRDB gives substantive improvement over pretrained weights.}}
    \label{fig:2ddet_trainprocedure}
 \end{figure}

\revisedTwo{The models in the previous experiment have been pretrained in a generic image datasets, MS COCO. To isolate the possible effect of this pretraining step, we compared the performance of i) a model pretrained in MS COCO and trained in JRDB original train split, ii) a model not pretrained and only training in JRDB, and iii) a model only pretrained in MS COCO. The results are summarized in Fig~\ref{fig:2ddet_trainprocedure}. Models only pretrained on MS COCO performed substantially worse than the models pretrained in MS COCO and further trained on JRDB. This indicates that training on JRDB provides a substantial benefit over generic pretraining, supporting the idea that 1) the data in JRDB is original and different to what is generically observed in image datasets, and 2) the original training set covers these special data distribution. Additionally, we observe that, as would be expected, if we do not pretrained the models on MS COCO, the performance drops. However, the drop between pretrained and not pretrained models is not significantly large, indicating that the proposed training set in JRDB is general and large enough to enable training of large detectors.}

\subsection{2D Person Tracking}
\label{ss_2dtrack}

We now analyze the performance of various popular and state-of-the-art methods for our 2D person tracking benchmark in JRDB: DeepSort and Tracktor. We also run our own baseline developed on the JRDB dataset, JRMOT. All these methods use the public detections generated by Faster-RCNN 2D to render fair the comparison of the tracking results.

\emph{DeepSort~\cite{DBLP:journals/corr/WojkeBP17}\footnote{\revised{Code: \url{https://github.com/nwojke/deep_sort}}}:} DeepSORT is a popular, online 2D tracking algorithm which follows the tracking-by-detection paradigm. It uses Kalman filtering for motion modelling, the Hungarian algorithm for the assignment problem, and a deep neural network to generate features of objects/pedestrians for re-identification and data association. \revised{In our experiments, we use the hyperparameters of the original publication with the exception of setting \texttt{nn\_budget} to 100 which limits the maximum size of appearance descriptors stored}.

\emph{JRMOT~\cite{shenoi2020jrmot}\footnote{\label{note2}\revised{Code: \url{https://github.com/StanfordVL/JRMOT_ROS}}}:} JRMOT is our provided baseline which performs both 2D and 3D tracking using the tracking-by-detection paradigm. JRMOT uses a multimodal (2D-3D) Kalman filter that updates the state based on measurements from both 2D RGB images and 3D point clouds. It applies m-best joint probabilistic data association (JPDA)~\cite{rezatofighiiccv2015} for probabilistic assignments, and 2D and 3D feature descriptor fusion for re-identification and data association. The feature descriptors are learned using the JRDB dataset. While it is primarily a 3D tracker, we project the resulting 3D tracks into 2D space to obtain 2D results that we can use to compare to the existing algorithms. \revised{We use the same set of hyperparameters described in the JRMOT manuscript.}

\emph{Tracktor~\cite{DBLP:journals/corr/abs-1903-05625}\footnote{\revised{Code: \url{https://github.com/phil-bergmann/tracking_wo_bnw}}}:} Tracktor is a state-of-the-art algorithm extensively used in the MOTChallenge dataset. This learning-based method uses an object detector that regresses the bounding boxes in consecutive frames and decides to preserve or terminate a track depending on the detector confidence scores. \revised{We use the provided implementation and approach for the detector and re-identification network and fine-tune it on JRDB. Specifically, the detector is fine-tuned for 27 epochs and the re-identification module for 25000 iterations, as prescribed by the authors.}

\subsubsection{Results \& Discussion}

Table~\ref{tab:2dtracking} summarizes the results of the three methods in the JRDB benchmark. \revised{According to the results, all these state-of-the-art online multi-object trackers perform poorly in our egocentric dataset, as indicated by low MOTA compared to the other tracking datasets and benchmarks and very high OSPA$^{(2)}_{IoU}$ (the values are very close to 1). 
All three methods are online trackers and use heuristics to handle occlusion or track termination. These heuristics have been proven to perform well for short occlusion periods in the tracks. However, JRDB contains considerably longer occluded trajectories compared to the other benchmark and datasets. We hypothesize this contributes to the low performance of these three state-of-the-art online trackers.} 

\revised{Comparing the trackers against each other, DeepSORT outperforms the two other trackers based on both MOT-Clear and OSPA$^{(2)}_{IoU}$ performance measures. Although JRMOT incorporate similar filtering strategy with more sensor information and has more reliable data-association technique compared to DeepSORT (\ie JPDA vs. Hungarian), the process of projecting 3D bounding boxes into the 2D image space to render JRMOT a 2D tracker introduces inaccuracies and leads to lower performance compared to Deepsort for 2D tracking. Rather surprisingly, we see worse performance for Tracktor, which is the state-of-the-art in the MOTChallenge benchmarks. However, Tracktor is a two-frame detector (based on the Faster R-CNN architecture) rather than a tracking algorithm. This might provide an advantage for this tracker in the benchmarks where the duration for object's occlusion is not very long and the performance improvement heavily depends on a better detection. However, due to its incapacity to maintain temporal information beyond two frames leads to this inferior performance in JRDB, caused by the unique features in the data (longer occlusions). As expected, Tracktor, a two frame detector, has the best localization error in OSPA$^{(2)}_{IoU}$ and similarly best MOTP value due to the best localization performance in MOT-clear.}

\emph{Indoor vs Outdoor scenes:} We see all methods perform better in indoor environments and have a similar drop in performance when running in outdoor environments. This indicates that outdoor environments are significantly more challenging. We hypothesize this is again caused by higher number of visible pedestrians and larger outdoor crowds at further away distances. However, neither in indoor nor in outdoor environments the evaluated solutions exhibit great performance, indicating the need to improve 2D tracking methods to support robot navigation in these type of environments underrepresented in existing datasets and challenges.

\revisedTwo{
\subsubsection{Re-identification after Long Occlusions}}

\revisedTwo{
Motivated by the prior analysis suggesting long occlusion windows are a driving force in tracking failure, specifically with ID switches, we propose a simple baseline method for re-identification after long occlusions (RALO). This method can be used with several of the existing state of the art baselines evaluated. Once tracks are terminated by a tracking solution, we propose to store them in a pool for an extended period of time, $T$. When new tracks are detected and initialized, we first compare them to the tracks in the pool. To that end, we use their appearance similarity measured based on the similarity of their AlignReID~\cite{zhang2017alignedreid} feature embeddings, and their similarity in position using GIoU~\cite{Rezatofighi_2018_CVPR}. If a new track obtains a sufficiently high similarity score (threshold adapted experimentally on the training data), we retrieve the matching track from the pool and the two are merge together.} \revisedF{This decreases ID switches while not affecting false negatives and false positives.}

\revisedTwo{As Table~\ref{tab:2docclreid} shows, we observe a significant decrease in ID switches thanks to our suggested baseline for re-identification after long occlusions. These results point out a necessary area where existing algorithms need to improve their performance in order to support robots in human crowded environments: handling correctly frequent and long occlusions. Methods that improve their performance under these conditions are likely to obtain better results in JRDB. For example, the integration of better forward and predictive models into tracking solutions and the exploitation of those to guide the detectors may improve further the re-identification and decrease further false negatives, the largest source of error for existing methods.
}

\begin{table}[h]
    \centering
    \begin{tabular}{|c||c|c|c||}
        \hline
         & DeepSORT & JRMOT (2D) & Tracktor \\ \hline
        Without RALO & 5296 & 7719 & 7026 \\ \hline
        With RALO & 5022 & 6612 & 6084 \\ \hline
    \end{tabular}
    \caption{\revisedTwo{Number of ID switches in 2D trackers with/without re-identification after long occlusions (RALO).} \revisedF{RALO leaves other MOTA components, \eg FP, FN, unaffected.}}
    \label{tab:2docclreid}
\end{table}
\vspace{-10pt}

\subsection{3D Person Detection}
\label{ss_3ddet}

\begin{table*}[tbh]
\revised{ \scriptsize
    \centering
\begin{tabular}{ |c||c| c |c||c| c |c c c||}
\hline
\multirow{2}{*}{\backslashbox{Method}{Metric}}& \multicolumn{3}{c||}{\%AP$_{IoU=.3}$$\uparrow$}& \multicolumn{5}{c||}{OSPA$_{IoU}$$\downarrow$}\\
\cline{2-9} 
 & Indoor & Outdoor & Overall & Indoor & Outdoor & Overall & Card. & Loc.  \\ \hline
 TANet & \bf52.16 & \bf 59.15&\bf 53.87 &\bf \revisedTwo{0.758}&\bf \revisedTwo{0.801}&\bf\revisedTwo{0.771}&\bf\revisedTwo{0.391}& \revisedTwo{0.380}\\ \hline
F-PointNet & 35.5 & 44.1 & 38.2&\revisedTwo{0.765}&\revisedTwo{0.844}& \revisedTwo{0.798}&\revisedTwo{0.564}&\bf\revisedTwo{0.234}\\ \hline
\end{tabular}
    \caption{3D detection results in the JRDB benchmark using average precision (AP) and OSPA$_{IoU}$. Card. $=$ Cardinality, Loc. $=$ Localization.}
    \label{tab:3ddetection}}
\end{table*}


\begin{figure}[tb]
\centering
\begin{subfigure}[b]{0.49\linewidth}
    \centering
	\includegraphics[width=\linewidth]{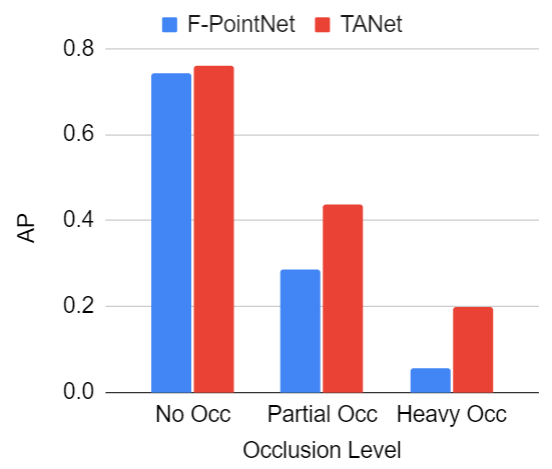}
	\caption{Average Precision (AP)} 
\end{subfigure}
\hfill
\begin{subfigure}[b]{0.49\linewidth}
    \centering
	\includegraphics[width=\linewidth]{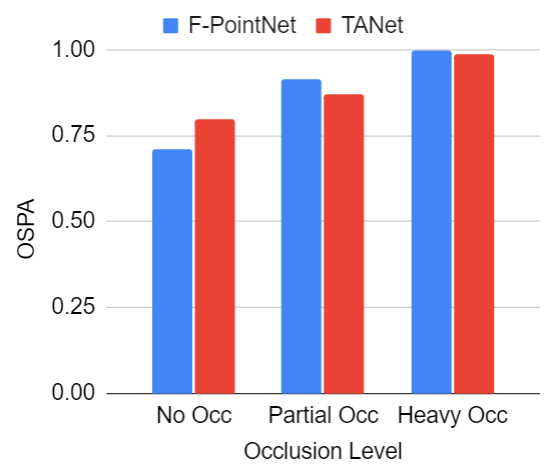}
	\caption{OSPA$_{IoU}$} 
\end{subfigure}
     \caption{\revised{Sensitivity to occlusion of methods in the JRDB 3D detection benchmark. We include average precision (AP) and OSPA$_{IoU}$ at 3 different occlusion (Occ.) levels. Both methods get significantly worse with occlusion, but TANet is slightly more robust to it.}}
    \label{fig:3ddetection_occ}
 \end{figure}

We now turn to evaluate 3d person detectors in the JRDB benchmark. We compare two state-of-the-art solutions from the community, Frustum PointNet and TANet, in our benchmark.

\emph{F-PointNet~\cite{fpointnet}\footnote{\revised{Code: \url{https://github.com/charlesq34/frustum-pointnets}}}:} Frustum PointNet is a conditional 3D detector that uses RGB data. It first uses a 2D detector in the RGB images to segment out the point cloud data inside the frustrum (rectangular pyramid) projected from the sensor through the corners of the bounding boxes. The segmented point cloud per detection is then fed into a PointNet neural network structure that predicts the oriented bounding box around the object of interest, pedestrians in our case. Our implementation uses the pretrained weights from the KITTI dataset~\cite{Geiger2012CVPR} and we fine-tune them on the JRDB dataset. We follow the same procedure as the original manuscript to prepare the JRDB dataset. Specifically, we generate training data by extracting the frustum of perturbed ground-truth 2D bounding boxes. We filter the ones that are too far away or has no enclosed points in its corresponding 3D bounding box. Then, we train the model on the processed JRDB dataset for 200 epochs using the default parameters.

\revised{\emph{TANet~\cite{liu2019tanet}\footnote{\revised{ Code: \url{https://github.com/happinesslz/TANet}}}:} TANet is a 3D detector that only uses point cloud data without RGB information. It relies on voxelizing the point cloud and applying a triple attention mechanism to obtain a feature representation for each voxel. These voxel feature representations are then arranged according to their original layout and fed into a regression module to produce 3D bounding boxes around objects of interest. We train TANet for pedestrian detection on JRDB, filtering boxes that are too far away or have less than 10 enclosed points, and train for 500,000 steps using the default parameters in the original manuscript.}

\subsubsection{Results \& Discussion}

\begin{figure}[tb]
\centering
\begin{subfigure}[b]{0.49\linewidth}
    \centering
	\includegraphics[width=\linewidth]{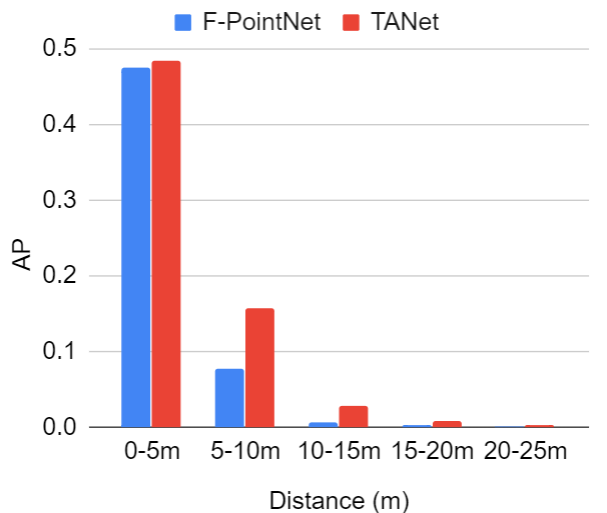}
	\caption{Average Precision (AP)} 
\end{subfigure}
\hfill
\begin{subfigure}[b]{0.49\linewidth}
    \centering
	\includegraphics[width=\linewidth]{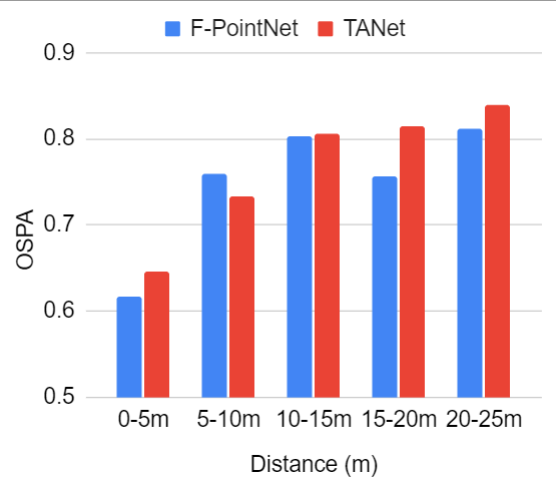}
	\caption{OSPA$_{IoU}$} 
\end{subfigure}
 	\caption{\revised{Results in the JRDB 3D detection benchmark by distance. We report both performance measures, AP (higher better) and OSPA$_{IoU}$ (lower better), at intervals of \SI{5}{\meter} from \SI{0}{\meter} to \SI{25}{\meter}. Both methods, F-PointNet and TANet, show a considerable drop in performance with increasing distance to the sensor, with a slightly more robust performance by TANet. Beyond \SI{15}{\meter} the results from both detectors are extremely noisy.}}
 	\label{fig:3ddetectiondistance}
 \end{figure}

We present overall results for the 3D detection methods in Table~\ref{tab:3ddetection}. \revised{Similar to 2D cases, these state-of-the-art 3D detectors perform poorly in our dataset in both AP score as well OSPA$^{(2)}_{IoU}$. The high value of cardinality error in OSPA$^{(2)}_{IoU}$ is an indicator of many missing detections in both detectors. The challenges posed by the JRDB unique data such as high level of partial and full occlusions due to very crowded scene and indoor natural occlusions, high sparsity of point clouds for smaller size object like human (compared to vehicles), high distance range of pedestrians from the senors and pedestrians above (or below) the ground plane due to stairs, cause these state-of-the art 3D detectors to perform rather poorly in the JRDB benchmark while performing well on the self-driving benchmarks datasets such as KITTI or Waymo.}

\revised{Compared to TANet, F-PointNet performs very poorly in our dataset. F-PointNet assumes only one object of interest is contained in the 3D data of each frustrum but this may not hold true in JRDB, especially in crowded scenes. Moreover, both detectors perform weakly in the estimation of person orientation which contributes to the low values in both performance metrics. To examine this hypothesis, we conducted additional experiments increasing the value of IoU threshold for AP from $0.3$ to $0.5$. The results of both F-PointNet and TANet noticeably dropped from $38.21$ and $53.87$ to $6.38$ and $4.17$ respectively. Since the value of $IoU$ is very sensitive to an error in orientation, a significant drop in AP with a small decrease in the value of IoU can be a reflective of their good localization but poor orientation estimation. Based on these results, where TANet exhibited overall better performance than F-PointNet, we decided to use TANet as a public detector for our 3D tracking benchmark.}

\emph{Indoor vs Outdoor Scenes:}
Considering AP (with a low $IoU$ threshold, \ie $0.3$ ) as performance measure, we surprisingly see a better performance for the 3D detectors in the outdoor scenario. We hypothesize that people cluster more tightly in indoor environments, and indoor scene includes significantly more occluders in the point cloud data, rendering 3D detection in indoor environment harder. This situation become worse for F-PointNet, which assumes one object per frustum and thus does not perform well in scenes with crowds or partial occlusions. However, considering a threshold-free metric like OSPA$_{IoU}$, both indoor and outdoor scenes seem to be almost equally challenging for these detectors.

\begin{table*}[tb]
\revised{\scriptsize
    \centering
\begin{tabular}{ |l||c|c|c|c|c|c|c||c|c|c c c||}
\hline
\multirow{3}{*}{\backslashbox[12em]{Method}{Metric}} & \multicolumn{7}{c||}{Clear-MOT$_{IoU=.3}$}&\multicolumn{5}{c||}{OSPA$^{(2)}_{IoU}$$\downarrow$}\\
\cline{2-8}
 & \multicolumn{3}{c|}{\%MOTA$\uparrow$} & \multirow{2}{*}{\%MOTP$\uparrow$}  & \multirow{2}{*}{$|\mathcal{B}_{ID}|\downarrow$} & \multirow{2}{*}{$|\mathcal{B}_{FP}|\downarrow$} & 
\multirow{2}{*}{$|\mathcal{B}_{FN}|\downarrow$} & \multicolumn{5}{c||}{}\\ \cline{2-4}\cline{9-13} 
 & Indoor & Outdoor & Overall&&&&&Indoor & Outdoor & Overall & Card. & Loc. 
 \\\hline
JRMOT & 20.9 &  18.6 & 20.2 & \bf 42.5 &  \bf 4207 & 19711 & 765907 &\bf0.976&\bf0.979&\bf0.979&\bf0.644&0.335 \\\hline
AB3DMOT (F-PointNet) & 19.9 & 18.3 & 19.4 & 42.0 & 6177 & \bf 13664 & 777946 &\bf0.976&0.984&0.983&0.756&0.227\\ \hline
AB3DMOT (TANet) & \bf 25.7 & \bf 27.1 & \bf 26.2 & 38.0 & 10081 & 61417 & \bf 658986 &0.983&0.989&0.987&0.782&\bf0.206\\ \hline
\end{tabular}
\caption{Results in the JRDB 3D tracking benchmark. We include Clear-MOT metrics and OSPA$_{IoU}^{(2)}$. IDS $=$ ID Switches, FP $=$ False Positives, FN $=$ False Negatives, Card. $=$ Cardinality, Loc. $=$ Localization.}
    \label{tab:3dtracking}}
\end{table*}

\emph{Performance against Occlusion Level:} As expected, we see \revised{a similarly} rapid decay in performance with increasing level of occlusion \revised{in both AP and OSPA$^{(2)}_{IoU}$ performance measures}. In particular, when there is no occlusion, the performance is far better than the overall AP and OSPA$^{(2)}_{IoU}$, but even under small occlusions, the performance rapidly decays. This shows that our JRDB dataset presents challenging scenes for existing 3D detectors, and that novel research needs to be carried out in this type of scenarios to develop new detectors that are more robust to the frequent levels of occlusions observed by human-sized robots in public human spaces. 

\emph{Performance Against Distance from the Robot:} As Figure~\ref{fig:3ddetectiondistance} shows, the performance of the 3D detectors quickly drops off as we go farther out from the robot. \revised{F-PointNet presents slightly better OSPA$^{(2)}_{IoU}$ than TANet for detections at far distances. A possible reason is the better cardinality score of F-PointNet, although both algorithms present poor localization capabilities. }At the range between \SI{5}{\meter} to \SI{10}{\meter}, the performance has already dropped below $0.1$ AP and the detections provided from both algorithms are extremely noisy. This indicates a clear opportunity for the development of new 3D detectors that can provide reliable hypotheses even with few points for further away objects.

\revisedTwo{\emph{Effect of the Size of JRDB on the Results:} Similar to the evaluation of 2D Person Detection, we would like to rule out the effect of the size of JRDB's training set in the poor performance observed in the evaluated state-of-the-art 3D detectors. Therefore, we perform an analogous analysis, training F-PointNet on different fractions of the combination of JRDB's original training and test sets, and testing on JRDB's original evaluation set. Training on approximately 50\% of the dataset, F-PointNet reaches $46.07$ AP and 0.766 OSPA$^{(2)}_{IoU}$. In comparison, using 90\% of the dataset for training F-PointNet leads to $48.67$ AP and 0.751 OSPA$^{(2)}_{IoU}$, a negligible difference. F-PointNet was not pretrained on any large generic dataset and therefore, we do not need to evaluate the effect of pretraining on these results. The results support the idea that the original training set split is large and diverse enough to cover the distribution of the test set, and its size is not the cause of the observed performance of 3D detectors, but rather the challenging special type of data included in JRDB.}

\subsection{3D Person Tracking}
\label{ss_3dtrack}

Finally, we evaluate 3D person tracking solutions in our new JRDB benchmark. We compare the performance of two state-of-the-art tracking solutions with available open-source implementations, AB3DMOT and JRMOT. This problem is significantly less studied than 2D tracking, and, while several works have been tackling 3D tracking in recent years, most of them come from autonomous vehicle companies that do not release open-source code. To assure a fair comparison between both 3D trackers, both methods use the same F-PointNet 3D detections. F-PointNet 3D detections are conditioned on 2D detections; we use the Faster R-CNN 2D detections provided with the dataset to query F-PointNet for 3D detections. \revised{For AB3MOT, we also present results using TANet 3D detections. However, JRMOT results are only obtained with F-PointNet as that specific detection method is required for the system to obtain a one-to-one mapping between 2D and 3D detections.}

\emph{AB3DMOT~\cite{DBLP:journals/corr/abs-1907-03961}\footnote{\revised{Code: \url{https://github.com/xinshuoweng/AB3DMOT}}}:} AB3DMOT is a light weight, online 3D tracker. \revised{This approach uses a simple 3D Kalman filter and the Hungarian algorithm for matching and data association. Similarity scores for the Hungarian algorithm are based on intersection over union of 3D boxes.} We use the publicly available implementation.

\emph{JRMOT~\cite{shenoi2020jrmot}\footnoteref{note2}:} JRMOT is our provided baseline, as introduced in Section~\ref{ss_2dtrack}. JRMOT is based internally in a F-PointNet like procedure that provides linked 2D and 3D detections to use in a multimodal Kalman update step. Therefore, we use JRMOT with Faster R-CNN 2D and F-PointNet associated 3D detections. Different to the version of Section~\ref{ss_2dtrack}, to benchmark 3D tracking on the JRDB benchmark, we use the full 3D tracking capabilities of JRMOT. We use the exact system as described in the JRMOT publication~\cite{shenoi2020jrmot}.

\subsubsection{Results \& Discussion}
\begin{figure}[tb]
\centering
\begin{subfigure}[b]{0.49\linewidth}
    \centering
	\includegraphics[width=\linewidth]{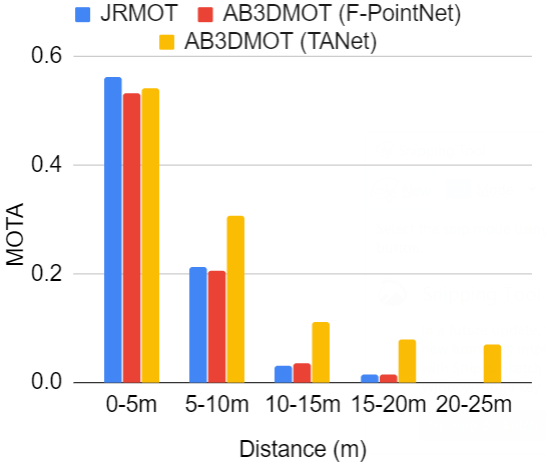}
	\caption{MOTA Results} 
\end{subfigure}
\hfill
\begin{subfigure}[b]{0.49\linewidth}
    \centering
	\includegraphics[width=\linewidth]{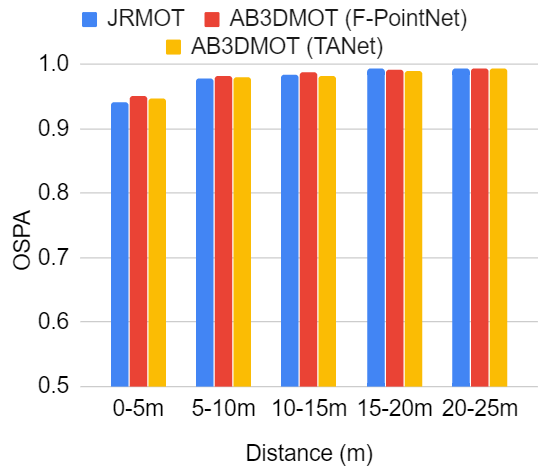}
	\caption{OSPA Results} 
\end{subfigure}
 	\caption{\revised{Results in the JRDB 3D tracking benchmark by distance. We report both performance measures, MOTA (higher better) and OSPA$^{(2)}_{IoU}$ (lower better), at intervals of \SI{5}{\meter} from \SI{0}{\meter} to \SI{25}{\meter}. JMOT and the both variants of AB3DMOT, with F-PointNet and TANet detections, perform best at closer distances, under \SI{10}{\meter}. Beyond \SI{15}{\meter} the tracking results are extremely noisy. AB3DMOT with TANet shows significantly better MOTA values than the other baselines at larger distances due to the less noisy input detections.}}
 	\label{fig:3dtrackingdistance}
 \end{figure}

In Table~\ref{tab:3dtracking}, we present the overall results of both tracking methods \revised{for both Clear-MOT and OSPA$_{IoU}^{(2)}$ performance measures}. \revised{ We see low overall scores in MOTA and and very high overall errors OSPA$_{IoU}^{(2)}$ for all methods primarily driven by the high rate of false negatives, indicating the challenging nature of this dataset.} \revised{Based on the OSPA$_{IoU}^{(2)}$ measure we observe that AB3DMOT using TANet detections presents worse localization than AB3DMOT with F-PointNet detections. This is surprising, since TANet presented better localization than F-PointNet in the 3D detection JRDB benchmark, based on OSPA$_{IoU}^{(2)}$. This indicates that AB3DMOT is the culprit of this worse localization and that there is room for better tracking based on the given detections. We hypothesize this is due to poor and unstable Kalman filtering on extremely noisy angles from the detector which ultimately make performance worse.}

\emph{Indoor vs Outdoor Scenes:} We see a small gain in \revised{both MOTA and OSPA$_{IoU}^{(2)}$ in indoor scenes compared to outdoor scenes. This is consistent across all methods. Similar to the observed performance in the JRDB benchmark for 2D tracking, we hypothesize this is the effect of challenging tracking cases in the JRDB: the existing algorithms cannot cope with the longer occlusion patterns in more crowded outdoor scenes observed in our novel dataset (see Fig.\ref{fig:2D_occlusion_stats}). Additionally, the more crowded scenes lead to poor detection performance (see Section~\ref{ss_3ddet}), which affects directly the performance of the 3D trackers.}

\emph{Performance Against Distance from the Robot:} In Fig.~\ref{fig:3dtrackingdistance}, we observe a rapid and remarkable degradation in performances across both methods as distance increases, indicating existing methods are still not very effective beyond \SI{10}{\meter}. \revised{We also observe that JRMOT performs slightly better than both variants of AB3DMOT at shorter distances, in the range between \SI{0}{\meter} to \SI{5}{\meter}, based on both MOTA and OSPA$_{IoU}^{(2)}$.} As detections and measurements are noisier at these distances (see Section~\ref{ss_3ddet}), the low quality of the detections is the most probable reason for the low performance of the trackers. Interestingly, we observe a significant difference in MOTP at larger distances between both approaches. We hypothesize that, by using information from the 2D bounding boxes, JRMOT obtains a more accurate estimation of the 3D bounding boxes and better fine-grained estimation of the object position, which leads to the observed higher MOTP. \revised{These limited performance of state-of-the-art solutions in our JRDB 3D tracking benchmark provide insights about future research areas to explore by the community. We believe that the low performance of the 3D trackers, both at large but also at short distances, is a handicap to develop autonomous navigating robots in human public spaces.} 

\revisedTwo{
\subsubsection{Re-identification through Long Occlusions}
To alleviate some of the special difficulties in JRDB, namely, the more frequent and temporally extended occlusions, we adapt and integrate into the 3D trackers the re-identification method introduced in Section 7.2.2.
As Table~\ref{tab:3docclreid} depicts, we observe a substantial decrease in the number of ID switches, indicating that an extended memory for tracks is a simple but effective way to reduce the effects of long occlusions. Future research in 3D tracking could dive deeper into extended memory methods to improve further the tracking capabilities in the presence of large occlusions. We plan to further improve tracking performance under long occlusions by incorporating more accurate predictive models based on egocentric motion and social cues~\cite{gupta2018social,kosaraju2019social,swofford2020improving}.
}

\begin{table}[h]
    \centering
    \begin{tabular}{|c||c|c||}
        \hline
         & JRMOT (3D) & Ab3DMOT \\ \hline
        Without RALO & 4207 & 8278 \\ \hline
        With RALO & 2673 & 3868 \\ \hline
    \end{tabular}
    \caption{\revisedTwo{Number of ID switches in 3D trackers with/without re-identification after long occlusions (RALO). \revisedF{RALO leaves other MOTA components, \eg FP, FN, unaffected.}}}
    \label{tab:3docclreid}
\end{table}
\vspace{-10pt}



\section{Conclusion and Future Work}
\label{s:conc}

We presented JRDB, a novel dataset containing multimodal streams acquired in human environments: university indoor buildings and pedestrian areas on campus. The dataset of temporally synchronized and calibrated data includes images from 360\degree stereo cylindrical cameras, LiDAR 3D point clouds, audio signals, IMU values and encoder readings from the robot's base. The data includes scenes where the robot navigates among humans. The dataset has been annotated with ground truth 2D bounding boxes and associated 3D cuboids around all persons in the scenes. We show that it is a challenging dataset for person detection and training, in both 2D and 3D for existing methods. Hence, we expect this dataset to support research in perception for autonomous agents in human and social contexts. Our future plans are to continue annotating ground truth values for individual and group activities, social grouping, and human posture. 

\section*{Acknowledgment}
This work was supported by the ONR, MURI award W911NF-15-1-0479.


%

\ifCLASSOPTIONcaptionsoff
  \newpage
\fi



\bibliographystyle{IEEEtran}
\bibliography{jrdb}
%
%
\begin{IEEEbiography}[{\includegraphics[width=1in,height=1.25in,clip,keepaspectratio]{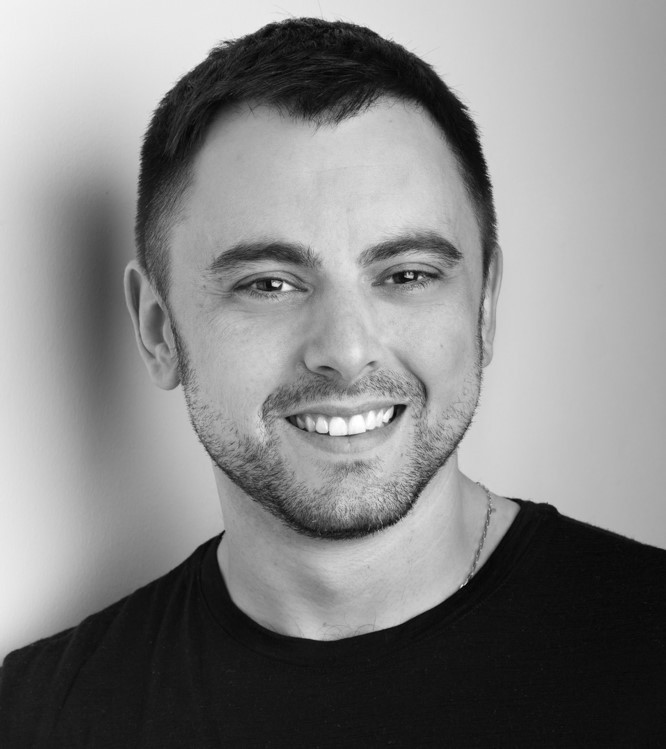}}]{Roberto Mart\'in-Mart\'in}
Roberto Mart\'in-Mart\'in is a postdoctoral scholar at the Stanford Vision and Learning Lab with Prof. Silvio Savarese and Prof. Fei-Fei Li. He coordinates research projects in two groups: the JackRabbot team, which works on mobile manipulation in human environments, and the People, AI \& Robots (PAIR) team, which works on visuo-motor learning skills for manipulation and planning. He obtained his PhD in Robotics at the Technische Universit\"at Berlin at the RBO group from Prof. Oliver Brock.
\end{IEEEbiography}
\vfill
\begin{IEEEbiography}[{\includegraphics[width=1in,height=1.25in,clip,keepaspectratio]{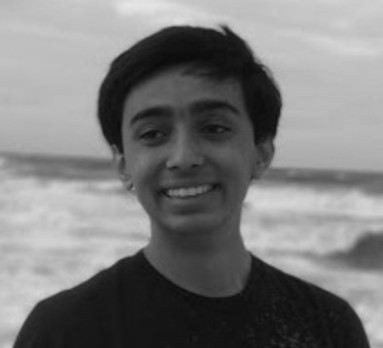}}]{Mihir Patel}
Mihir Patel is pursuing a Bachelor's degree in the Mathematics department and a simultaneous Master's degree in the Computer Science department at Stanford University. He currently works at the Stanford Vision and Learning Lab with Prof. Silvio Savarese and Prof. Fei-Fei Li and is part of the JackRabbot team. His interests are in vision and generative modeling. 
\end{IEEEbiography}
\vfill
\begin{IEEEbiography}[{\includegraphics[width=1in,height=1.25in,clip,keepaspectratio]{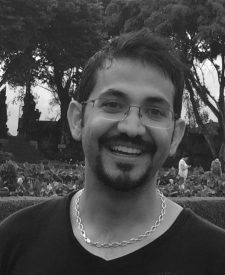}}]{Hamid Rezatofighi}
Hamid Rezatofighi is a lecturer at Faculty of Information Technology, Monash University, Australia. Before that, he was an Endeavour Research Fellow at the Stanford Vision Lab (SVL), Stanford University and a Senior Research Fellow at the Australian Institute for Machine Learning (AIML), the University of Adelaide. His main research interest focuses on computer vision and vision-based perception for robotics, including object detection, multi-object  tracking, human trajectory and pose forecasting and human collective activity recognition.  He has also research expertise in Bayesian filtering, estimation and learning using point process and finite set statistics.
\end{IEEEbiography}
\vfill
\begin{IEEEbiography}[{\includegraphics[width=1in,height=1.25in,clip,keepaspectratio]{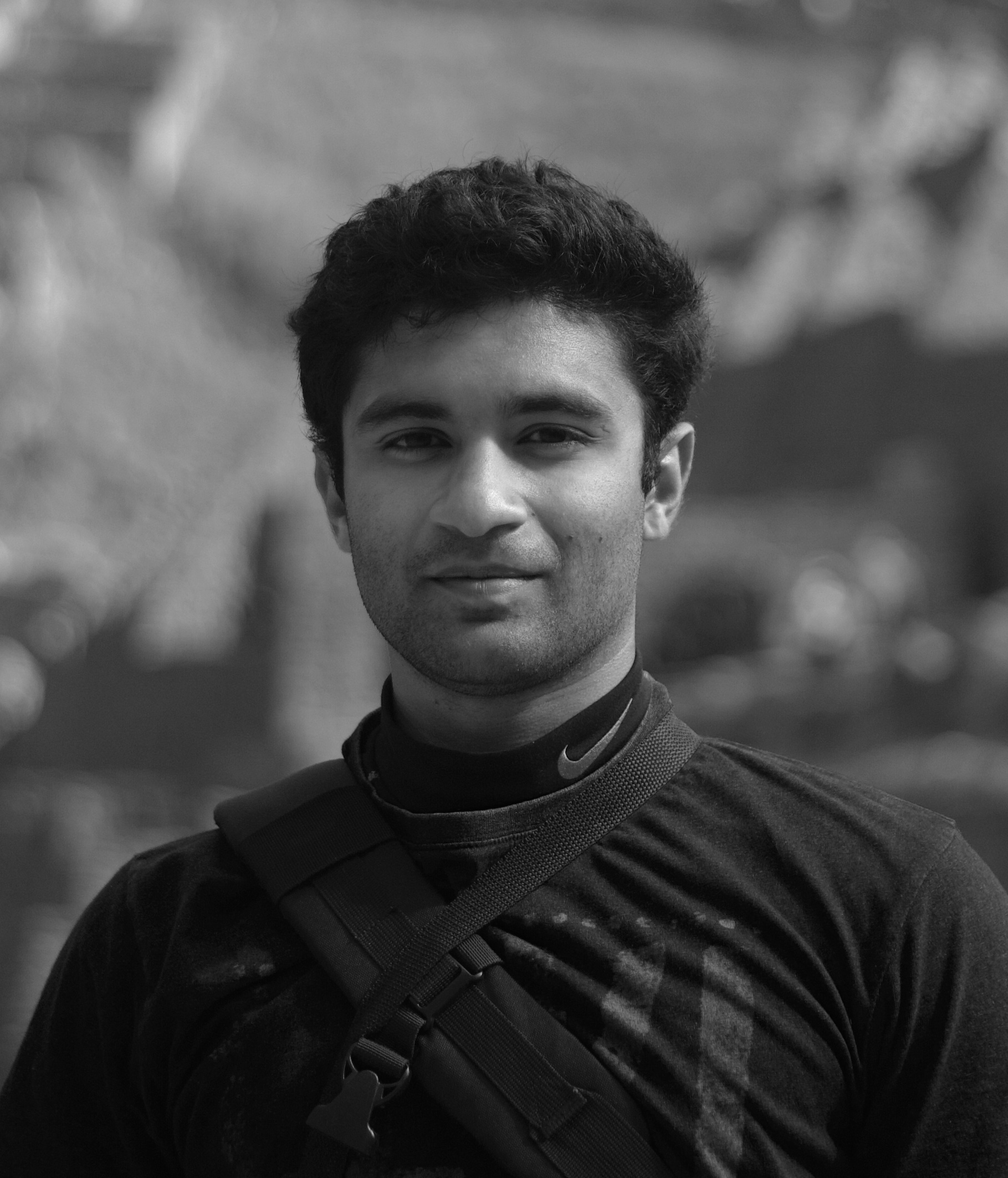}}]{Abhijeet Shenoi}
Abhijeet Shenoi is an algorithm engineer at AiBee US Corporation. He earned his Master's degree in the Computer Science department at Stanford University in 2019, while working under the supervision of Prof. Silvio Savarese in the Stanford Vision and Learning Lab. His research interests include semantic understanding, 3D tracking and 3D computer vision.
\end{IEEEbiography}
\vfill
\begin{IEEEbiography}[{\includegraphics[width=1in,height=1.25in,clip,keepaspectratio]{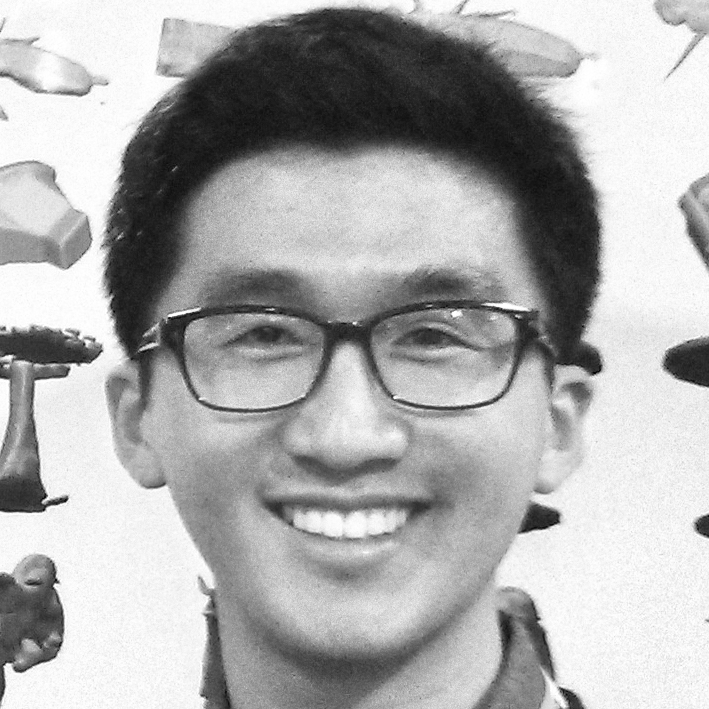}}]{JunYoung Gwak}
JunYoung Gwak is a PhD student at the Stanford Vision and Learning Lab with Prof. Silvio Savarese. His current research focus is on 3D vision and scene understanding.
\end{IEEEbiography}
\begin{IEEEbiography}[{\includegraphics[width=1in,height=1.25in,clip,keepaspectratio]{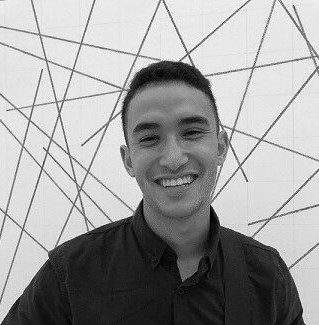}}]{Eric Frankel}
Eric Frankel is a research assistant in the JackRabbot Laboratory under Prof. Silvio Savarese on the JackRabbot project, which studies and improves autonomous agent activity in dynamic environments. He is studying mathematics and computer science at Stanford University under Prof. Emmanuel Cand\`es. His research interests include computer vision and representation learning.
\end{IEEEbiography}
\vfill
\begin{IEEEbiography}[{\includegraphics[width=1in,height=1.25in,clip,keepaspectratio]{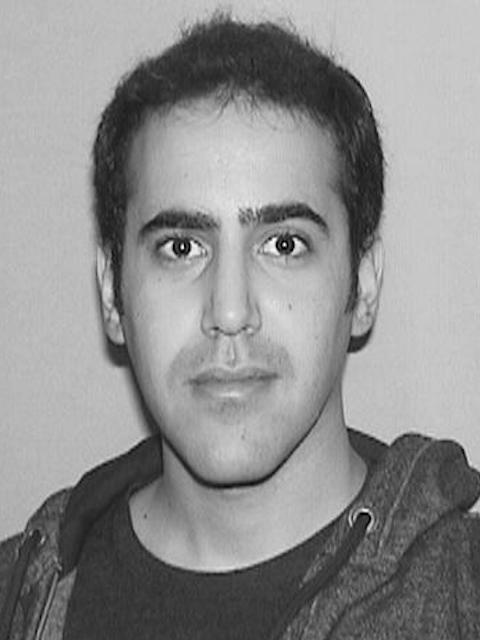}}]{Amir Sadeghian}
Amir Sadeghian is an Algorithm Scientist and founding member at Aibee US Corporation, where he leads the shopping mall team. His team works on building systems to digitalize the shopping malls, enabling better management and opening new ways to monetize off-line traffic.
He received his PhD from Stanford University in Jan 2019, where he worked in Stanford Vision and Learning Lab with Prof. Silvio Savarese. His research interests primarily focus on computer vision and perception for robotics. During his PhD, he was leading the JackRabbot team that has been featured in several major news outlets including CBS, ABC, and BBC.
\end{IEEEbiography}
\vfill
\begin{IEEEbiography}[{\includegraphics[width=1in,height=1.25in,clip,keepaspectratio]{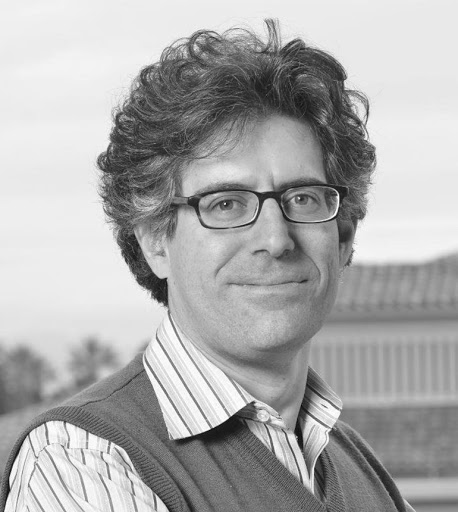}}]{Silvio Savarese}
Silvio Savarese is an Associate Professor of Computer Science at Stanford University and the inaugural Mindtree Faculty Scholar. He earned his Ph.D. in Electrical Engineering from the California Institute of Technology in 2005 and was a Beckman Institute Fellow at the University of Illinois at Urbana-Champaign from 2005–2008. He joined Stanford in 2013 after being Assistant and then Associate Professor of Electrical and Computer Engineering at the University of Michigan, Ann Arbor, from 2008 to 2013. His research interests include computer vision, robotic perception and machine learning. He is recipient of several awards including a Best Student Paper Award at CVPR 2016, the James R. Croes Medal in 2013, a TRW Automotive Endowed Research Award in 2012, an NSF Career Award in 2011 and Google Research Award in 2010. In 2002 he was awarded the Walker von Brimer Award for outstanding research initiative.
\end{IEEEbiography}


\vfill

\enlargethispage{-5in}



\renewcommand{\thesection}{\Alph{section}}
\renewcommand{\thesubsection}{A.\arabic{subsection}}

\setcounter{figure}{0} \renewcommand{\thefigure}{A.\arabic{figure}}

\setcounter{table}{0} \renewcommand{\thetable}{A.\arabic{table}}

\section{Appendix}

\vspace{1cm}
\subsection{360 Cylindrical Image Composition}
\label{sec:stitching}

The raw data in our dataset is obtained from two parallel rows of five RGB cameras each. The synchronized images from each row can be used to compose a 360\degree image, as shown in our visualizations (\eg Fig.~2 and Fig.~\ref{fig:scenesvis}). We assume that the individual cameras of each row (5 in total) can be modeled as pinhole cameras, each providing images of resolution $752 \times 480$ pixels. To compose a single 360\degree cylindrical RGB image from them, we adopt the following procedure. 

First, we construct the function that will map pixels in the cylindrical image to pixels in the five individual images, $c_i$. To do so, we assume that every pixel in the cylindrical image, $(u_{\mathit{cyl}},v_{\mathit{cyl}})$, is at a constant distance $r$ from the camera center. Assuming a Cartesian frame with the axis $y$ pointing upwards, $x$ pointing to the front and $z$ pointing to the right, the 3D location of the pixel in the image cylinder would have the coordinates $(r\cos{\theta}, v_{\mathit{cyl}}, r\sin{\theta})$, where $\theta$ is the angle between the line connecting the origin and the pixel, and the positive $x$ axis of the reference frame. We then project this pixel onto each of the five individual cameras based on their extrinsic parameters $R_i$ and $T_i$, intrinsic parameters, $K_i$, and distortion coefficients, $D_i$. The projected point $(\hat{u}_i, \hat{v}_i)$ is given by:
$$ \begin{pmatrix}
\hat{u}_i \\
\hat{v}_i
\end{pmatrix} = K_i f_{D_i}\left(\begin{bmatrix} R_i & T_i \end{bmatrix} \begin{pmatrix}
r \cos{\theta} \\
v_{\mathit{cyl}} \\
r\sin{\theta}\\
1
\end{pmatrix}\right)$$
where $f_D$ is the distortion function parameterized by the distortion coefficients, $D$. To eliminate cases where the point is behind the camera onto which we are projecting, we first transform the 3D location of the pixel in the cylindrical image to each of the camera reference frames $$\begin{pmatrix}
x_i \\
y_i \\
z_i
\end{pmatrix} = \begin{bmatrix} R_i & T_i \end{bmatrix} \begin{pmatrix}
r \cos{\theta} \\
v_{\mathit{cyl}} \\
r\sin{\theta}\\
1
\end{pmatrix}$$ 

Since the pointing direction of the camera is aligned with the positive $z$ axis, $z_i < 0$ indicates that the 3D location of the pixel in the cylindrical image corresponds to a point behind camera $c_i$ and we do not consider its projection. We also eliminate correspondences beyond the limits of the individual images, i.e. when $\hat{u}_i$ or $\hat{v}_i$ lie outside the boundaries of the image from $c_i$. The result of this process is the mapping between $(u_{\mathit{cyl}},v_{\mathit{cyl}})$ and $(\hat{u}_i, \hat{v}_i)$ for each of the individual cameras, $c_i$. 

Given these mappings, we can transform the pixels from the original five images into the corresponding pixels in the cylindrical image, followed by smoothing of the individual images onto a cylindrical surface, resulting in the 360\degree RGB cylindrical image. 


\newpage

\subsection{JRDB Data Structure}
\begin{figure}[th!]
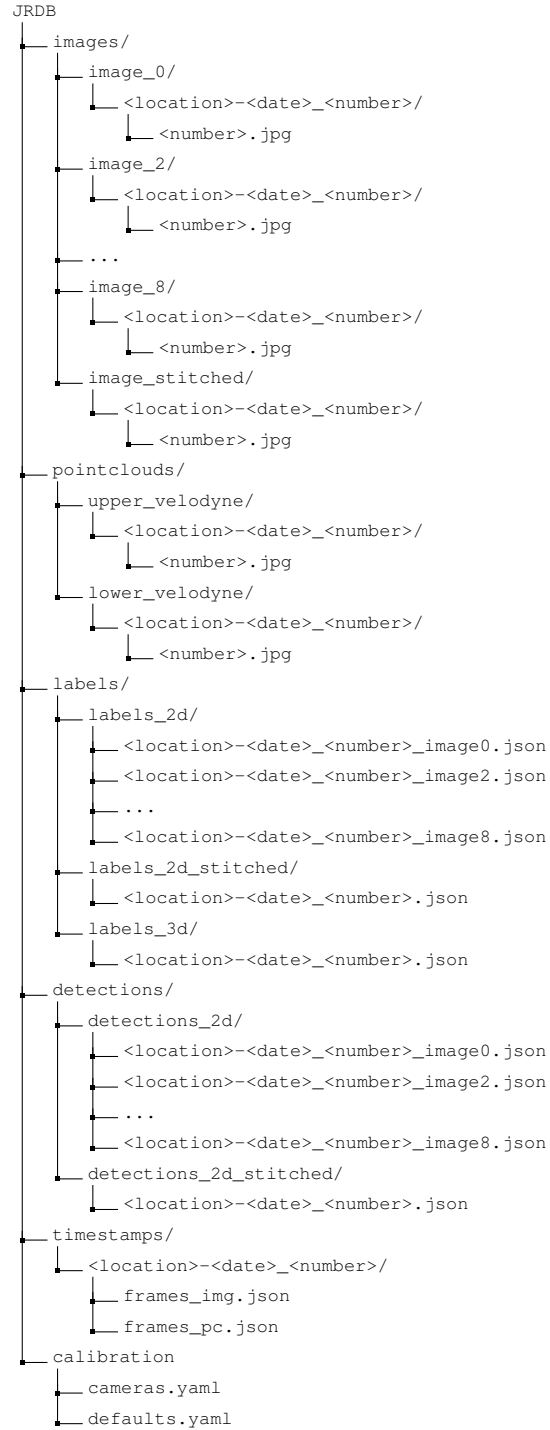

\scriptsize
\dirtree{%
.1 JRDB.
.2 images/.
.3 image\_0/.
.4 <location>-<date>\_<number>/.
.5 <number>.jpg.
.3 image\_2/.
.4 <location>-<date>\_<number>/.
.5 <number>.jpg.
.3 \ldots.
.3 image\_8/.
.4 <location>-<date>\_<number>/.
.5 <number>.jpg.
.3 image\_stitched/.
.4 <location>-<date>\_<number>/.
.5 <number>.jpg.
.2 pointclouds/.
.3 upper\_velodyne/.
.4 <location>-<date>\_<number>/.
.5 <number>.jpg.
.3 lower\_velodyne/.
.4 <location>-<date>\_<number>/.
.5 <number>.jpg.
.2 labels/.
.3 labels\_2d/.
.4 <location>-<date>\_<number>\_image0.json.
.4 <location>-<date>\_<number>\_image2.json.
.4 \ldots.
.4 <location>-<date>\_<number>\_image8.json.
.3 labels\_2d\_stitched/.
.4 <location>-<date>\_<number>.json.
.3 labels\_3d/.
.4 <location>-<date>\_<number>.json.
.2 detections/.
.3 detections\_2d/.
.4 <location>-<date>\_<number>\_image0.json.
.4 <location>-<date>\_<number>\_image2.json.
.4 \ldots.
.4 <location>-<date>\_<number>\_image8.json.
.3 detections\_2d\_stitched/.
.4 <location>-<date>\_<number>.json.
.2 timestamps/.
.3 <location>-<date>\_<number>/.
.4 frames\_img.json.
.4 frames\_pc.json.
.2 calibration.
.3 cameras.yaml.
.3 defaults.yaml.
}
\caption{The dataset contains image and point cloud data, along with labels, sample detections, timestamps and calibration information. Further detail is in text.}
\label{fig:dirtree}
\end{figure}

\newpage
\subsection{Validation Split}
\begin{figure}[!hb]
    \centering
\begin{subfigure}[b]{0.95\linewidth}
   \begin{subfigure}[b]{0.48\linewidth}
    \centering
	\includegraphics[width=1.05\linewidth]{figs/area_distribution_train_2.pdf}
	\caption{Train Set} 
\end{subfigure}
\hfill
\begin{subfigure}[b]{0.48\linewidth}
    \centering
	\includegraphics[width=1.05\linewidth]{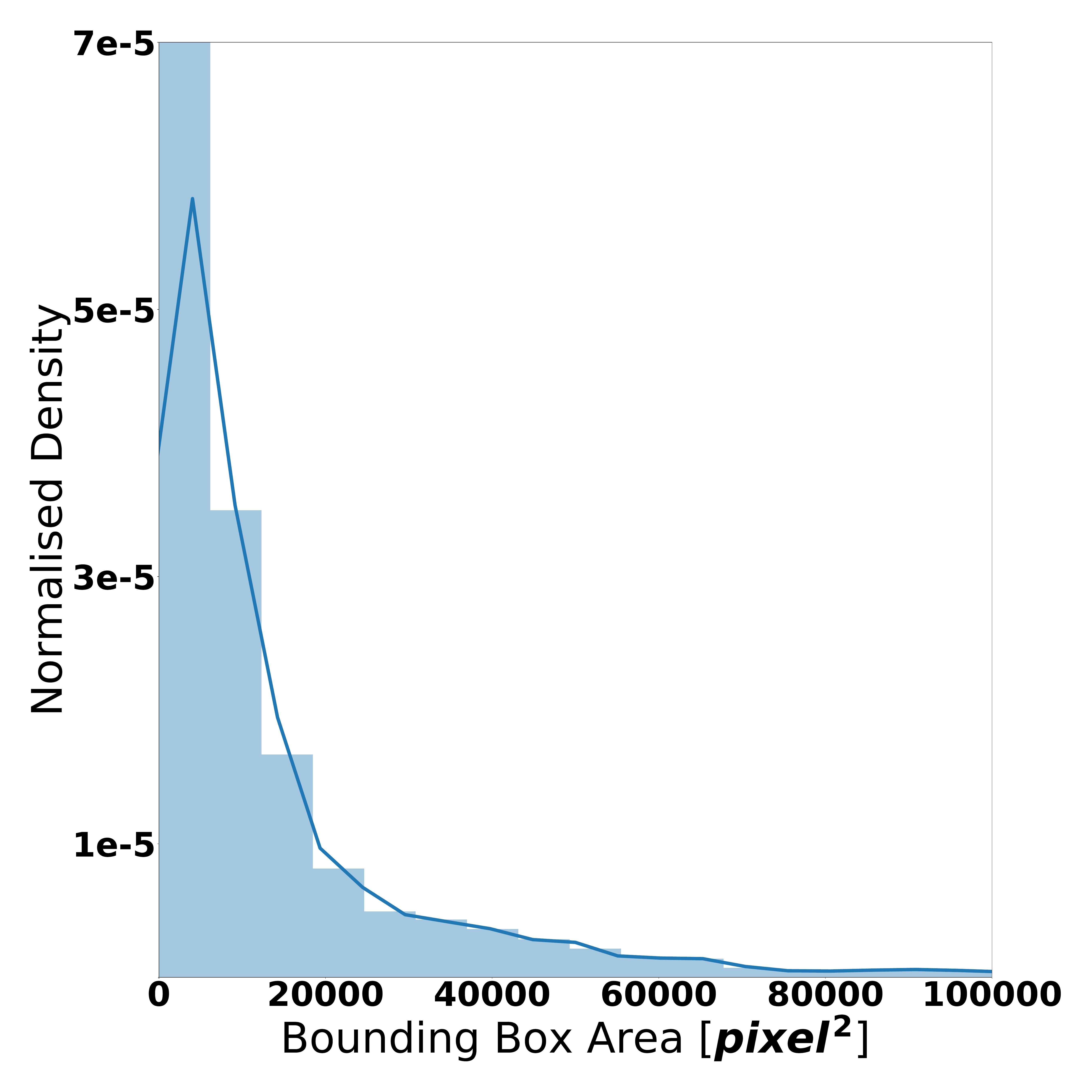}
	\caption{\revised{Validation Set}}
\end{subfigure}
    \caption*{Histogram of areas of 2D bounding boxes}
\end{subfigure}
\begin{subfigure}[b]{0.95\linewidth}
\begin{subfigure}[b]{0.48\linewidth}
    \centering
	\includegraphics[width=1.05\linewidth]{figs/distance_distribution_train_2.pdf}
	\caption{Train Set} 
\end{subfigure}
\hfill
\begin{subfigure}[b]{0.48\linewidth}
    \centering
	\includegraphics[width=1.05\linewidth]{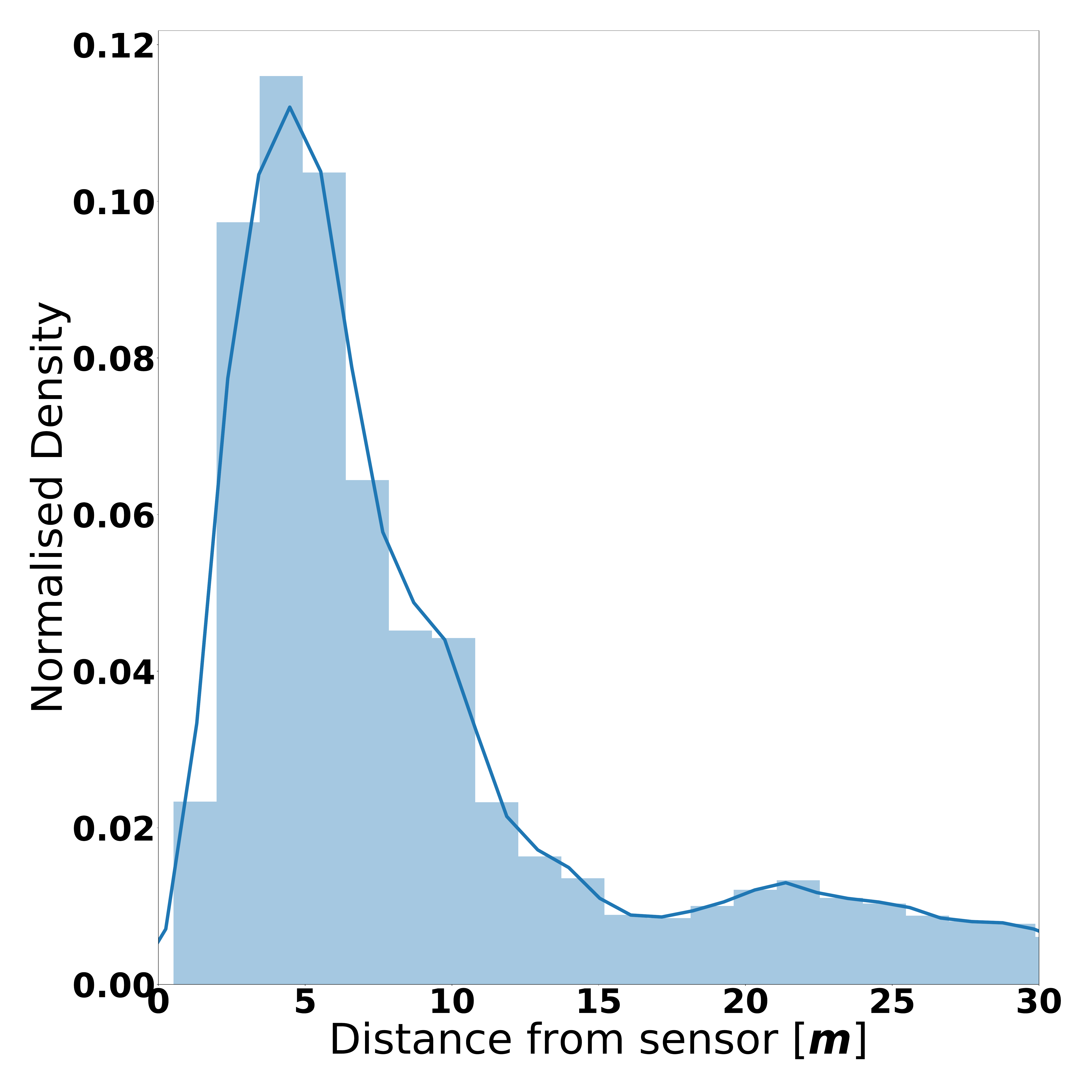}
	\caption{\revised{Validation Set}}
\end{subfigure}
 	\caption*{Histogram of distances between 3D bounding box centers and sensor origin}
 \end{subfigure}
\begin{subfigure}[b]{0.95\linewidth}
\begin{subfigure}[b]{0.48\linewidth}
    \centering
	\includegraphics[width=1.05\linewidth]{figs/depth_map_train.pdf}
	\caption{Train Set} 
\end{subfigure}
\hfill
\begin{subfigure}[b]{0.48\linewidth}
    \centering
	\includegraphics[width=1.05\linewidth]{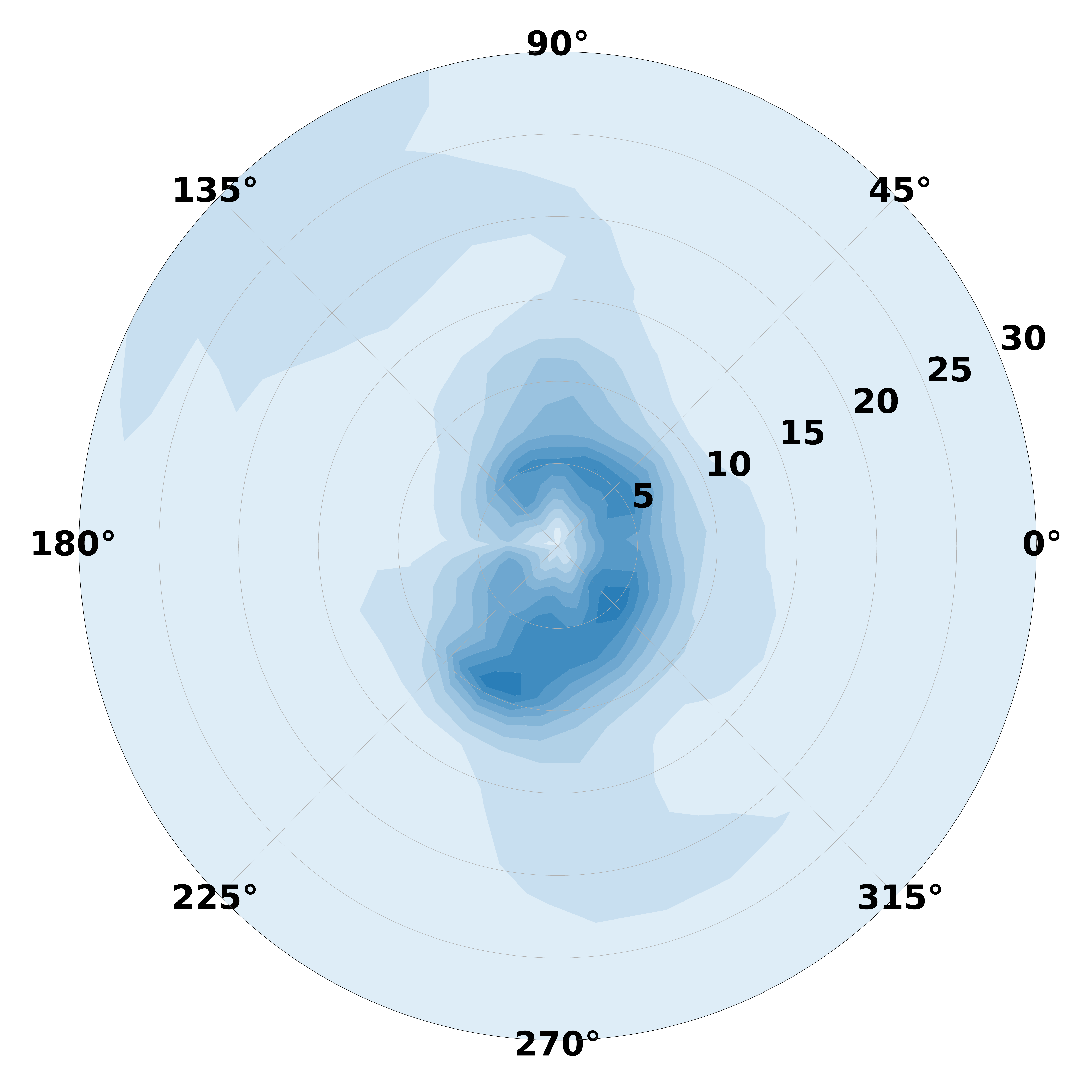}
	\caption{\revised{Validation Set}}
\end{subfigure}
 	\caption*{Kernel Density Estimates of the spatial distribution of people around the robot}
 \end{subfigure}
 \begin{subfigure}[b]{0.95\linewidth}
\begin{subfigure}[b]{0.48\linewidth}
    \centering
	\includegraphics[width=\linewidth]{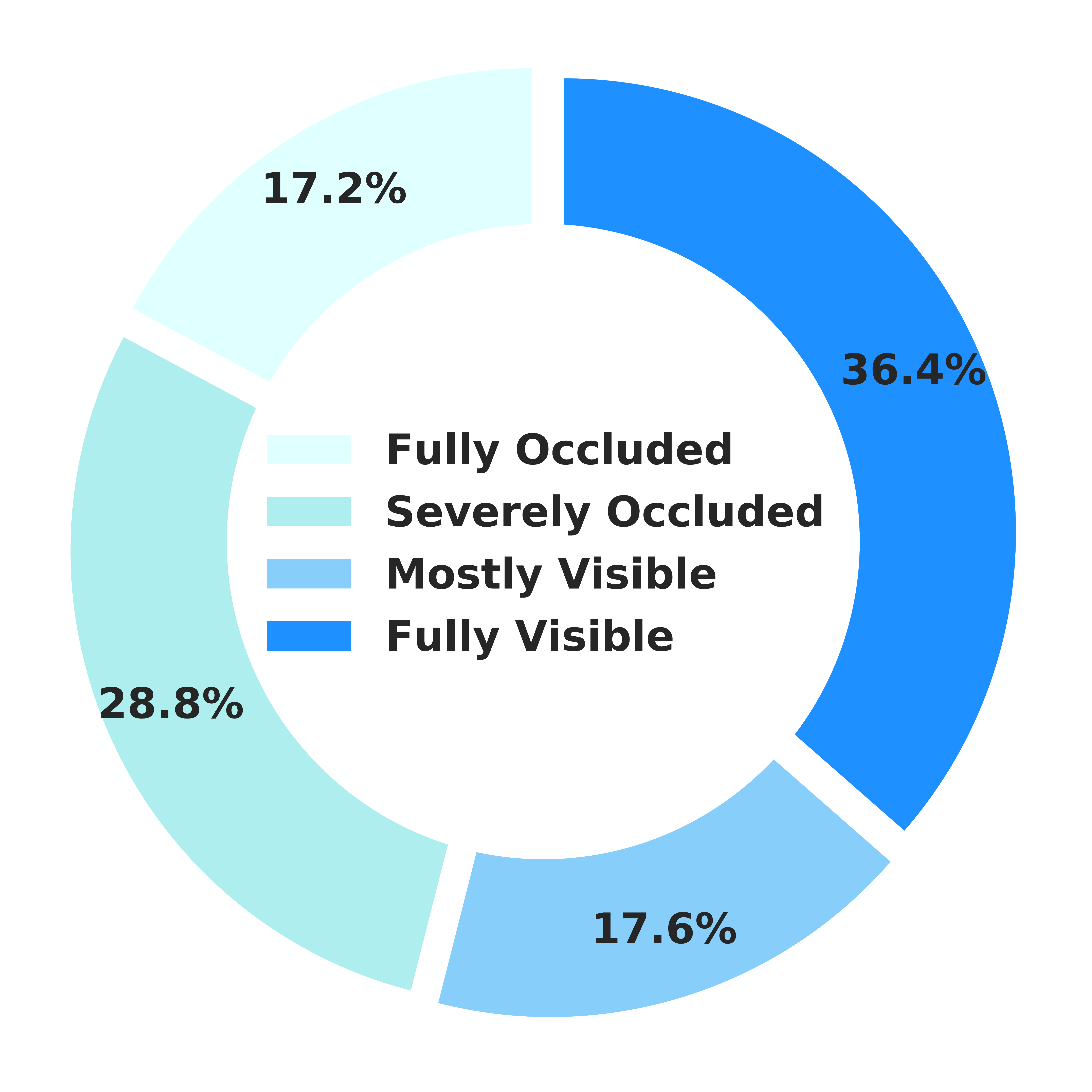}
	\caption{Train Set} 
\end{subfigure}
\hfill
\begin{subfigure}[b]{0.48\linewidth}
    \centering
	\includegraphics[width=1.05\linewidth]{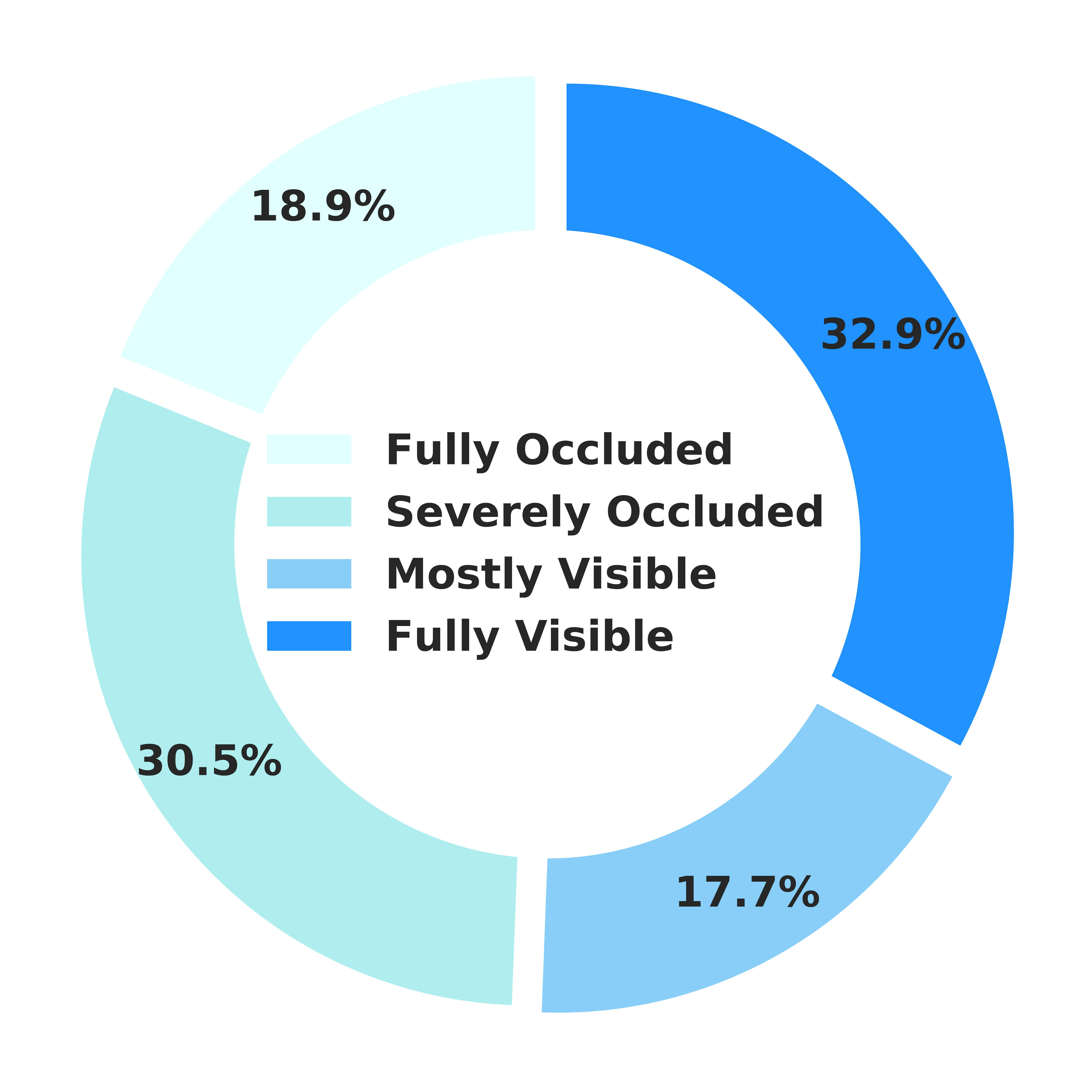}
	\caption{\revised{Validation Set}}
\end{subfigure}
 	\caption*{Distribution of occlusion levels (four levels)}
 \end{subfigure}
 \caption{\revised{Comparison of data in train and validation sets. The data distribution is very similar on the observed metrics.}}
 \label{fig:stattrainval}
\end{figure}

\newpage

\subsection{Visualization of JRDB Scenes}
\begin{figure}[!hb]
\begin{subfigure}{.5\textwidth}
  \centering
  \includegraphics[width=\linewidth]{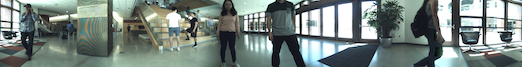}
\end{subfigure}
\hspace{-4pt}
\begin{subfigure}{.5\textwidth}
  \centering
  \includegraphics[width=\linewidth]{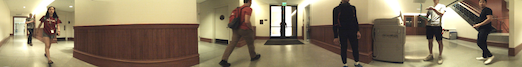}
\end{subfigure}
\begin{subfigure}{.5\textwidth}
  \centering
  \includegraphics[width=\linewidth]{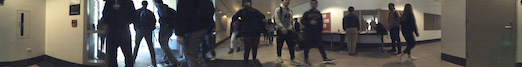}
\end{subfigure}
\hspace{-4pt}
\begin{subfigure}{.5\textwidth}
  \centering
  \includegraphics[width=\linewidth]{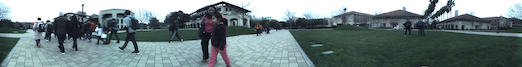}
\end{subfigure}
\begin{subfigure}{.5\textwidth}
  \centering
  \includegraphics[width=\linewidth]{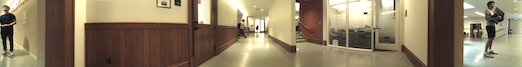}
\end{subfigure}
\hspace{-4pt}
\begin{subfigure}{.5\textwidth}
  \centering
  \includegraphics[width=\linewidth]{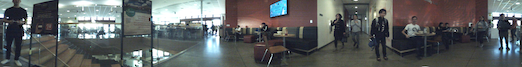}
\end{subfigure}
\begin{subfigure}{.5\textwidth}
  \centering
  \includegraphics[width=\linewidth]{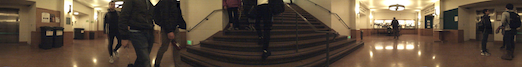}
\end{subfigure}
\hspace{-4pt}
\begin{subfigure}{.5\textwidth}
  \centering
  \includegraphics[width=\linewidth]{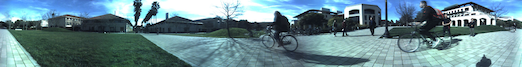}
\end{subfigure}
\begin{subfigure}{.5\textwidth}
  \centering
  \includegraphics[width=\linewidth]{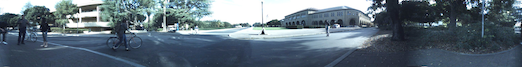}
\end{subfigure}
\hspace{-4pt}
\begin{subfigure}{.5\textwidth}
  \centering
  \includegraphics[width=\linewidth]{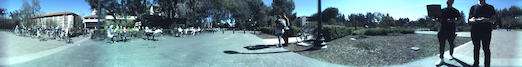}
\end{subfigure}
\begin{subfigure}{.5\textwidth}
  \centering
  \includegraphics[width=\linewidth]{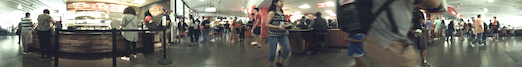}
\end{subfigure}
\hspace{-4pt}
\begin{subfigure}{.5\textwidth}
  \centering
  \includegraphics[width=\linewidth]{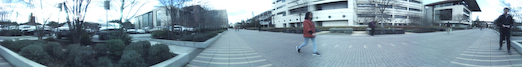}
\end{subfigure}
\begin{subfigure}{.5\textwidth}
  \centering
  \includegraphics[width=\linewidth]{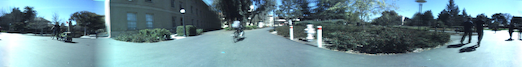}
\end{subfigure}
\hspace{-4pt}
\begin{subfigure}{.5\textwidth}
  \centering
  \includegraphics[width=\linewidth]{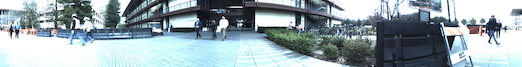}
\end{subfigure}
\begin{subfigure}{.5\textwidth}
  \centering
  \includegraphics[width=\linewidth]{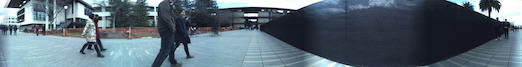}
\end{subfigure}
\hspace{-4pt}
\begin{subfigure}{.5\textwidth}
  \centering
  \includegraphics[width=\linewidth]{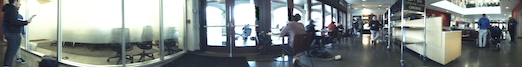}
\end{subfigure}
\caption{\revised{Visualization of the 54 sequences in the JRDB. The scenes include indoor and outdoor areas in a university campus, with different lighting conditions, motion, number of pedestrians, and activities. These include locations similar to roads, strip malls, sidewalks, restaurants, plazas, parks, halls, classrooms, laboratories, and office buildings. This scenes span a large degree of variability and cover scenerios that a social robot will encounter during operation (crowds, walking groups, queues, talking formations, and more).}}
\label{fig:scenesvis2}
\end{figure}

\begin{figure*}[tbh]
\ContinuedFloat
\begin{subfigure}{.5\textwidth}
  \centering
  \includegraphics[width=\linewidth]{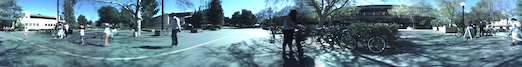}
\end{subfigure}
\hspace{-4pt}
\begin{subfigure}{.5\textwidth}
  \centering
  \includegraphics[width=\linewidth]{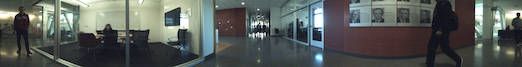}
\end{subfigure}
\begin{subfigure}{.5\textwidth}
  \centering
  \includegraphics[width=\linewidth]{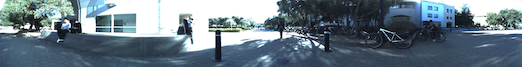}
\end{subfigure}
\hspace{-4pt}
\begin{subfigure}{.5\textwidth}
  \centering
  \includegraphics[width=\linewidth]{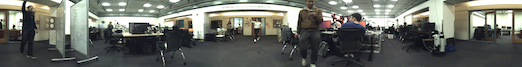}
\end{subfigure}
\begin{subfigure}{.5\textwidth}
  \centering
  \includegraphics[width=\linewidth]{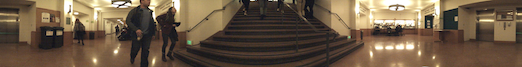}
\end{subfigure}
\hspace{-4pt}
\begin{subfigure}{.5\textwidth}
  \centering
  \includegraphics[width=\linewidth]{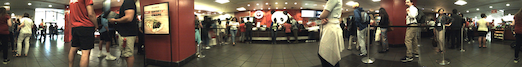}
\end{subfigure}
\begin{subfigure}{.5\textwidth}
  \centering
  \includegraphics[width=\linewidth]{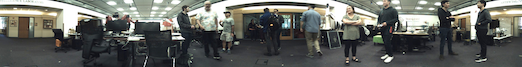}
\end{subfigure}
\hspace{-4pt}
\begin{subfigure}{.5\textwidth}
  \centering
  \includegraphics[width=\linewidth]{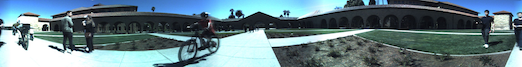}
\end{subfigure}
\begin{subfigure}{.5\textwidth}
  \centering
  \includegraphics[width=\linewidth]{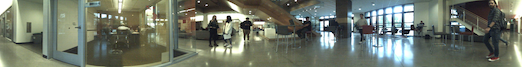}
\end{subfigure}
\hspace{-4pt}
\begin{subfigure}{.5\textwidth}
  \centering
  \includegraphics[width=\linewidth]{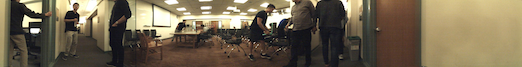}
\end{subfigure}
\begin{subfigure}{.5\textwidth}
  \centering
  \includegraphics[width=\linewidth]{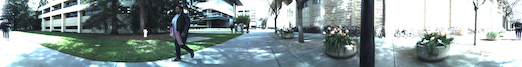}
\end{subfigure}
\hspace{-4pt}
\begin{subfigure}{.5\textwidth}
  \centering
  \includegraphics[width=\linewidth]{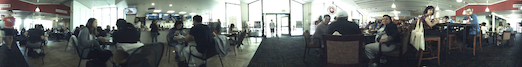}
\end{subfigure}
\begin{subfigure}{.5\textwidth}
  \centering
  \includegraphics[width=\linewidth]{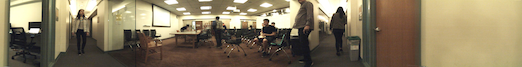}
\end{subfigure}
\hspace{-4pt}
\begin{subfigure}{.5\textwidth}
  \centering
  \includegraphics[width=\linewidth]{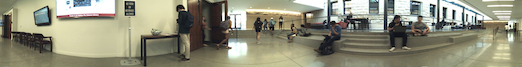}
\end{subfigure}
\begin{subfigure}{.5\textwidth}
  \centering
  \includegraphics[width=\linewidth]{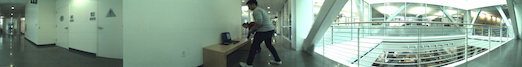}
\end{subfigure}
\hspace{-4pt}
\begin{subfigure}{.5\textwidth}
  \centering
  \includegraphics[width=\linewidth]{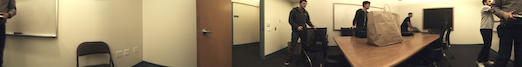}
\end{subfigure}
\begin{subfigure}{.5\textwidth}
  \centering
  \includegraphics[width=\linewidth]{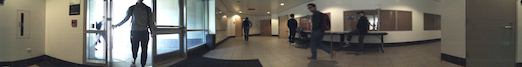}
\end{subfigure}
\hspace{-4pt}
\begin{subfigure}{.5\textwidth}
  \centering
  \includegraphics[width=\linewidth]{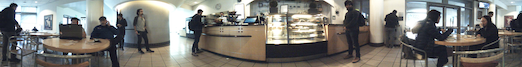}
\end{subfigure}
\begin{subfigure}{.5\textwidth}
  \centering
  \includegraphics[width=\linewidth]{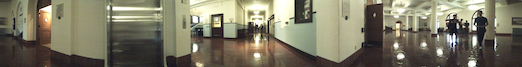}
\end{subfigure}
\hspace{-4pt}
\begin{subfigure}{.5\textwidth}
  \centering
  \includegraphics[width=\linewidth]{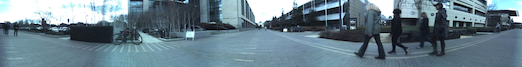}
\end{subfigure}
\begin{subfigure}{.5\textwidth}
  \centering
  \includegraphics[width=\linewidth]{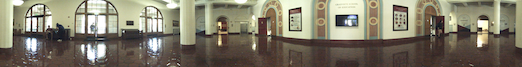}
\end{subfigure}
\hspace{-4pt}
\begin{subfigure}{.5\textwidth}
  \centering
  \includegraphics[width=\linewidth]{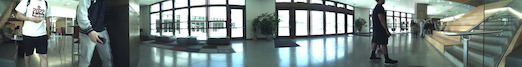}
\end{subfigure}
\begin{subfigure}{.5\textwidth}
  \centering
  \includegraphics[width=\linewidth]{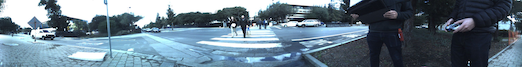}
\end{subfigure}
\hspace{-4pt}
\begin{subfigure}{.5\textwidth}
  \centering
  \includegraphics[width=\linewidth]{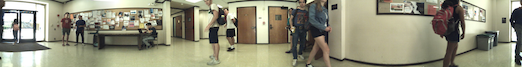}
\end{subfigure}
\begin{subfigure}{.5\textwidth}
  \centering
  \includegraphics[width=\linewidth]{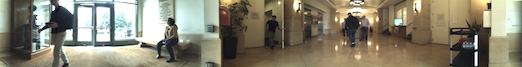}
\end{subfigure}
\hspace{-4pt}
\begin{subfigure}{.5\textwidth}
  \centering
  \includegraphics[width=\linewidth]{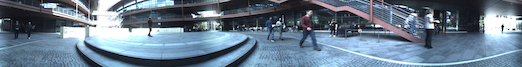}
\end{subfigure}
\begin{subfigure}{.5\textwidth}
  \centering
  \includegraphics[width=\linewidth]{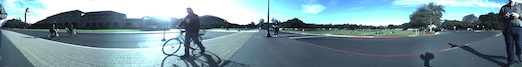}
\end{subfigure}
\hspace{-4pt}
\begin{subfigure}{.5\textwidth}
  \centering
  \includegraphics[width=\linewidth]{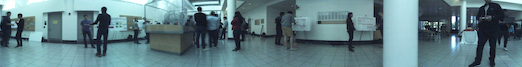}
\end{subfigure}
\begin{subfigure}{.5\textwidth}
  \centering
  \includegraphics[width=\linewidth]{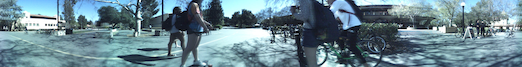}
\end{subfigure}
\hspace{-4pt}
\begin{subfigure}{.5\textwidth}
  \centering
  \includegraphics[width=\linewidth]{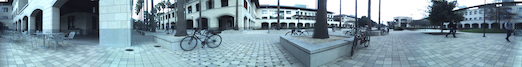}
\end{subfigure}
\begin{subfigure}{.5\textwidth}
  \centering
  \includegraphics[width=\linewidth]{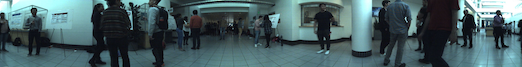}
\end{subfigure}
\hspace{-4pt}
\begin{subfigure}{.5\textwidth}
  \centering
  \includegraphics[width=\linewidth]{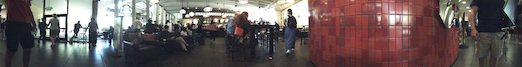}
\end{subfigure}
\begin{subfigure}{.5\textwidth}
  \centering
  \includegraphics[width=\linewidth]{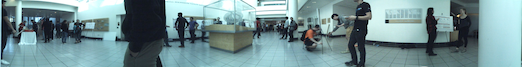}
\end{subfigure}
\hspace{-4pt}
\begin{subfigure}{.5\textwidth}
  \centering
  \includegraphics[width=\linewidth]{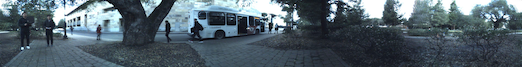}
\end{subfigure}
\begin{subfigure}{.5\textwidth}
  \centering
  \includegraphics[width=\linewidth]{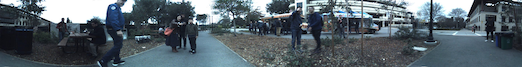}
\end{subfigure}
\hspace{-4pt}
\begin{subfigure}{.5\textwidth}
  \centering
  \includegraphics[width=\linewidth]{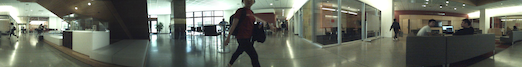}
\end{subfigure}
\begin{subfigure}{.5\textwidth}
  \centering
  \includegraphics[width=\linewidth]{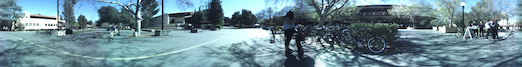}
\end{subfigure}
\hspace{-4pt}
\begin{subfigure}{.5\textwidth}
  \centering
  \includegraphics[width=\linewidth]{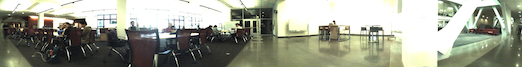}
\end{subfigure}
\caption{\revised{Visualization of the 54 sequences in the JRDB (cont.)}}
\label{fig:scenesvis}
\end{figure*}

\end{document}